\documentclass[12pt,authoryear,final]{elsarticle}
\usepackage[table]{xcolor}
\usepackage{colortbl}
\usepackage{amsmath}
\usepackage{enumitem}
\usepackage{amsmath}
\usepackage{float,times}
\usepackage{amssymb,amsthm,mathrsfs}
\usepackage{lineno}
\usepackage{epsfig}
\usepackage{color}
\usepackage{setspace}
\usepackage{latexsym}
\usepackage{tabularx}
\biboptions{square,sort&compress} 
\usepackage{algpseudocode}
\usepackage{wrapfig} % 用于插入biography
\usepackage{flushend} % 用于让最后一页的biography以行对齐
\usepackage{natbib}
\setcitestyle{numbers,square}
\usepackage{array, multirow, threeparttable}
\usepackage{tikz}
\usetikzlibrary{tikzmark}
\usepackage{subfig,graphicx,epstopdf}
\usepackage{exscale,relsize}
\usepackage{algorithm}
\usepackage[colorlinks=true,linkcolor=cyan,citecolor=cyan,urlcolor=cyan]{hyperref}
\usepackage[doipre={doi:~}]{uri}
\usepackage[capitalize]{cleveref}
\usepackage{longtable}
\allowdisplaybreaks[4]

\usepackage{booktabs}
\usepackage{adjustbox}
\usepackage{threeparttable}
\usepackage{ragged2e}
\usepackage{subcaption}
\usepackage{longtable}
\usepackage{multirow}
\usepackage{array}
\usepackage{caption}
\usepackage{amsmath} % 必须加载amsmath宏包
\usepackage{breqn}   % 加载breqn宏包
\usepackage{graphicx}
\usepackage{amsmath}
\usepackage{autobreak}
\usepackage{lscape} % 用于旋转表格
\usepackage{geometry}
\usepackage{changepage}

\newdefinition{lem}{Lemma }[section]
\newdefinition{dnt}{Definition}[section]
\newdefinition{rem}{Remark }[section]
\newdefinition{ass}{Assumption}[section]
\newdefinition{exam}{Example}[section]

\usepackage{arydshln} % 用于虚线
\usepackage{makecell}  % 提供颜色支持
\makeatletter
\def\ps@pprintTitle{%
  \let\@oddhead\@empty
  \let\@evenhead\@empty
  \def\@oddfoot{\reset@font\hfil\thepage\hfil} % 这一行保留了居中的页码
  \let\@evenfoot\@oddfoot
}
\makeatother

\begin{document}

\begin{frontmatter}

\title{Large Language Model for Discrete Optimization Problems: Evaluation and Step-by-step Reasoning}

%% Authors
\author[author1]{Tianhao Qian}
\author[author2]{Guilin Qi\corref{cor1}}
\author[author3]{Z.Y. Wu}
\author[author3]{Ran Gu}
\author[author3]{Xuanyi Liu}
\author[author1]{Canchen Lyu}

%% Affiliations
\affiliation[author1]{%
  organization={School of Mathematics, Southeast University},
  city={Nanjing},
  country={China}
}

\affiliation[author2]{%
  organization={School of Computer Science and Engineering, Southeast University},
  city={Nanjing},
  country={China}
}

\affiliation[author3]{%
  organization={School of Mathematical Sciences, Chongqing Normal University},
  city={Chongqing},
  country={China}
}

%% Corresponding author footnote
\cortext[cor1]{Corresponding authors: Guilin Qi}

\begin{abstract}

This work investigated the capabilities of different models, including the Llama-3 series of models and CHATGPT, with different forms of expression in solving discrete optimization problems by testing natural language datasets. In contrast to formal datasets with a limited scope of parameters, our dataset included a variety of problem types in discrete optimization problems and featured a wide range of parameter magnitudes, including instances with large parameter sets, integrated with augmented data. It aimed to (1) provide an overview of LLMs’ ability in large-scale problems, (2) offer suggestions to those who want to solve discrete optimization problems automatically, and (3) regard the performance as a benchmark for future research. These datasets included original, expanded and augmented datasets. Among these three datasets, the original and augmented ones aimed for evaluation while the expanded one may help finetune a new model. In the experiment, comparisons were made between strong and week models, CoT methods and No-CoT methods on various datasets. The result showed that stronger model performed better reasonably. Contrary to general agreement, it also showed that CoT technique was not always effective regarding the capability of models and disordered datasets improved performance of models on easy to-understand problems, even though they were sometimes with high variance, a manifestation of instability. Therefore, for those who seek to enhance the automatic resolution of discrete optimization problems, it is recommended to consult the results, including the line charts presented in the Appendix, as well as the conclusions drawn in this study for relevant suggestions.

\end{abstract}


\end{frontmatter}

\section{Introduction}\label{sec1}
Operations Research(OR) problems permeate nearly all areas of application, particularly in industrial production and automated control management systems. Among these problems, discrete optimization problems represent one significant type. These problems are typically solved by establishing a mathematical model followed by selecting appropriate solving methods based on the problem's scale, such as heuristic algorithms, dynamic programming\cite{shao2402deepseekmath}, neural network approaches\cite{nair2020solving}, etc. Automatically solving discrete optimization problems helps decision-making process, enabling dynamic resource allocation adjustments based on existing information to reduce costs and enhance efficiency. Moreover, the advancement of large language models (LLMs) presents promising future for further development in this domain. When solving OR problems, LLMs demonstrate unique ability to automatically integrate massive information to construct Mathematics models and select solution methods intelligently according to parameter scales.
    
Chat Generative Pre-trained Transformer, CHATGPT has dominated the online discussion since November 2022. Many researches were dedicated to improving the performance of classic downstream tasks, such as sentence classification, sentiment analysis, and Named-entity recognition, etc. With the help of techniques such as Chain of Thought(CoT)\textemdash a series of intermediate reasoning steps\cite{wei2022chain}, Program of Thought(PoT)\textemdash using language models to generate text and programming language statements, and finally an answer\cite{chen2022program}, or Retrieval-Augmented Generation (RAG) \cite{lewis2020retrieval}. After focusing on traditional NLP tasks, researchers turned to Mathematics problems and tested different models on basic Mathematics, such as MathQA and GSM8K\cite{amini2019mathqa,cobbe2021training}. In 2024, ORLM\cite{tang2024orlm}, a model to deal with OR problems, was finetuned based on LLaMA-3-8B-Instruct\cite{grattafiori2024llama}, but limited to its rigid responses. However, none of them has tested the performance of LLMs on discrete optimization problems featuring a wide range of parameter magnitudes. They focused on continuous thinking ability rather than decision-making. Our work was dedicated to bridging this gap while establishing a new dataset. 

In our work, large-scale datasets were established, including various types of discrete optimization problems in the form of natural language based on the OR library and VRP. Then, we expanded the background and changed the way the problems are narrated. Finally, the datasets were augmented by changing the order of the sentences. In terms of performance testing, the implicit form was mainly considered, even though the explicit form was also under consideration with some drawbacks. While you might query the difference when we just changed the parameter given in the problems, it had an impact on the methods LLMs utilize and further influenced the quality of the solution. The result was given in the appendix so that you can check it at any time as a reference if needed. For timeout limit, which was set at 300 seconds, it aimed to validate whether LLMs can choose the right methods to deal with problems within a limited time. Based on the experiment result, it showed that methods were not chosen randomly. 

Further, we tested the performance of our datasets, including the original one and the disordered one. To evaluate the response of LLMs, 4 indicators were adopted, including pass rate(PR), accuracy rate(AR), mean absolute percentage error(MAPE), and timeout rate(TR). With the help of these four indicators, advantages and disadvantages of the disordered datasets over those of the original ones, and performance of different models with different technique, such as CoT and PoT, were comprehensively analyzed. In addition, comparison among weak and strong models were evaluated based on these four indicators. Then, error types as well as timeout rate of different models were later explained. These analysis aimed to fully illustrate the dataset itself and identify the deterministic factor of the quality of the solution.

In conclusion, our contributions can be summarized as follows:
\begin{itemize}
	\item \textbf{Evaluation for testing LLMs' ability on large-scale discrete optimization problems: }
We introduced a new natural-language dataset, featuring a wide range of parameter magnitudes, to test LLMs' ability on various types of discrete optimization problems; see Subsection \ref{dataset}. It consisted of Assignment, 1D-Binpacking, Crew-Scheduling, Steiner, UBQP, CVRP, MDVRP, PVRP, Aircraft Landing, Generalised Assignment, Multi-dimensional Knapsack, Capacitated Warehouse Location and 2D-cutting Packing-constrained Guillotine problems in different types, such as original, expanded and disordered ones; see Table \ref{table1}.\ref{table2}. The expanded datasets changed the background automatically, transferred from other datasets, so that it could be generated endlessly, while the disordered one randomized the sentences to disturb LLMs and test whether they just rely on pattern matching to understand optimization problems.
        \item \textbf{Multiple comparisons and analysis across four indicators:} 
Based on our evaluation, we showed for what type of problems the disordered dataset has advantages over the original one, in which situation CoT technique is better than no-CoT one, the difference between weaker models and stronger models, the error of each problem and the conclusion with explanation of each one. This study could benefit future efforts to improve the overall performance. Our analysis was similar to the Mathematical Capabilities of ChatGPT\cite{frieder2023mathematical} but different on the topic itself.
        \item \textbf{Insight for which models with which technique to choose:} 
For any model with different techniques, we gave clear figures with average lines in the appendix \ref{statistic}. These figures allowed interested readers to independently decide which models and techniques to use for automatically solving large-scale discrete optimization problems. With a summary of our findings provided, it focused on certain aspects and was not align with specific needs. For this reason, the complete results were presented for public access.

\end{itemize}

\section{RELATED WORK}
\label{related work}

The automation of discrete optimization problems has been receiving a long-standing focus in research. Its significance lies in the fact that this field can be applied to solve complex issues in industry and management. Before the popularity of LLMs, scholars typically relied on pioneering and effective algorithms, building a series of repositories in Python, such as PULP, CVXPY, and Gurobi, to provide tools for the public to use conveniently. For users, they pay more attention to the effectiveness of the tool. In 2021, Hendrycks D, et al.\cite{hendrycks2021measuring} constructed a new dataset MATH, each data with step-by-step solutions. The test results on the MATH dataset demonstrated the low accuracy of machine learning methods in solving Math contest problems, even with enormous Transformer models. Besides, scaling in transformers disables solution in MATH and increasing budgets, along with the number of parameters, is impractical. Considering the low accuracy and weak transferability of traditional methods, scholars gradually resort to LLMs.

Initially, researchers just concentrated on simple mathematics problems. Chowdhery A, et al.\cite{chowdhery2023palm} tested the ability of Pathways Language Models on basic Mathematics based on some open-source datasets, such as MathQA and GSM8K \cite{amini2019mathqa, cobbe2021training}. They trained the PLM and demonstrated an outstanding performance in tasks, such as linguistic understanding and source code generation, in few-shot scenarios. Frieder S, et al.\cite{frieder2023mathematical} observed that researchers have been oblivious to the ability of ChatGPT on Mathematical reasoning by 2023 and then evaluated six subdatasets. They found that 9-January-2023 version of ChatGPT could not deal with high-quality proof and calculation, the ability of which on different advanced Mathematics problems was unstable so that ChatGPT could only act as an assistant at that time. Later, in 2024, Ahn, et al conducted a work for progresses and challenges of LLMs on mathematical reasoning. It concluded LLM-oriented techniques, factors and concerns affecting LLMs in solving math, providing a holistic perspective on the current state, accomplishments and future challenges\cite{ahn2024large}. Then, Mirzadeh, et al. introduced GSM-Symbolic, an improved benchmark created from symbolic templates that allow for the generation of a diverse set of questions, which aims to provide key insights and more reliable metrics for measuring the reasoning capabilities of models\cite{mirzadeh2024gsm}. 

Based on the Mathematical ability, researchers try to directly give LLMs tasks, including optimization and regression, to test the ability of LLMs. Yang, et al.\cite{yang2023large} viewed LLMs as an optimizer, improving the accuracy of linear regression and salesman problems on GSM8K\cite{cobbe2021training} datasets. Meanwhile, they found that the best prompting strategy was “Take a deep breath and work on this problem step-by-step.”. However, researchers have observed a significant drawback: in linear regression problems, different initializations of weights w and biases b can greatly affect the number of iterations needed to achieve the same performance and the number of parameters is exponentially related to the number of iterations. Such characteristic is unfriendly to the scenario when the time is limited. Similarly, Liu, et al.\cite{liu2402large} utilized LLMs to help Bayesian Optimization directly to fine-tune the parameters directly by iteratively proposing and evaluating promising solutions. Xiao, et al.\cite{xiao2024verbalized} introduced a new paradigm of machine learning: Verbalized Machine Learning(VML). Its approach is similar to that of the work\cite{yang2023large}, with the distinction that this paper focuses on various regression tasks, such as Linear Regression, Polynomial Regression and Sinusoidal Regression. They visualized the deterministic boundaries learned by LLMs to significantly demonstrate the strong ability of VML in solving traditional machine learning problems. Beyond traditional problems, Pluhacek, et al.\cite{pluhacek2023leveraging} concentrated on taking advantage of LLMs to generate novel hybrid swarm intelligence optimization algorithms. Several tasks were separated with prompts designed delicately but expert supervision and rigorous evaluation were still required. 

At the same time, another group of scholars aimed to address the instability and high latency of LLMs in solving OR problems in industry by fine-tuning them. In terms of optimization problems, Tang, et al.\cite{tang2024orlm} utilized self-built datasets in industry and data-augmentation technique to enlarge datasets with different classification and background based on LLAMA3-8B\cite{grattafiori2024llama}, Deepseek-Math-7B-Base\cite{bellman1966dynamic} and Mistral-7B\cite{jiang2024identifying}. Its’ fine-tuned LLAMA3-8B slightly outperformed GPT4 on simple OR problems. This research provides the industry with a multi-context OR problems dataset and the potential for training specified LLMs. Guo, et al.\cite{guo2025deepseek} developed DeepSeek-R1 by referring to reinforcement learning methods and finally overcame the disadvantage of supervised learning, improving the mathematical reasoning ability of LLMs. In terms of basic mathematics, Luo, et al.\cite{luo2023wizardmath} developed Reinforcement Learning from Evol-Instruct Feedback(RLEIF) framework to train many open-sourced models, called WizardMath, to enhance the mathematical CoT reasoning abilities of LLMs.

There have also been studies investigating human–machine collaboration using large language models on operations research(OR) problems other than utilizing LLMs to deal with them directly. Wasserkrug, et al.\cite{wasserkrug2025enhancing} combined LLMs with OR optimization to enhance decision-making. While we need to create accurate and efficient optimization models, LLMs can actually help accelerate expert work in creating formal models. 

\section{Benchmarking Architecture \& Data Representation}
\label{overview}

While previous study typically deals with mathematics problems with small-scale optimization tasks (with a small number of variables), this study aims to test the ability of LLMs on various discrete optimization problems involving a large number of variables. We will measure the performance of LLMs, analyze error types with underlying causes, as well as the description of datasets with the distribution of the number of variables. To achieve our goal, the following questions wait to be answered:

1.\textbf{RQ1:} How are the datasets constructed? What's unique in this dataset? 

2.\textbf{RQ2:} Can prompt engineering help in improving the solution or lowering the errors? Is LLMs sensitive to input
text?

3.\textbf{RQ3:} What plays a significant role in the performance of LLMs on discrete optimization problems?

4.\textbf{RQ4:} What strategies should I adopt?

To answer the first question, our constructed datasets with data augmentation will be first introduced. For the second question, the evaluation metrics will be introduced, then performance of LLMs will be measured and analyzed. For the third question, our standard datasets will be compared with augmented datasets. For the last question, see Section \ref{conclusion}.

\subsection{Dataset} \label{dataset}
\label{overview}
\textbf{Statistics:} Datasets are collected from part of OR Library and VRP, including 1D-Binpacking, Assignment, Crew Scheduling, Steiner, UBQP, a series of VRP problems, Aircraft Landing and other combinatorial problems. OR-Library is a collection of test data sets for a variety of OR problems originally described in J.E. Beasley and VRP was collected by NEO Research Group. Table \ref{table1}.\ref{table11} show the basic information of the datasets.

\begin{table}[htbp]
  \caption{Basic Information of Each Type of Problems (Part 1)}
  \label{table1}
  \centering
  \scriptsize            % 如需更小字号可改为 \tiny
  \renewcommand{\arraystretch}{1.2}  % 调整行高
  \setlength{\tabcolsep}{4pt}        % 缩小列间距
  \begin{adjustbox}{} 
    \begin{tabularx}{\textwidth}{@{}lXr@{\hspace{6mm}}lXr@{}}
      \toprule
      \multicolumn{3}{c@{\hspace{6mm}}}{\textbf{Left Section}} & \multicolumn{3}{c}{\textbf{Right Section}} \\
      \cmidrule(lr){1-3} \cmidrule(l){4-6}
      \textbf{Problems} & \textbf{Mathematical Model} & \textbf{Number of Instances}
        & \textbf{Problems} & \textbf{Mathematical Model} & \textbf{Number of Instances} \\
      \midrule

      Assignment &
      $\displaystyle
      \begin{aligned}[t]
\min\;Z &= \sum_{i=1}^n\sum_{j=1}^n c_{ij}\,x_{ij},\\
\text{s.t. } 
& \sum_{i=1}^n x_{ij} = 1,\quad \forall\,j=1,\dots,n,\\
& \sum_{j=1}^n x_{ij} = 1,\quad \forall\,i=1,\dots,n,
\end{aligned}$ & 12
      &
      Steiner$^*$ &
      $\displaystyle
      \begin{aligned}[t]
&\min_{\textbf p\in\mathbb R^2}Z 
= \sum_{i=1}^n \bigl\|\textbf p - (x_i, y_i)\bigr\|_2 \\ 
&= \sum_{i=1}^n \sqrt{(p_x - x_i)^2 + (p_y - y_i)^2},
\end{aligned}$ & 196 \\
      \addlinespace[3pt]

      1D-Binpacking &
      $\displaystyle
      \begin{aligned}[t]
\min\;Z &= m,\\
\text{s.t. } 
& \sum_{j=1}^m x_{ij} = 1,\quad \forall\,i=1,\dots,n,\\
& \sum_{i=1}^n w_i\,x_{ij} \le C,\quad \forall\,j=1,\dots,m,
\end{aligned}$ & 160
      &
      2D-cutting &
      $\displaystyle
      \begin{aligned}[t]
\max\;Z &= \sum_{j=1}^n p_j\,x_j,\\
\text{s.t. } 
& \sum_{j=1}^n a_j\,x_j \le A,\\
& \sum_{j=1}^n b_j\,x_j \le B,
\end{aligned}$ & 3 \\
      \addlinespace[3pt]

      Crew-scheduling &
      $\displaystyle
      \begin{aligned}[t]
\min\;Z &= \sum_{i=1}^n\sum_{j=1}^m c_{ij}\,x_{ij},\\
\text{s.t. } 
& \sum_{j=1}^m x_{ij} = 1,\quad \forall\,i=1,\dots,n,\\
& \sum_{i=1}^n t_i\,x_{ij} \le T_j,\quad \forall\,j=1,\dots,m,
\end{aligned}$ & 10
      &
      UBQP &
      $\displaystyle
      \begin{aligned}[t]
\min\;Z &= \sum_{i=1}^n\sum_{j=1}^n Q_{ij}\,x_i\,x_j,\\
\text{s.t. } 
& x_i \in \{0,1\},\quad i=1,\dots,n,
\end{aligned}$ & 45 \\
      \bottomrule
    \end{tabularx}
  \end{adjustbox}
\end{table}

\begin{table}[htbp]
  \caption{Basic Information of Each Type of Problems (Part 2)}
  \label{table11}
  \centering
  \scriptsize            % 如需更小字号可改为 \tiny
  \renewcommand{\arraystretch}{1.2}  % 调整行高
  \setlength{\tabcolsep}{4pt}        % 缩小列间距
  \begin{adjustbox}{} 
    \begin{tabularx}{\textwidth}{@{}lXr@{\hspace{6mm}}lXr@{}}
      \toprule
      \multicolumn{3}{c@{\hspace{6mm}}}{\textbf{Left Section}} & \multicolumn{3}{c}{\textbf{Right Section}} \\
      \cmidrule(lr){1-3} \cmidrule(l){4-6}
      \textbf{Problems} & \textbf{Mathematical Model} & \textbf{Number of Instances}
        & \textbf{Problems} & \textbf{Mathematical Model} & \textbf{Number of Instances} \\
      \midrule
        CVRP &
      $\displaystyle
      \begin{aligned}[t]
\min\;Z &= \sum_{i=0}^n\sum_{j=0}^n c_{ij}\,x_{ij},\\
\text{s.t. } 
& \sum_{j=1}^n x_{0j} = k,\\
& \sum_{j=0}^n x_{ij} = 1,\quad \forall\,i=1,\dots,n,\\
& \sum_{i=1}^n q_i\,x_{ij} \le Q,\quad \forall\,j=1,\dots,k,
\end{aligned}$ & 202
      &
      PVRP &
      $\displaystyle
      \begin{aligned}[t]
\min\;Z &= \sum_{i=0}^n\sum_{j=1}^n\sum_{k=1}^K c_{ij}\,x_{ijk},\\
\text{s.t. } 
& \sum_{k=1}^K y_{jk} = 1,\quad \forall\,j=1,\dots,n,\\
& \sum_{j=0}^n x_{0jk} = 1,\quad \forall\,k=1,\dots,K,\\
& \sum_{i=1}^n q_i\,x_{ijk} \le Q_k,\quad \forall\,k=1,\dots,K,
\end{aligned}$ & 42 \\
      \addlinespace[3pt]

      MDVRP &
      $\displaystyle
      \begin{aligned}[t]
\min\;Z &= \sum_{i=0}^n\sum_{j=1}^n\sum_{k=1}^K c_{ij}\,x_{ijk},\\
\text{s.t. } 
& \sum_{k=1}^K y_{jk} = 1,\quad \forall\,j=1,\dots,n,\\
& \sum_{j=0}^n x_{0jk} = 1,\quad \forall\,k=1,\dots,K,\\
& \sum_{i=1}^n q_i\,x_{ijk} \le Q_k,\quad \forall\,k=1,\dots,K,
\end{aligned}$ & 33
      &
      Aircraft Landing &
      $\displaystyle
      \begin{aligned}[t]
\min\;Z &= \sum_{i=1}^n d_i\,L_i,\\
\text{s.t. } 
& L_i \ge L_{i-1} + \sum_{j=1}^n T_{ij}\,x_{ij},\quad \forall\,i=2,\dots,n,\\
& \sum_{j=1}^n x_{ij} = 1,\quad \forall\,i=1,\dots,n,
\end{aligned}$ & 13 \\
      \addlinespace[3pt]

      Generalised Assignment &
      $\displaystyle
      \begin{aligned}[t]
\min\;Z &= \sum_{i=1}^n\sum_{j=1}^m c_{ij}\,x_{ij},\\
\text{s.t. } 
& \sum_{j=1}^m x_{ij} = 1,\quad \forall\,i=1,\dots,n,\\
& \sum_{i=1}^n w_{ij}\,x_{ij} \le W_j,\quad \forall\,j=1,\dots,m,
\end{aligned}$ & 60
      &
      Multi-Knapsack &
      $\displaystyle
      \begin{aligned}[t]
\max\;Z &= \sum_{i=1}^m\sum_{j=1}^n v_j\,x_{ij},\\
\text{s.t. } 
& \sum_{j=1}^n w_{ij}\,x_{ij} \le W_i,\quad \forall\,i=1,\dots,m,\\
& \sum_{i=1}^m x_{ij} \le 1,\quad \forall\,j=1,\dots,n,
\end{aligned}$ & 277 \\
      \addlinespace[3pt]

      Warehouse Location &
      $\displaystyle
      \begin{aligned}[t]
\min\;Z &= \sum_{j=1}^m f_j\,y_j \;+\; \\ & \sum_{i=1}^n\sum_{j=1}^m c_{ij}\,x_{ij},\\
\text{s.t. } 
& \sum_{j=1}^m x_{ij} = 1,\quad \forall\,i=1,\dots,n,\\
& \sum_{i=1}^n d_i\,x_{ij} \le Q_j\,y_j,\quad \forall\,j=1,\dots,m,
\end{aligned}$ & 34
      & & & \\
      \bottomrule
    \end{tabularx}
  \end{adjustbox}
\end{table}

In these two datasets, the data is raw so that it needs transforming into the form of natural language. Generally, we refer to related paper and use expert-annotated method to achieve our goal.
In detail, we constructed the PVRP datasets by referring a work\cite{golden2008vehicle}; Aircraft Landing datasets by referring to the work of Pang Y et al.\cite{pang2024machine} and Beasley J E et al.\cite{beasley2004displacement}; Generalised Assignment problems datasets by referring to the work of Maniezzo V et al.\cite{maniezzo2021matheuristics}; Multi-dimensional Knapsack problems datasets by referring to the work of Fréville A\cite{freville2004multidimensional} and Puchinger J et al.\cite{puchinger2010multidimensional}; Capacitated Warehouse Location problems datasets by referring to the work of Kelly D L et al.\cite{kelly1982capacitated}; 2D-cutting Packing with Constrained Guillotine problems problems datasets by referring to the work of Morabito R et al.\cite{morabito1996staged} and Martin M et al.\cite{martin2020models}. The remaining datasets not mentioned above are annotated using the expert-annotated method. However, both of these methods require a lot of energy and even the paper-based method that fails to transform the problem into different forms significantly impacts our evaluation of LLMs’ capabilities in solving discrete optimization problems. Additionally, the uniformity of the dataset scenarios hinders the fine-tuning of the base model. Thus, the diversity of this dataset should be increased.

\textbf{Data Generation and Expansion:}
For data generation, annotated samples are used to formulate a standard dataset manually. The samples include several implicit problems and one explicit problem, which means the mathematics model is given explicitly. For data expansion, Ilia Shumailov, et al. conducted a work\cite{shumailov2024ai} published on Nature, which reveals that the use of LLMs at scale to publish content on the Internet will pollute the collection of data to train their successors\cite{shumailov2024ai}. This finding suggests that generating data by LLMs should be approached seriously. In this way, other datasets whose problems of data contain background were first selected. Then, GPT-4o-mini was utilized to extract these backgrounds and asked to creatively continue the story and stored as a database. Finally, for any piece of implicit problems, a new background was randomly selected from the database to take place of the older one. However, these types of datasets are still too perfect. We aimd to introduce some noise to this natural language data for the purpose of data augmentation.

\begin{figure}[htbp!]
\centering
\includegraphics[width=0.9\textwidth,keepaspectratio]{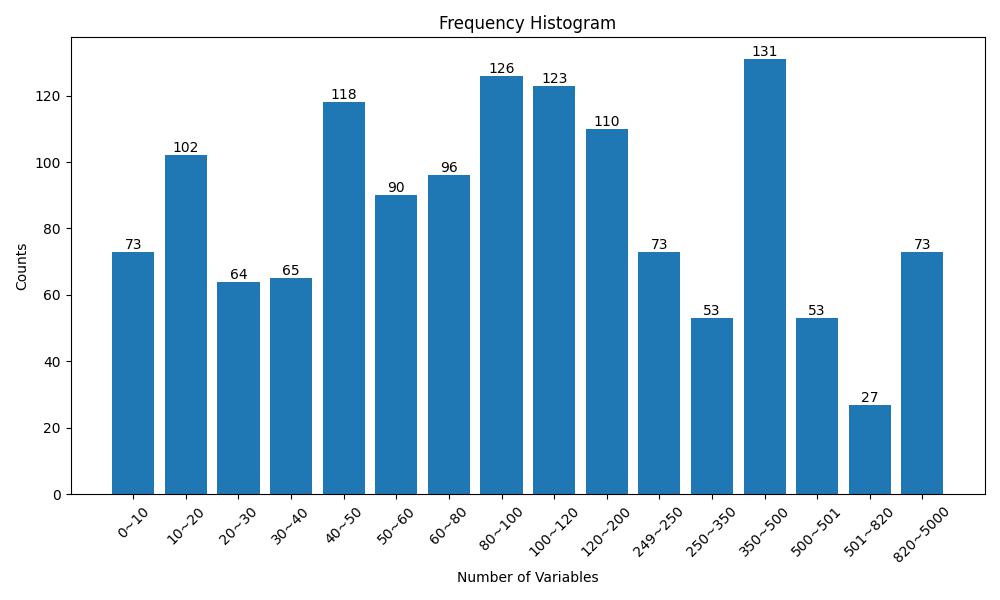}
\caption{Statistics on the Number of Variables}
\label{fig1}
\end{figure}

\textbf{Data Augmentation:}
The sentences will be broken up into segments randomly, which destroys the logic of the sentences. It aims to test whether LLMs fully understand problems themselves or just do pattern matching, which will also be analyzed in the next section. \textbf{Table} \ref{table2}. and \textbf{Figure} \ref{fig1}. show the number of processed, expanded and augmented datasets and statistic on the number of variables.

%经典三线表
\begin{table}[H]
\caption{Statistics on Various Datasets}%标题
\label{table2}
\centering%把表居中
\begin{tabular}{cccc}%四个c代表该表一共四列，内容全部居中
\toprule%第一道横线
PROBLEMS&MANUAL&EXPANDED&AUGMENTED\\
\midrule%第二道横线 
Assignment&60&48&108 \\
1D-Binpacking&960&640&1600 \\
Crew-scheduling&150&100&250 \\
Steiner&980&980&1960 \\
UBQP&135&90&225 \\
CVRP&606&404&1010 \\
MDVRP&132&88&220 \\
PVRP&126&84&210 \\
Aircraft Landing&26&26&52 \\
Generalised assignment&180&120&300 \\
Multi-dimensional knapsack&1108&841&1949 \\
Capacitated Warehouse Location&102&68&170 \\
2D-cutting Packing-constrained Guillotine&18&12&30 \\
\bottomrule%第三道横线
\end{tabular}
\end{table}

\subsection{Evaluation Benchmark} 

In our study, evaluation benchmark is defined to systematically compare the performance of our datasets on different models and techniques.

A preliminary experiment showed that it’s impossible for LLMs to implement complex iterations without code. Our strategy is to optimize the problems after obtaining the result by executing the code generated by LLMs with some instructions necessary to instruct models to generate correct codes. For instance, the data may be too large for LLMs to read so that it’s stored in the format of text in advance. It avoids hallucination when reading the data, as well as reduces the generation cost. Besides, we set a threshold for codes executed so that the dead end will not interfere with other test data. The evaluation is conducted with prompts to regularize the output generated by LLMs without disturbing the performance and the instructions and codes are given in the appendix. 

\textbf{Evaluation Metrics(EM)} We adopt four evaluation metrics, i.e., Exact Match, Pass Rate, Mean Absolute Percentage Error and Timeout Rate denoted as AR, PR, MAPE and TR respectively. Specifically, $I$ represents the indicator function and $|I_{A}|$ represents measure of the domain that meets $x\in A$, where $x$ refers to a problem. In detail, $\lvert I_{optimal}\rvert$ indicates the measure of $\left\{x\right\}$\textemdash the solution of $x$ reaches or surpasses the optimal result, $\lvert I_{pass}\rvert$ indicates the measure of $\left\{x\right\}$\textemdash the solution code of $x$ can be executed successfully, $\lvert I_{\leq 300s}\rvert$ indicates the measure of $\left\{x\right\}$\textemdash the execution time of the code within 300s, and $\lvert I_{overall}\rvert$ indicates the measure of $\left\{x\right\}$.

\textbf{Accuracy Rate(AR)} aims to capture whether LLMs find the optimal value. When the objective is to minimize the result, the output is considered correct if the executed result is less than the reference value. Similarly, for maximization, the result is valid if it exceeds the reference. In this way, we have:

\begin{flalign}
  && \textbf{AR} = \frac{|\textbf{I}_{optimal}|}{|\textbf{I}_{overall}|} &&
\end{flalign}

\textbf{Pass Rate(PR)} is presented to measure whether the generated codes are executable. If the number of failed execution of generated codes exceeds a specified threshold, such as 6 times, it will be counted as a failure; otherwise, the code of the solution will be accepted. In detail, when the execution fails, the error information will be given to LLMs to help them correct mistakes. In this way, we have:

\begin{flalign}
    &&
    \textbf{PR}=\frac{\lvert I_{pass}\rvert}{\lvert I_{overall}\rvert}
    &&
\end{flalign}

\textbf{Mean Absolute Percentage Error(MAPE)} serves as a complement to AR. For continuous data, it can represent the disparity between predicted values and actual values, which can further elucidate the results in AR metrics. In this way, we have:

\begin{flalign}
    &&
    \textbf{MAPE}=\frac{100\%}{n}\sum_{i=1}^{n}{|\frac{\hat{y}_i-y_i}{y_i}|}
    &&
\end{flalign}

\textbf{Timeout Rate(TR)} is presented to measure the percentage of solution within limited time, set 300 seconds heare This indicator aims evaluate whether different models can autonomously select the most suitable optimization method to solve problems within constrained timeframes across diverse problem sets. In this way, we have:

\begin{flalign}
    &&
    \textbf{TR}=\frac{\lvert I_{\leq 300s}\rvert}{\lvert I_{pass}\rvert}
    &&
\end{flalign}

\textbf{Baselines:} We adopted different techniques, such as CoT\cite{wei2022chain} and PoT\cite{chen2022program}, and models such as GPT-4o-mini, DeepSeek-R1\cite{guo2025deepseek}, LLAMA3-8B-Instruct\cite{grattafiori2024llama} using float16 precision and ORLM\cite{tang2024orlm} to test our original datasets. In this work, the first two models are considered stronger models, while the latter two models are regarded as weaker ones. In detail, GPT-4o-mini and LLAMA3-8B are general-purpose LLMs, ORLM is finetuned based on LLAMA3-8B and DeepSeek-R1 with strong Mathematical ability, is also based on LLAMA3-8B. CoT prompting is completed by GPT-4o-mini, who gives the complete instructions of steps to deal with a certain kind of problems. In the following text, LLAMA3-8B-Instruct will be simplified as LLAMA3-8B.

\subsection{Error Categories}

\begin{table}[h]
    \centering
    \caption{Error Types and Descriptions}
    \begin{tabular}{lp{10cm}} % 设置描述列宽度为10cm
        \toprule
        \textbf{Error Types} & \textbf{Description} \\
        \midrule
        IndexError & Raised when trying to access an index that is out of range for a list or a sequence. \\
        FileNotFoundError & Raised when a file or directory is requested but cannot be found. \\
        OverflowError & Raised when a calculation exceeds the limits of the data type, typically in numeric operations. \\
        NameError & Raised when a local or global name is not found, meaning the variable or function is not defined. \\
        ValueError & Raised when a function receives an argument of the right type but an inappropriate value, such as converting a non-numeric string to an integer. \\
        AttributeError & Raised when an invalid attribute reference is made, meaning the attribute does not exist on the object. \\
        KeyError & Raised when attempting to access a dictionary with a key that does not exist. \\
        SyntaxError & Raised when the Python interpreter encounters invalid syntax in the code, preventing execution. \\
        TypeError & Raised when an operation or function is applied to an object of inappropriate type, such as adding a string to an integer. \\
        ImportError & Raised when a module or package cannot be imported, often due to it not being found or having issues within it. \\
        PulpError & Raised in the context of optimization problems, indicating issues such as infeasibility or unbounded solutions. \\
        SolutionError & Raised when the solution of a problem is invalid, often related to disobeying constraints. \\
        ResultError & Raised when a function does not yield an expected form of result, often due to output error by LLMs. \\
        \bottomrule
    \end{tabular}
    \label{DESCRIPTION}
\end{table}

When generated codes, results or solutions make mistakes, these errors will be recorded. Here, we list several error types and descriptions in Table \ref{DESCRIPTION}. The preliminary experiment showed that IndexError, ValueError, TypeError, KeyError and SyntaxError are the most frequent among these errors. Typically, IndexError arises from omissions that occur when LLMs attempt to load data sequentially. Then, the interpreter tries to access an index that is out of range and raises this error. ValueError arises from output error, which means the codes fail to output result as is regulated by prompt engineering. For TypeError, it orrcurs when using Python’s built-in solver to construct a mathematical model, resulting in a symbolic error. Namely the code fails to correctly use the solver. SyntaxError results from the code missing a bracket or parentheses. The other errors are described in Table \ref{DESCRIPTION}.

\subsection{Representations}
We suppose that different forms of expression of one problem will lead to different performance. One problem can be explicit or implicit. One example along with that in disorder is given in the Table \ref{EXAMPLE}. 

Typically, datasets with explicit form are more costly but without limitation while datasets with implicit form are cheaper but limited to PoT techniques. Besides, the former will meet long sequential problems with LLMs slacking off, omitting essential data, while the later will be limited to the performance of OR problem solvers without fully utilizing the computational advantages of LLMs. Therefore, both forms of expression have their respective potential applications.

\section{Experiments}
Before analyzing the result, we suppose that a larger number of parameters will result in worse performance, including a decrease in PR and AR, and an increase in MAPE, caused by long sequential problems. Besides, We assume that CoT is supposed to improve the overall performance based on previous research. Before testifying to this initial assumption, it must be clarified that in our experiments, LLMs tend to omit data when dealing with explicit forms of problems or may produce omissions due to hallucination issues when reproducing data. Additionally, LLMs may take shortcuts by fabricating data, which is the reason why explicit forms of problems are not used for testing later. Furthermore, this approach is also costly and is not recommended for addressing discrete optimization problems.

\begin{table}[!htbp]
    \centering
    \caption{A case study of 1D-Binpacking problem}
    \label{EXAMPLE}
    \begin{threeparttable}
        \begin{tabular}{p{0.4\textwidth} p{0.4\textwidth} @{}c@{}}
            \toprule
            \textbf{\textcircled{1}Implicit} & \textbf{\textcircled{1}Explicit} & 
            \multirow{4}{*}{\shortstack{\textcircled{1} \\ \rule{0.4pt}{15pt} \\ \rule{0.4pt}{15pt} \\\rule{0.4pt}{15pt} \\\rule{0.4pt}{15pt} \\\rule{0.4pt}{15pt} \\\rule{0.4pt}{15pt} \\\rule{0.4pt}{15pt} \\ \rule{0.4pt}{15pt} \\\rule{0.4pt}{15pt} \\\rule{0.4pt}{15pt} \\\rule{0.4pt}{15pt} \\\rule{0.4pt}{15pt} \\\rule{0.4pt}{15pt} \\\rule{0.4pt}{15pt} \\$\downarrow$}} \\ % 合并4行
            \cmidrule(lr){1-2}
            \raggedright
            There are 120 items S and several boxes B with a capacity of 150. The volumes of the 120 items are \textbf{[BLANK]}, and it is given that $s_i \leq 150$ ($1\leq i \leq 120$). The packing problem seeks to pack all items into the boxes in such a way that the number of boxes used is minimized. The following table shows the volumes of the items: \textbf{[BLANK]}  
            & There are 120 items S and several boxes B with a capacity of 150. The volumes of the 120 items are $s_1, s_2, \ldots, s_{120}$, and it is given that $s_i \leq 150$ ($1 \leq i \leq 120$). The packing problem seeks to pack all items into the boxes in such a way that the number of boxes used is minimized. The following table shows the volumes of the items:
            $x_1$,$x_2$,{\ldots}, $x_{119}$,$x_{120}$&\\
            \cmidrule(lr){1-2}
            \multicolumn{2}{c}{\textbf{\textcircled{1}Implicit-Disorder}} &  \\
            \cmidrule(lr){1-2}
            \multicolumn{2}{p{0.8\textwidth}}{%
                \raggedright
                The packing problem seeks to pack all items into the boxes in such a way that the number of boxes used is minimized The following table shows the volumes of the items: $x_1$ $x_2$ {\ldots} $x_{119}$ $x_{120} s_{120}$, and it is given that $s_i \leq 150$ ($1\leq i\leq120$) 1.There are 120 items S and several boxes B with a capacity of 150. The volumes of the 120 items are $s_{1}, s_{2},$
            }\\
            \cmidrule(lr){1-2}
            \multicolumn{2}{c}{\textbf{\textcircled{2}CoT Prompting(Optional)}} & \multirow{4}{*}{\shortstack{\textcircled{2} \\ \rule{0.4pt}{15pt} \\ \rule{0.4pt}{15pt} \\\rule{0.4pt}{15pt} \\\rule{0.4pt}{15pt} \\\rule{0.4pt}{15pt} \\\rule{0.4pt}{15pt} \\\rule{0.4pt}{15pt} \\ \rule{0.4pt}{15pt} \\\rule{0.4pt}{15pt} \\\rule{0.4pt}{15pt} \\\rule{0.4pt}{15pt} \\\rule{0.4pt}{15pt} \\\rule{0.4pt}{15pt} \\\rule{0.4pt}{15pt} \\$\downarrow$}} \\ % 合并4行
            \cmidrule(lr){1-2}
            \multicolumn{2}{p{0.8\textwidth}}{%
  \raggedright
  Dealing with 1D bin‑packing problems involves several key steps. Here’s a concise outline: \\[0.5ex]
  1. Define the Problem: Clearly state the items to be packed (sizes and quantities) and the capacity of the bins. \\ 
  2. Choose a Strategy: Select a bin‑packing algorithm or heuristic (First‑Fit, Best‑Fit, FFD, BFD). \\  
  3. Sort the Items: If using a decreasing strategy, sort the items in descending order. \\ 
  4. Initialize Bins: Create an empty list or array to represent the bins. \\ 
  5. Pack the Items: Iterate through the sorted list and place each item into the first suitable bin. \\ 
  6. Handle Overflows: If an item cannot fit, create a new bin. \\ 
  7. Track Usage: Record how many bins are used and their contents. \\ 
  8. Evaluate the Solution: Assess packing efficiency (number of bins used, waste). \\ 
  9. Optimize (if needed): Explore further optimization techniques. \\ 
  10. Implement and Test: Run with various inputs to ensure correctness and efficiency.
}\\
            \cmidrule(lr){1-2}

            \multicolumn{2}{c}{\textbf{\textcircled{3}Common Prompting}} & \textcircled{3} \\ % 合并4行 \\
            \cmidrule(lr){1-2}
            \multicolumn{2}{p{0.8\textwidth}}{%
                \raggedright
                Output the python program that stores the optimum value into result and then into file: output.txt(only store optimum value without any redundant expression).
            } & \\
            \bottomrule
        \end{tabular}
    \end{threeparttable}
\end{table}

\clearpage

In this part, we will first present performance metrics along with a comparison among datasets, including PR, AR, MAPE and TR. Next, we will conclude the trend of LLMs on discrete optimization problems, and what is worth noticing for each type of problems. Then, each error of problems will be given and explained. Finally, TR will be presented at the end of this section.

\begin{table}[ht]
\centering
\caption{PR Task Performance Comparison}
\label{PR}
\resizebox{\textwidth}{!}{% 
\begin{tabular}{llcccccccc}
\toprule
\multirow{2}{*}{Task} & \multirow{2}{*}{Dataset} & \multicolumn{2}{c}{GPT-4o-mini} & \multicolumn{2}{c}{LLAMA3-8B} & \multicolumn{2}{c}{ORLM} & \multicolumn{2}{c}{DeepSeek-R1}\\
\cmidrule(lr){3-4} \cmidrule(lr){5-6}\cmidrule(lr){7-8}\cmidrule(lr){9-10}
 & & PoT & PoT+CoT & PoT & PoT+CoT &PoT & PoT+CoT &PoT & PoT+CoT \\
\midrule
\multirow{3}{*}{Assignment} 
& Explicit-origin & \textbackslash{}  & $76.67$ & \textbackslash{}   &  \textbackslash{} &  \textbackslash{} &  \textbackslash{} &  \textbackslash{} &  \textbackslash{}\\
& Implicit-origin & $80.88$  & $91.67$ & $9.09$  & $43.33$ &$0.00$&$8.33$ &\textbf{\textbf{98.33}}&$85.00$\\
& Implicit-disorder & \textbackslash{}  & \textbf{\textbf{92.59}} & $13.89$  & $39.81$ & $0.93$  & $9.26$ & \textbackslash{}  & \textbackslash{}  \\
\cmidrule(lr){1-10}
\multirow{3}{*}{1D-Binpacking}
& Explicit-origin & \textbackslash{}  & $83.37$ & \textbackslash{}  & \textbackslash{} & \textbackslash{} & \textbackslash{} & \textbackslash{} & \textbackslash{}  \\
& Implicit-origin & $64.06$  & $75.73$ & $55.21$  & $12.50$ & $3.54$ & $8.12$ & $80.47$ & \textbf{\textbf{82.55}}  \\
& Implicit-disorder & \textbackslash{}  & \textbf{\textbf{83.19}} & $7.35$  & $9.96$ & $3.25$  & $3.98$ & \textbackslash{}  & \textbackslash{} \\
\cmidrule(lr){1-10}
\multirow{3}{*}{Crew-scheduling}
& Explicit-origin & \textbackslash{}  & $76.00$ & \textbackslash{}  & \textbackslash{} & \textbackslash{} & \textbackslash{} & \textbackslash{} & \textbackslash{}  \\
& Implicit-origin & $52.00$  & $52.67$ & $4.67$  & $3.33$ & $0.00$ & $2.00$ & \textbf{\textbf{79.33}} & $67.59$  \\
& Implicit-disorder & \textbackslash{}  & $44.31$ & $3.60$  & $1.60$& \textbf{\textbf{72.31}}  & $0.80$ & \textbackslash{}  & \textbackslash{}  \\
\cmidrule(lr){1-10}
\multirow{3}{*}{Steiner}
& Explicit-origin & \textbackslash{}  & $84.48$ & \textbackslash{}  & \textbackslash{} & \textbackslash{} & \textbackslash{} & \textbackslash{} & \textbackslash{}  \\
& Implicit-origin & $87.86$  & $93.67$ & $20.73$  & $14.29$ & $1.12$ & $0.10$ & $76.30$ & \textbf{\textbf{97.10}}  \\
& Implicit-disorder & \textbackslash{}  & \textbf{\textbf{23.54}} & $3.06$  & $13.68$ & $0.92$  & $1.86$ & \textbackslash{}  & \textbackslash{}  \\
\cmidrule(lr){1-10}
\multirow{3}{*}{UBQP}
& Explicit-origin & \textbackslash{}  & 60.74 & \textbackslash{}  & \textbackslash{} & \textbackslash{} & \textbackslash{} & \textbackslash{} & \textbackslash{}  \\
& Implicit-origin & $17.78$  & $62.69$ & $16.30$  & $8.89$ & $1.48$ & $2.96$ & $64.44$ & \textbf{\textbf{93.33}}  \\
& Implicit-disorder & \textbackslash{}  & \textbf{\textbf{30.46}} & $14.67$  & $4.00$ & $0.89$  & $5.78$& \textbackslash{}  & \textbackslash{}  \\
\cmidrule(lr){1-10}
\multirow{3}{*}{CVRP}
& Explicit-origin & \textbackslash{}  & $64.00$ & \textbackslash{}  & \textbackslash{} & \textbackslash{} & \textbackslash{} & \textbackslash{} & \textbackslash{}  \\
& Implicit-origin & $36.15$  & $47.06$ & $2.12$  & $0.49$ & $0.16$ & $0.49$ & $71.16$ & \textbf{\textbf{83.87}}  \\
& Implicit-disorder & \textbackslash{}  & \textbf{\textbf{49.31} } & $3.04$  & $1.10$ & $0.59$  & $0.49$& \textbackslash{}  & \textbackslash{}  \\
\cmidrule(lr){1-10}
\multirow{3}{*}{MDVRP}
& Explicit-origin & \textbackslash{}  & $53.79$ & \textbackslash{}  & \textbackslash{} & \textbackslash{} & \textbackslash{} & \textbackslash{} & \textbackslash{}  \\
& Implicit-origin & $75.76$  & $77.27$ & $0.76$  & $5.30$ & $0.00$ & $0.00$ & \textbf{\textbf{93.18}} & $46.21$ \\
& Implicit-disorder & \textbackslash{}  & \textbf{\textbf{75.76}} & $6.57$  & $2.02$& $0.51$  & $1.52$& \textbackslash{}  & \textbackslash{}  \\
\cmidrule(lr){1-10}
\multirow{3}{*}{PVRP}
& Explicit-origin & \textbackslash{}  & $71.43$ & \textbackslash{}  & \textbackslash{} & \textbackslash{} & \textbackslash{} & \textbackslash{} & \textbackslash{}  \\
& Implicit-origin & $72.22$  & $67.46$ & $2.38$  & $2.38$&$0.79$&$0.00$&\textbf{\textbf{79.37}}&$71.31$  \\
& Implicit-disorder & \textbackslash{}  & \textbf{\textbf{73.81}} & $2.38$  & $2.38$& $0.00$  & $0.95$& \textbackslash{}  & \textbackslash{}  \\
\cmidrule(lr){1-10}
\multirow{3}{*}{Aircraft Landing}
& Explicit-origin & \textbackslash{}  & $56.82$ & \textbackslash{}  & \textbackslash{} & \textbackslash{} & \textbackslash{} & \textbackslash{} & \textbackslash{}  \\
& Implicit-origin & \textbf{\textbf{77.27}}  & $68.18$ & $11.36$  & $9.09$&$0.00$&$0.00$&$68.18$&$0.00$  \\
& Implicit-disorder & \textbackslash{}  & \textbf{\textbf{67.05}} & $18.18$  & $7.95$& $0.00$  & $1.14$& \textbackslash{}  & \textbackslash{}  \\
\cmidrule(lr){1-10}
\multirow{3}{*}{Generalised Assignment}
& Explicit-origin & \textbackslash{}  & $70.56$ & \textbackslash{}  & \textbackslash{} & \textbackslash{} & \textbackslash{} & \textbackslash{} & \textbackslash{}  \\
& Implicit-origin & \textbf{\textbf{76.11}}  & $58.89$ & $6.67$  &$6.67$&$1.67$&$1.67$&$24.00$&$53.85$  \\
& Implicit-disorder & \textbackslash{}  & \textbf{\textbf{76.00}} & $8.00$  & $5.00$& $2.33$  & $1.33$& \textbackslash{}  & \textbackslash{}  \\
\cmidrule(lr){1-10}
\multirow{3}{*}{Multi-dimensional Knapsack}
& Explicit-origin & \textbackslash{}  & $43.12$ & \textbackslash{}  & \textbackslash{} & \textbackslash{} & \textbackslash{} & \textbackslash{} & \textbackslash{}  \\
& Implicit-origin & \textbf{\textbf{42.42}}  & $41.47$ & $3.88$  & $4.24$&$0.19$&$0.72$&$29.56$&$31.34$  \\
& Implicit-disorder & \textbackslash{}  & \textbf{\textbf{43.17}} & $2.64$  & $2.54$& $0.37$  & $1.33$& \textbackslash{}  & \textbackslash{}  \\
\cmidrule(lr){1-10}
\multirow{3}{*}{Capacitated Warehouse Location}
& Explicit-origin & \textbackslash{}  & $31.37$ & \textbackslash{}  & \textbackslash{} & \textbackslash{} & \textbackslash{} & \textbackslash{} & \textbackslash{}  \\
& Implicit-origin & $25.49$  & $30.39$ & $0.00$  &$ 0.00$&$1.96$&$0.00$&\textbf{\textbf{53.92}}&$16.67$  \\
& Implicit-disorder & \textbackslash{}  &\textbf{\textbf{ 41.10}} & $2.35$  & $0.59$& $0.00$  & $1.10$& \textbackslash{}  & \textbackslash{}  \\
\cmidrule(lr){1-10}
\multirow{3}{*}{2D-cutting Packing}
& Explicit-origin & \textbackslash{}  & $83.33$ & \textbackslash{}  & \textbackslash{} & \textbackslash{} & \textbackslash{} & \textbackslash{} & \textbackslash{}  \\
& Implicit-origin & $77.78$  & \textbf{\textbf{83.33}} & $16.67$  & $11.11$ & $0.00$ & $0.00$ & $38.89$ & $50.00$  \\
& Implicit-disorder & \textbackslash{}  & \textbf{\textbf{72.41}} & $0.00$  & $0.00$ & $0.00$  & $0.00$ & \textbackslash{}  & \textbackslash{}  \\
\bottomrule
\end{tabular}%
}
\end{table}

\begin{table}[ht]
\centering
\caption{AR Task Performance Comparison}
\label{AR}
\resizebox{\textwidth}{!}{ % 使用 \resizebox 来调整表格大小
\begin{tabular}{llcccccccc}
\toprule
\multirow{2}{*}{Task} & \multirow{2}{*}{Dataset} & \multicolumn{2}{c}{GPT-4o-mini} & \multicolumn{2}{c}{LLAMA3-8B} & \multicolumn{2}{c}{ORLM} & \multicolumn{2}{c}{DeepSeek-R1}\\
\cmidrule(lr){3-4} \cmidrule(lr){5-6}\cmidrule(lr){7-8}\cmidrule(lr){9-10}
 & & PoT & PoT+CoT & PoT & PoT+CoT &PoT & PoT+CoT &PoT & PoT+CoT \\
\midrule

\multirow{3}{*}{Assignment} 
& Explicit-origin & \textbackslash{}  & $0.00$ & \textbackslash{}   &  \textbackslash{} &  \textbackslash{} &  \textbackslash{} &  \textbackslash{} &  \textbackslash{}\\
& Implicit-origin & $4.41$  & \textbf{\textbf{10.00}} & $2.27$  & $5.00$ &$0.00$&$0.00$&$8.33$&$6.67$ \\
& Implicit-disorder & \textbackslash{}  & \textbf{\textbf{11.11}} & $0.00$  & $4.63$& $0.00$  & $0.93$& \textbackslash{}  & \textbackslash{}  \\
\cmidrule(lr){1-10}

\multirow{3}{*}{1D-Binpacking}
& Explicit-origin & \textbackslash{}  & $10.85$ & \textbackslash{}  & \textbackslash{} & \textbackslash{} & \textbackslash{} & \textbackslash{} & \textbackslash{}  \\
& Implicit-origin & $5.94$  & $4.79$ & $6.04$  & $1.25$&$0.62$&$0.00$&$9.65$&\textbf{\textbf{10.07}}  \\
& Implicit-disorder & \textbackslash{}  & \textbf{\textbf{4.48}} & $0.56$  & $1.72$ & $0.44$  & $0.54$& \textbackslash{}  & \textbackslash{} \\
\cmidrule(lr){1-10}

\multirow{3}{*}{Crew-scheduling}
& Explicit-origin & \textbackslash{}  & $0.00$ & \textbackslash{}  & \textbackslash{} & \textbackslash{} & \textbackslash{} & \textbackslash{} & \textbackslash{}  \\
& Implicit-origin & $1.33$  & $0.00$ & $0.00$  & $0.00$&$0.00$&$0.00$& \textbf{\textbf{27.33}} & $25.00$  \\
& Implicit-disorder & \textbackslash{}  & $0.00$ & $0.00$  & $0.00$& \textbf{\textbf{21.54}}  & $0.00$& \textbackslash{}  & \textbackslash{}  \\
\cmidrule(lr){1-10}

\multirow{3}{*}{Steiner}
& Explicit-origin & \textbackslash{}  & $7.46$ & \textbackslash{}  & \textbackslash{} & \textbackslash{} & \textbackslash{} & \textbackslash{} & \textbackslash{}  \\
& Implicit-origin & $6.84$  & $7.24$ & $2.02$  & $1.63$&$0.20$&$0.00$& \textbf{\textbf{75.52}}& $6.47$  \\
& Implicit-disorder & \textbackslash{}  & $0.37$ & $0.41$  & \textbf{\textbf{1.10}}& $0.05$  & $0.26$& \textbackslash{}  & \textbackslash{}  \\
\cmidrule(lr){1-10}

\multirow{3}{*}{UBQP}
& Explicit-origin & \textbackslash{}  & 0.74 & \textbackslash{}  & \textbackslash{} & \textbackslash{} & \textbackslash{} & \textbackslash{} & \textbackslash{}  \\
& Implicit-origin & $0.00$  & $0.75$ &$ 0.74$  & $0.74$&$0.00$&$0.74$& \textbf{\textbf{3.70}}&$2.22$  \\
& Implicit-disorder & \textbackslash{}  & \textbf{\textbf{0.57}} & $0.44$  & $0.00$& $0.00$  & $0.44$& \textbackslash{}  & \textbackslash{}  \\
\cmidrule(lr){1-10}

\multirow{3}{*}{CVRP}
& Explicit-origin & \textbackslash{}  & $3.11$ & \textbackslash{}  & \textbackslash{} & \textbackslash{} & \textbackslash{} & \textbackslash{} & \textbackslash{}  \\
& Implicit-origin & $5.41$  & $3.59$ & $0.00 $ & $0.00$ & $0.00$ & $0.00$& \textbf{\textbf{6.15}}&$2.69$  \\
& Implicit-disorder & \textbackslash{}  & \textbf{\textbf{3.14}} & $0.10$  & $0.00$& $0.00$  & $0.10$& \textbackslash{}  & \textbackslash{}  \\
\cmidrule(lr){1-10}
\multirow{3}{*}{MDVRP}
& Explicit-origin & \textbackslash{}  & $11.36$ & \textbackslash{}  & \textbackslash{} & \textbackslash{} & \textbackslash{} & \textbackslash{} & \textbackslash{}  \\
& Implicit-origin & $6.06$ & \textbf{\textbf{8.33}} & $0.00$  &$ 0.00$ & $0.00$ & $0.00$& $4.55$&$1.52$  \\
& Implicit-disorder & \textbackslash{}  & \textbf{\textbf{11.62}} & $1.01$  & $0.00$& $0.00$  & $0.00$& \textbackslash{}  & \textbackslash{}  \\
\cmidrule(lr){1-10}
\multirow{3}{*}{PVRP}
& Explicit-origin & \textbackslash{}  & $3.97$ & \textbackslash{}  & \textbackslash{} & \textbackslash{} & \textbackslash{} & \textbackslash{} & \textbackslash{}  \\
& Implicit-origin & $10.32$  & $7.94$ & $0.00$  & $0.00$&$0.00$&$0.00$& \textbf{\textbf{11.90}}&$11.48$  \\
& Implicit-disorder & \textbackslash{}  &\textbf{\textbf{ 11.43}} & $0.00$  & $0.00$& $0.00$  & $0.00$& \textbackslash{}  & \textbackslash{}  \\
\cmidrule(lr){1-10}
\multirow{3}{*}{Aircraft Landing}
& Explicit-origin & \textbackslash{}  & $4.55$ & \textbackslash{}  & \textbackslash{} & \textbackslash{} & \textbackslash{} & \textbackslash{} & \textbackslash{}  \\
& Implicit-origin & $11.36$  & $4.55$ & $0.00$  &$0.00$&$0.00$&$0.00$& \textbf{\textbf{45.45}}& $0.00$  \\
& Implicit-disorder & \textbackslash{}  & \textbf{\textbf{5.68}} & $1.14$  & $0.00$& $0.00$  & $0.00$& \textbackslash{}  & \textbackslash{}  \\
\cmidrule(lr){1-10}
\multirow{3}{*}{Generalised Assignment}
& Explicit-origin & \textbackslash{}  & $22.78$ & \textbackslash{}  & \textbackslash{} & \textbackslash{} & \textbackslash{} & \textbackslash{} & \textbackslash{}  \\
& Implicit-origin & \textbf{\textbf{12.22}}  & $11.67$ & $3.33$  &$2.78$&$0.56$&$0.56$& $8.00$&$11.54$  \\
& Implicit-disorder & \textbackslash{}  & \textbf{\textbf{23.00}} & $3.33$  & $2.33$& $0.67$  & $0.33$& \textbackslash{}  & \textbackslash{}  \\

\cmidrule(lr){1-10}
\multirow{3}{*}{Multi-dimensional Knapsack}
& Explicit-origin & \textbackslash{}  & $3.75$ & \textbackslash{}  & \textbackslash{} & \textbackslash{} & \textbackslash{} & \textbackslash{} & \textbackslash{}  \\
& Implicit-origin & $0.72$  & $0.59$ & $0.36$  & $0.27$&$0.00$&$0.00$& \textbf{\textbf{0.94}}& $0.00$  \\
& Implicit-disorder & \textbackslash{}  & \textbf{\textbf{1.09}} & $0.18$  & $0.15$& $0.00$  & $0.00$& \textbackslash{}  & \textbackslash{}  \\

\cmidrule(lr){1-10}
\multirow{3}{*}{Capacitated Warehouse Location}
& Explicit-origin & \textbackslash{}  & $5.88$ & \textbackslash{}  & \textbackslash{} & \textbackslash{} & \textbackslash{} & \textbackslash{} & \textbackslash{}  \\
& Implicit-origin & $0.00$  & $0.00$ & $0.00$  & $0.00$&$0.00$&$0.00$& \textbf{\textbf{11.76}}&$2.94$  \\
& Implicit-disorder & \textbackslash{}  & \textbf{\textbf{5.52}} & $0.59$  & $0.00$& $0.00$  & $0.00$& \textbackslash{}  & \textbackslash{}  \\
\cmidrule(lr){1-10}

\multirow{3}{*}{2D-cutting Packing}
& Explicit-origin & \textbackslash{}  & 27.78 & \textbackslash{}  & \textbackslash{} & \textbackslash{} & \textbackslash{} & \textbackslash{} & \textbackslash{}  \\
& Implicit-origin & $5.56$  &$ 11.11$ & $0.00$  & $0.00$ & $0.00$ & $0.00$& $16.67$&\textbf{\textbf{33.33}}  \\
& Implicit-disorder & \textbackslash{}  & \textbf{\textbf{24.14}} & $0.00$  & $0.00$& $0.00$  & $0.00$& \textbackslash{}  & \textbackslash{}  \\

% 其余任务按相同模式添加...
\bottomrule
\end{tabular}
}
\end{table}

\begin{table}[ht]
\centering
\caption{MAPE Task Performance Comparison}
\label{MAPE}
\resizebox{\textwidth}{!}{ % 使用 \resizebox 来调整表格大小
\begin{tabular}{llcccccccc}
\toprule
\multirow{2}{*}{Task} & \multirow{2}{*}{Dataset} & \multicolumn{2}{c}{GPT-4o-mini} & \multicolumn{2}{c}{LLAMA3-8B} & \multicolumn{2}{c}{ORLM} & \multicolumn{2}{c}{DeepSeek-R1}\\
\cmidrule(lr){3-4} \cmidrule(lr){5-6}\cmidrule(lr){7-8}\cmidrule(lr){9-10}
 & & PoT & PoT+CoT & PoT & PoT+CoT &PoT & PoT+CoT &PoT & PoT+CoT \\
\midrule

\multirow{3}{*}{Assignment} 
& Explicit-origin & \textbackslash{}  & - & \textbackslash{}   &  \textbackslash{} &  \textbackslash{} &  \textbackslash{} &  \textbackslash{} &  \textbackslash{}\\
& Implicit-origin & $1.969$  & $1.969$ & -  & $1.876$ &$0.00$&\textbf{\textbf{0.534}}& $1.847$&$1.944$ \\
& Implicit-disorder & \textbackslash{}  & $1.838$ & -  & $2.276$& -  & \textbf{\textbf{0.220}}& \textbackslash{}  & \textbackslash{}  \\
\cmidrule(lr){1-10}

\multirow{3}{*}{1D-Binpacking}
& Explicit-origin & \textbackslash{}  & $0.089$ & \textbackslash{}  & \textbackslash{} & \textbackslash{} & \textbackslash{} & \textbackslash{} & \textbackslash{}  \\
& Implicit-origin & $0.2284$  & \textbf{\textbf{0.092}} & $0.727$  & $0.691$ & $1.267$ &$1.182$& $0.249$& $0.292$  \\
& Implicit-disorder & \textbackslash{}  & \textbf{\textbf{0.181}} & $0.811$  & $0.539$ & $1.526$  & $1.839$& \textbackslash{}  & \textbackslash{} \\
\cmidrule(lr){1-10}

\multirow{3}{*}{Crew-scheduling}
& Explicit-origin & \textbackslash{}  & $1.694$ & \textbackslash{}  & \textbackslash{} & \textbackslash{} & \textbackslash{} & \textbackslash{} & \textbackslash{}  \\
& Implicit-origin & -  & $0.801$ & $2.691$  & - & - & -& $0.671$& \textbf{\textbf{0.421}}  \\
& Implicit-disorder & \textbackslash{}  & $1.866$ & -  & -& \textbf{\textbf{0.973}}  & -& \textbackslash{}  & \textbackslash{}  \\
\cmidrule(lr){1-10}

\multirow{3}{*}{Steiner}
& Explicit-origin & \textbackslash{}  & $0.758$ & \textbackslash{}  & \textbackslash{} & \textbackslash{} & \textbackslash{} & \textbackslash{} & \textbackslash{}  \\
& Implicit-origin & $1.269$  & $0.704$ & $1.751$  & $1.548$&$1.895$ & - & \textbf{\textbf{0.033}} & $0.926$  \\
& Implicit-disorder & \textbackslash{}  & $1.474$ & \textbf{\textbf{1.345}}  & $1.677$& $2.095$  & $1.390$& \textbackslash{}  & \textbackslash{}  \\
\cmidrule(lr){1-10}

\multirow{3}{*}{UBQP}
& Explicit-origin & \textbackslash{}  & $0.579$ & \textbackslash{}  & \textbackslash{} & \textbackslash{} & \textbackslash{} & \textbackslash{} & \textbackslash{}  \\
& Implicit-origin & $0.391$  & \textbf{\textbf{0.315}} & -  & $0.577$ & - & - & $0.347$& $0.459$ \\
& Implicit-disorder & \textbackslash{}  & \textbf{\textbf{0.336}} & $0.601$  & - & -  & -& \textbackslash{}  & \textbackslash{}  \\
\cmidrule(lr){1-10}

\multirow{3}{*}{CVRP}
& Explicit-origin & \textbackslash{}  & $0.428$ & \textbackslash{}  & \textbackslash{} & \textbackslash{} & \textbackslash{} & \textbackslash{} & \textbackslash{}  \\
& Implicit-origin & $0.887$  & $0.424$ & $3.429$  & - & - & -& $0.641$& \textbf{\textbf{0.083}}  \\
& Implicit-disorder & \textbackslash{}  & \textbf{\textbf{0.722}} & -  & - & -  & -& \textbackslash{}  & \textbackslash{}  \\
\cmidrule(lr){1-10}
\multirow{3}{*}{MDVRP}
& Explicit-origin & \textbackslash{}  & $1.226$ & \textbackslash{}  & \textbackslash{} & \textbackslash{} & \textbackslash{} & \textbackslash{} & \textbackslash{}  \\
& Implicit-origin & $0.836$ & $0.719$ & -  & - & - & - & \textbf{\textbf{0.698}}& $0.975$  \\
& Implicit-disorder & \textbackslash{}  & $1.300$ & \textbf{\textbf{0.451}}  & - & -  & -& \textbackslash{}  & \textbackslash{}  \\
\cmidrule(lr){1-10}
\multirow{3}{*}{PVRP}
& Explicit-origin & \textbackslash{}  & $1.401$ & \textbackslash{}  & \textbackslash{} & \textbackslash{} & \textbackslash{} & \textbackslash{} & \textbackslash{}  \\
& Implicit-origin & $1.501$  & $0.884$ & -  & - & - & -& \textbf{\textbf{0.798}}& $1.086$  \\
& Implicit-disorder & \textbackslash{}  & \textbf{\textbf{0.887}} & $3.200$  & - & -  & -& \textbackslash{}  & \textbackslash{}  \\
\cmidrule(lr){1-10}
\multirow{3}{*}{Aircraft Landing}
& Explicit-origin & \textbackslash{}  & - & \textbackslash{}  & \textbackslash{} & \textbackslash{} & \textbackslash{} & \textbackslash{} & \textbackslash{}  \\
& Implicit-origin & $3.187$  & - & -  &$1.774$&-&- & \textbf{\textbf{0.892}}& -  \\
& Implicit-disorder & \textbackslash{}  & \textbf{\textbf{0.381}} & -  & - & -  & -& \textbackslash{}  & \textbackslash{}  \\
\cmidrule(lr){1-10}
\multirow{3}{*}{Generalised Assignment}
& Explicit-origin & \textbackslash{}  & $1.371$ & \textbackslash{}  & \textbackslash{} & \textbackslash{} & \textbackslash{} & \textbackslash{} & \textbackslash{}  \\
& Implicit-origin & $1.053$  & $1.048$ & $0.741 $ & \textbf{\textbf{0.024}} &$0.437$&-& $0.543$&$0.606$  \\
& Implicit-disorder & \textbackslash{}  & \textbf{\textbf{0.868}} & $2.397$  & $0.872$& -  & -& \textbackslash{}  & \textbackslash{}  \\

\cmidrule(lr){1-10}
\multirow{3}{*}{Multi-dimensional Knapsack}
& Explicit-origin & \textbackslash{}  & $4.387$ & \textbackslash{}  & \textbackslash{} & \textbackslash{} & \textbackslash{} & \textbackslash{} & \textbackslash{}  \\
& Implicit-origin & $1.776$  & $1.756$ & \textbf{\textbf{1.087}}  & $1.163$ & - & $2.068$&$ 2.236$& $2.345$  \\
& Implicit-disorder & \textbackslash{}  & $1.953$ & $1.800$  & $1.691$& \textbf{\textbf{0.453}}  & -& \textbackslash{}  & \textbackslash{}  \\

\cmidrule(lr){1-10}
\multirow{3}{*}{Capacitated Warehouse Location}
& Explicit-origin & \textbackslash{}  & $0.842$ & \textbackslash{}  & \textbackslash{} & \textbackslash{} & \textbackslash{} & \textbackslash{} & \textbackslash{}  \\
& Implicit-origin & -  & - & -  & - & - & -& \textbf{\textbf{0.016}}&$0.017 $ \\
& Implicit-disorder & \textbackslash{}  & \textbf{\textbf{0.064}} & $2.089$  & -& -  & -& \textbackslash{}  & \textbackslash{}  \\
\cmidrule(lr){1-10}

\multirow{3}{*}{2D-cutting Packing}
& Explicit-origin & \textbackslash{}  & $0.341$ & \textbackslash{}  & \textbackslash{} & \textbackslash{} & \textbackslash{} & \textbackslash{} & \textbackslash{}  \\
& Implicit-origin & $0.507$  & $0.353$ & $0.562$  & - & - & -& $0.776$&\textbf{\textbf{0.213}}  \\
& Implicit-disorder & \textbackslash{}  & \textbf{\textbf{0.576}} & -  & - & -  & -& \textbackslash{}  & \textbackslash{}  \\

% 其余任务按相同模式添加...
\bottomrule
\end{tabular}
}
\end{table}

\renewcommand{\thesubsection}{\Alph{subsection}} % 将子章节编号格式更改为字母

\subsection{Quantitative Analysis}

For PR metric, DeepSeek-R1 demonstrates overwhelming performance on implicit expression problems from the original dataset, only underperforming GPT-4o-mini on Aircraft Landing, Generalized Assignment, and Multi-dimensional Knapsack tasks. This aligns with existing research highlighting DeepSeek-R1's exceptional mathematical reasoning capabilities. Notably, CoT prompting doesn't universally improve PR. In detail, stronger models may benefit CoT while weaker models may not. That's because CoT divides the problems into several steps and weaker models can not deal with these steps consistently. While there is an increase in PR with disordered datasets for GPT-4o-mini, a stronger model, it's just the opposite for weaker models in general. Also, ORLM exhibits persistent format compliance issues regardless of CoT application, likely due to overfitting in training, making PR evaluations less meaningful for this model. In summary, in terms of PR metric, the comprehensive ability in understanding problems are of great importance(That's why stronger model performs better.) and disordered datasets may benefit stronger models in some types of problems by providing the optimization goal ahead of time. Based on the previouse work in 2025, we learn that LLMs broadly follow Bayesian posterior updates, with deviations primarily due to miscalibrated priors rather than flawed updates\cite{gupta2025enough}. Therefore, when the problem is randomized, with optimization objective brought forward inadvertently, the Bayesian probability distribution is updated, leading the LLMs to interpret subsequently presented inputs more precisely. The specific underlying mechanism falls outside the scope of our work and may be addressed in future work. However, PR does not necessarily means better performance for the results of the solution have not been compared.

For AR metric, GPT-4o-mini demonstrates significant advantages with CoT on original implicit datasets, while DeepSeek-R1 excels with PoT. In general, DeepSeek-R1 is more powerful. Typically, CoT has its auxiliary nature and thus fails to help reach precise solutions. However, though its prompts may influence the selection of built-in optimization solvers in Python, some of which can be invoked by providing the type of problems with formatted data and others of which needs artificial coding of constraints. That choice will lead to different performance. Similar to PR, disordered datasets help achieve better performance for stronger models but the opposite for weaker models. While we can not explain it for GPT-4o-mini, ORLM's computational limitations and formatting issues jointly explain its generally poor AR. Despite this, its performance on Crew Scheduling problems suggests mathematical potential slightly inferior to DeepSeek-R1 with proper refinement. Comprehensively, ORLM performs the best in Crew-scheduling problems if the scale of the model is taken into consideration. Thus, finetuned model is effective to some degree. 
In summary, while stronger models may perform better in solving discrete optimization problems, CoT is not necessarily effective for them, which depends on models themselves.  For weaker models, we recommend no-CoT technique without disordered datasets while for stronger models, CoT technique may help improve the performance and disordered datasets can be explored if further improvement is needed.

For Mean Absolute Percentage Error(MAPE) metric, null(-) means the result can not be accepted, resulting from timeouts or extreme outliers $\geq$ 500\%, necessitating data exclusion. The result shows that LLAMA3 and ORLM frequently produce null MAPE values, typically due to outliers by checking Fig.\ref{timeout}. For stronger models, DeepSeek-R1 generally perform better with either CoT or No-CoT technique while GPT-4o-mini performs better with CoT technique only. However, GPT-4o-mini on disordered datasets just performs worse than original ones. Consequently, combined with AR metric, the variance of the result is comparably high, which means that one needs to overtake the risk if it is used.

In summary, for both stronger and weaker models, CoT technique is strongly recommended to ensure performance robustness under any condition. Of course, practitioners may strategically employ disordered datasets when emphasizing prediction accuracy, but need to overtake the risk in instability.

\subsection{Error Case Analysis}
In this part, a case study is conducted for each task. From Table \ref{tab:error_analysis_en}, the Assignment, Crew Scheduling, UBQP, and Generalized Assignment Problem frequently encounter errors related to list operations, resulting in IndexError. In contrast, the 1D Bin Packing, MDVRP, Multi-dimensional Knapsack Problems, and 2D Cutting Packing Problems often experience errors during data reading, leading to ValueError. The CVRP and PVRP problems typically encounter issues when invoking Python’s built-in solvers, such as incorrect function calls and parameter mismatches, resulting in TypeError. The Aircraft Landing problem, due to its frequent involvement in calculations, tends to produce SyntaxError. Lastly, while the Steiner problem rarely encounters errors, it may experience timeout issues when dealing with large parameter sets, resulting in TimeoutError.
\begin{table}[htbp]
\centering
\caption{Error Case Analysis in Common Research Problems}
\label{tab:error_analysis_en}
\begin{tabularx}{1.1\textwidth}{p{2cm}>{\raggedright\arraybackslash\hsize=1.2\hsize}X>{\raggedright\arraybackslash\hsize=0.7\hsize}X}
\toprule
\textbf{Task} & \textbf{Error} & \textbf{Reason} \\
\midrule
Assignment & cost\_matrix[i, j] = c& IndexError: index 3301 is out of bounds for axis 1 with size 3000 \\
1D-Binpacking & with open('data123.txt', 'r') as file:
\quad \quad \quad \quad \quad S = [float(line.strip()) for line in file]& ValueError: could not convert string to float: '[$\ldots$]' \\
\addlinespace
Crew-scheduling & cost[i][j] = c & IndexError: list assignment index out of range  \\
\addlinespace
Steiner & Timeout & TimeoutError: Operation Timed Out   \\
\addlinespace
UBQP & matrix[i][j] = c\_ij & IndexError: list assignment index out of range  \\
\addlinespace
CVRP & routing.AddDimensionWithVehicleCapacity(
    demand\_callback\_index,
    0,
    data,  
    True, 
    'Capacity'
) & TypeError: 'float' object cannot be interpreted as an integer \\
\addlinespace
MDVRP & 
    with open(filename, 'r') as f:   
\quad \quad \quad \quad \quad \quad data = f.read().strip().split('\textbackslash n') 
\quad \quad \quad \quad \quad \quad return [list(map(float, line.strip().split())) for line in data]
& ValueError: could not convert string to float: '[$\ldots$]' \\
\addlinespace
PVRP & routing = pywrapcp.RoutingModel(n\_nodes, num\_vehicles, depot\_index) & TypeError: Wrong number or type of arguments for overloaded function 'new\_RoutingModel'. \\
\addlinespace
Aircraft Landing & model += land\_time[i] $\geq$ land\_time[j] + sep\_matrix[j][i] - M * (before + (1 - same\_rw))) & SyntaxError: unmatched ')' \\
\addlinespace
Generalised Assignment & prob += lpSum(p[i][j] * x[i][j] for i in bins for j in items) & IndexError: list index out of range \\
\addlinespace
Multi-dimensional Knapsack & 
with open('data1.json', 'r') as f:
\quad\quad \quad \quad \quad \quad c = list(map(float, f.read().strip().split()))& ValueError: could not convert string to float: '[$\ldots$]'  \\
\addlinespace
Capacitated Warehouse Location & assert len(F) == 25 and len(d) == 50 and len(K) == 25 and len(c) == 25, 'Data dimensions incorrect' & AssertionError: Data dimensions incorrect \\
\addlinespace
2D-cutting Packing & l = int(float(l\_str)) &ValueError: could not convert string to float: '[$\ldots$]'  \\
\addlinespace
\bottomrule
\end{tabularx}
\end{table}

\clearpage

\subsection{Explanation}

The complete analysis of different problems are given in Appendix \ref{AppendixA}. Here we just show a part of the conclusion with explanation.

\textbf{1. ValueError is the most common error for Assignment problems.}\ref{Assignment}

When trying to assign a sequence (such as a list or another array) to an element of a NumPy array, the dimensions and shapes of the data are incompatible. If the shapes do not align, a ValueError will be raised. And, when reading the data, if preprocessing is not done, a ValueError will also be raised.

\textbf{2. IndexError takes the lead for Crew-sheduling problems.}\ref{Crewscheduling}

For Crew-scheduling problems, we have to do multiple operations on matrices, combined with the presence of both a cost matrix and a time matrix in this problem, have led to an increase in complexity, which in turn has caused a rise in IndexError occurrences, such as "list index out of range".
    
\textbf{3. SyntaxError occurrences arise when LLMs use CoT and PoT techniques while ValueError and AttributeError occurrences arise when LLMs use just PoT technique for steiner problems.}\ref{Steiner}

The SyntaxError occurs due to the incorrect format of an f-string in Python. Models utilizing CoT may generate code that includes a “print” statement at the end, sometimes accompanied by a write operation and sometimes not. It also occurs when the parenthesis or bracket are unclosed or quotation marks are mismatched, which is a common dilemma faced by codes generated by LLMs.

The AttributeError occurs when codes generated by LLMs try to call a method of a class constructed by a certain built-in solver in Python. However, the method is unavailable caused by different version of solvers or just misuse. And, when reading the data, if preprocessing is not done, a ValueError will also raised.

\textbf{4. It is more likely for weaker models to raise a ValueError, while stronger models are more inclined to raise an IndexError for UBQP problems.}\ref{UBQP}

For ValueError, besides reading error, "not enough values to unpack" is usually raised, while for IndexError,  "list assignment index out of range" is usually raised. This is because weaker models fail to take into account the consistency of contextual output while stronger models are not sensitive to numbers. 

\textbf{5. Weaker models still tend to raise a SyntaxError while stronger models tend to raise KeyError and TypeError for VRP problems.}\ref{VRP}

While SyntaxError is the same as that in the third statement, TypeError occurs when codes generated by LLMs try to call a function constructed by a ceartain built-in solver in Python. In detail, Incorrect argument type is given. Besides, KeyError occurs when LLMs tend to access a key of dictionary mapped between the DEPoT number and its' other properties. For example, if the beginning dePoT is 0 in datasets but LLMs regard it as 1 on default(By convention, we count from one.), then when access key "0", it will raise such kind of Error.

\textbf{6. Noticeably, the CoT method or the absence of the CoT method results in an interchange of errortypes between TypeError and SyntaxError for models such as Deepseek-R1 and LLAMA3 for 2D-cutting packing-constrained guillotine problems.}\ref{2D-Binpacking}

Pitifully, we don't know what happened here and this issue necessitates further investigation in subsequent studies. 

\subsection{Timeout}

In this part, Timeout rate is presented in Table.\ref{timeout} to supplement additional information except for PR, AR, MAPE and different types of errors. Usually, beside the built-in solvers, dynamic programming is a regular customer in No-CoT prompting, which may result in the increase in Timeout rate. Table.\ref{timeout} shows that stronger models are more likely to generate codes that are easy to make timeout error while GPT-4o-mini without CoT shows the similar trend compared to that with CoT. But such trend is not significant for DeepSeek-R1. In general, CoT technique is a regular customer with low timeout rate while No-CoT does with higher one, and disordered datasets may lead to higher timeout rate for stronger models but just the opposite for weaker models. Actually, the more the timeout rate goes up, the more solutions could not be obtained within limited time. Therefore, if we want to obtain a feasible solution within limited time, LLMs with CoT on No-disordered datasets is a common approach within limited time.

\begin{table}[ht]
\centering
\caption{Timeout Task Performance Comparison}
\label{timeout}
\resizebox{\textwidth}{!}{ % 使用 \resizebox 来调整表格大小
\begin{tabular}{llcccccccc}
\toprule
\multirow{2}{*}{Task} & \multirow{2}{*}{Dataset} & \multicolumn{2}{c}{GPT-4o-mini} & \multicolumn{2}{c}{LLAMA3-8B} & \multicolumn{2}{c}{ORLM} & \multicolumn{2}{c}{DeepSeek-R1}\\
\cmidrule(lr){3-4} \cmidrule(lr){5-6}\cmidrule(lr){7-8}\cmidrule(lr){9-10}
 & & PoT & PoT+CoT & PoT & PoT+CoT &PoT & PoT+CoT &PoT & PoT+CoT \\
\midrule

\multirow{3}{*}{Assignment} 
& Explicit-origin & \textbackslash{}  & 0.00 & \textbackslash{}   &  \textbackslash{} &  \textbackslash{} &  \textbackslash{} &  \textbackslash{} &  \textbackslash{}\\
& Implicit-origin & 0.00  & 0.00 & 0.00  & 3.33 &0.00&1.67& 1.67 &$\overline{\textbf{11.67}}$ \\
& Implicit-disorder & \textbackslash{}  & $\underline{\textbf{0.93}}$ & 1.85  & 1.85& 1.85  & $\overline{\textbf{2.78}}$& \textbackslash{}  & \textbackslash{}  \\
\cmidrule(lr){1-10}

\multirow{3}{*}{1D-Binpacking}
& Explicit-origin & \textbackslash{}  & 11.55 & \textbackslash{}  & \textbackslash{} & \textbackslash{} & \textbackslash{} & \textbackslash{} & \textbackslash{}  \\
& Implicit-origin & 15.00  & 4.62 & 6.04  & $\underline{\textbf{0.94}}$ & 2.92 &4.17& $\overline{\textbf{16.74}}$ &15.10  \\
& Implicit-disorder & \textbackslash{}  & 0.80 & $\overline{\textbf{1.49}}$  & 0.77 & $\underline{\textbf{0.63}}$  & 1.08& \textbackslash{}  & \textbackslash{} \\
\cmidrule(lr){1-10}

\multirow{3}{*}{Crew-scheduling}
& Explicit-origin & \textbackslash{}  & 12.67 & \textbackslash{}  & \textbackslash{} & \textbackslash{} & \textbackslash{} & \textbackslash{} & \textbackslash{}  \\
& Implicit-origin & $\overline{\textbf{20.67}}$  & 16.00 & 1.33  & 3.33 & $\underline{\textbf{0.67}}$ & 3.33& 9.33 &9.26  \\
& Implicit-disorder & \textbackslash{}  & $\overline{\textbf{24.39}}$ & $\underline{\textbf{0.40}}$  & 0.80& 13.85  & 1.20& \textbackslash{}  & \textbackslash{}  \\
\cmidrule(lr){1-10}

\multirow{3}{*}{Steiner}
& Explicit-origin & \textbackslash{}  & 10.15 & \textbackslash{}  & \textbackslash{} & \textbackslash{} & \textbackslash{} & \textbackslash{} & \textbackslash{}  \\
& Implicit-origin & $\overline{\textbf{10.41}}$  & 5.92 & 9.96  & 7.76&2.35&0.92& $\underline{\textbf{0.78}}$ &1.56  \\
& Implicit-disorder & \textbackslash{}  & $\overline{\textbf{22.79}}$ & $\underline{\textbf{1.63}}$  & 6.23& 1.33  & 2.76& \textbackslash{}  & \textbackslash{}  \\
\cmidrule(lr){1-10}

\multirow{3}{*}{UBQP}
& Explicit-origin & \textbackslash{}  & 35.56 & \textbackslash{}  & \textbackslash{} & \textbackslash{} & \textbackslash{} & \textbackslash{} & \textbackslash{}  \\
& Implicit-origin & $\overline{\textbf{64.44}}$  & 21.64 & 2.22  & $\underline{\textbf{0.74}}$ & $\underline{\textbf{0.74}}$ & 2.96& 31.85 &4.44  \\
& Implicit-disorder & \textbackslash{}  & $\overline{\textbf{21.26}}$ & 0.00  & 0.00& 0.00  & 1.78& \textbackslash{}  & \textbackslash{}  \\
\cmidrule(lr){1-10}

\multirow{3}{*}{CVRP}
& Explicit-origin & \textbackslash{}  & 1.33 & \textbackslash{}  & \textbackslash{} & \textbackslash{} & \textbackslash{} & \textbackslash{} & \textbackslash{}  \\
& Implicit-origin & 20.78  & 6.37 & 0.16  & 0.00 & 0.98 & 0.16& $\overline{\textbf{24.82}}$ &4.57  \\
& Implicit-disorder & \textbackslash{}  & $\overline{\textbf{8.33}}$ & 1.27  & $\underline{\textbf{0.20}}$& 0.78  & 1.08& \textbackslash{}  & \textbackslash{}  \\
\cmidrule(lr){1-10}
\multirow{3}{*}{MDVRP}
& Explicit-origin & \textbackslash{}  & 3.03 & \textbackslash{}  & \textbackslash{} & \textbackslash{} & \textbackslash{} & \textbackslash{} & \textbackslash{}  \\
& Implicit-origin & $\overline{\textbf{4.55}}$ & 3.03 & 0.76  & 0.76 & 0.00 & 0.00& 0.00 &1.52  \\
& Implicit-disorder & \textbackslash{}  & $\overline{\textbf{3.54}}$ & 1.01  & 0.51& 0.51  & 0.00& \textbackslash{}  & \textbackslash{}  \\
\cmidrule(lr){1-10}
\multirow{3}{*}{PVRP}
& Explicit-origin & \textbackslash{}  & 11.90 & \textbackslash{}  & \textbackslash{} & \textbackslash{} & \textbackslash{} & \textbackslash{} & \textbackslash{}  \\
& Implicit-origin & 11.90  & 11.11 & 0.79  & 0.00 & 0.79 & 0.00& 8.73 &$\overline{\textbf{14.75}}$  \\
& Implicit-disorder & \textbackslash{}  & $\overline{\textbf{12.86}}$ & 1.43  & 0.00& 0.48  & 0.48& \textbackslash{}  & \textbackslash{}  \\
\cmidrule(lr){1-10}
\multirow{3}{*}{Aircraft Landing}
& Explicit-origin & \textbackslash{}  & 0.00 & \textbackslash{}  & \textbackslash{} & \textbackslash{} & \textbackslash{} & \textbackslash{} & \textbackslash{}  \\
& Implicit-origin & 9.09  & 0.00 & 0.00  &0.00&0.00&0.00& $\overline{\textbf{18.18}}$ &9.09  \\
& Implicit-disorder & \textbackslash{}  & $\overline{\textbf{7.95}}$ & 0.00  & 1.14& 0.00  & 0.00& \textbackslash{}  & \textbackslash{}  \\
\cmidrule(lr){1-10}
\multirow{3}{*}{Generalised Assignment}
& Explicit-origin & \textbackslash{}  & 0.00 & \textbackslash{}  & \textbackslash{} & \textbackslash{} & \textbackslash{} & \textbackslash{} & \textbackslash{}  \\
& Implicit-origin & 1.11  & $\overline{\textbf{2.78}}$ & 0.00  & 0.00 &0.56&0.56& 0.00 &0.00  \\
& Implicit-disorder & \textbackslash{}  & 1.33 & 0.00  & 0.00& $\overline{2.00}$  & 0.00& \textbackslash{}  & \textbackslash{}  \\

\cmidrule(lr){1-10}
\multirow{3}{*}{Multi-dimensional Knapsack}
& Explicit-origin & \textbackslash{}  & 5.63 & \textbackslash{}  & \textbackslash{} & \textbackslash{} & \textbackslash{} & \textbackslash{} & \textbackslash{}  \\
& Implicit-origin & 8.66  & 15.59 & 0.99  & 0.90 & 0.85 & $\underline{\textbf{0.72}}$& 38.36 &$\overline{\textbf{57.71}}$  \\
& Implicit-disorder & \textbackslash{}  & $\overline{\textbf{19.95}}$ & 0.35  & $\underline{\textbf{0.15}}$& 0.37  & 0.20& \textbackslash{}  & \textbackslash{}  \\

\cmidrule(lr){1-10}
\multirow{3}{*}{Capacitated Warehouse Location}
& Explicit-origin & \textbackslash{}  & 2.94 & \textbackslash{}  & \textbackslash{} & \textbackslash{} & \textbackslash{} & \textbackslash{} & \textbackslash{}  \\
& Implicit-origin & 0.00  & 0.00 & 0.00  & 0.00 & 0.00 & 0.00& 0.00 &$\overline{\textbf{0.98}}$  \\
& Implicit-disorder & \textbackslash{}  & $\overline{\textbf{3.07}}$ & 0.00  & 0.00& 0.59  & 0.00& \textbackslash{}  & \textbackslash{}  \\
\cmidrule(lr){1-10}

\multirow{3}{*}{2D-cutting Packing}
& Explicit-origin & \textbackslash{}  & 5.56 & \textbackslash{}  & \textbackslash{} & \textbackslash{} & \textbackslash{} & \textbackslash{} & \textbackslash{}  \\
& Implicit-origin & 0.00  & 5.56 & 0.00  & 0.00 & 0.00 & 0.00& 11.11 &$\overline{\textbf{38.89}}$  \\
& Implicit-disorder & \textbackslash{}  & $\overline{\textbf{6.90}}$ & 0.00  & 0.00& 0.00  & 0.00& \textbackslash{}  & \textbackslash{}  \\

% 其余任务按相同模式添加...
\bottomrule
\end{tabular}
}
\end{table}

\subsection{Case Study: Potential for Solution Improvement}

\subsubsection{Framework Overview}

For each instance, the framework runs:
\begin{enumerate}[leftmargin=*]
    \item \textbf{Input normalization:} convert the raw instance into a clean, stable representation
    \item \textbf{Iterative search (rounds $\times$ simulations):}
    maintain a search graph whose nodes store candidate solutions. Each simulation:
    (i) selects a promising node, (ii) constructs an augmented prompt that includes the problem, current solution, formatting constraints, and guidance modules, (iii) queries the LLM to generate a new candidate, and (iv) parses the candidate into structured JSON.
    \item \textbf{Verification \& scoring:} check feasibility against hard constraints; compute the objective value; update the global best if improved.
    \item \textbf{Cross-round guidance aggregation:} cluster/aggregate traces into ``best'' vs ``worst'' patterns and summarize them into short guidance for the next round, to discourage repeated failure modes and reinforce successful heuristics.
\end{enumerate}

Core search: Enhanced Graph-of-Thought (MCTS-like).

The search loop is a pragmatic MCTS variant:
\begin{itemize}[leftmargin=*]
    \item \textbf{Selection:} choose promising leaves using scores (objective-based) with stochasticity (temperature-like sampling).
    \item \textbf{Expansion:} generate new candidates with the LLM under strict output-format requirements.
    \item \textbf{Evaluation:} parse $\rightarrow$ feasibility check $\rightarrow$ objective compute.
    \item \textbf{Backprop-like update:} update node scores / best-so-far / history, which then influences later selections.
\end{itemize}

\subsubsection{}

\section{Conclusion}\label{conclusion}
In the part of conclusion, four questions put forward at the beginning of the article will be answered.

\textbf{RQ1:} How are the datasets constructed? What’s unique in these datasets?

The datasets used in this study with data collected from OR Library and generated in the form of natural language from several paper, include the original datasets, expanded datasets and those processed through rearrangement (disorder). The original datasets are composed of natural language, and these datasets can be used for fine-tuning to assist in training OR-specific LLMs. Additionally, the expanded datasets are formed by extracting contextual information from various types of datasets, creating domain-specific datasets that can enhance the context in which events may occur, thereby improving the robustness of OR-specific LLMs. However, we have not yet testified this aspect, which is a task we plan to complete in the future. Finally, The disordered datasets help us assess the LLMs’ understanding of various problems, allowing users to quickly gauge the model’s comprehension abilities even when they know nothing about specific types of questions. Based on the test results presented in this paper, users can adopt corresponding techniques to enhance model performance.

\textbf{RQ2:}  Can prompt engineering help in improving the solution or lowering the errors? Is LLMs sensitive to input text?

Theoretically, while CoT provides guidance, the performance of CoT should be higher than that of that without CoT. However, due to the potential differences in the solvers used, the relationship between the two in terms of AR and MAPE is difficult to ascertain. In addition, CoT just benefits stronger models and worsens weaker models in PR. In our work, the result shows that the PR of disordered datasets improves compared to that of original ones for GPT-4o-mini, even though the disordered datasets disrupted the structure of the problems. This trend is observed in the AR metric but disobeyed in MAPE metric.

This research finding indicates that prompt engineering is helpful especially to stronger models in general, but backfire for certain models, such as DeepSeek-R1. By comparison, it's not friendly to weaker models. Notably, the model exhibits slight performance variations across different datasets. In detail, specific types of problems,
such as generalized assignment problems and capacitated warehouse location problems, achieve significant perfor-
mance improvements with the datasets in disorder, indicating that the expression of the task should be considered
when asking LLMs to deal with discrete optimization problems according to the type of problems. Other detailed results have been included in the Appendix \ref{statistic} for the convenience of reference. Besides, LLMs exhibit high sensitivity to input text, which is validated by comparing disorder datasets with original datasets. The result shows that the disordered datasets lead to high variance of the model for stronger models. Thus, such method is radical along with the risk of instability. But for weaker models, disordered datasets do not perform well. 

Also, we can say that different models have their own advantages over different problems. For instance, even ORLM, a weak model can be slightly inferior to DeepSeek-R1 in crew-scheduling problems.  But how to use effective prompt engineering and when to use disordered datasets? \textbf{RQ3} gives the answer.

\textbf{RQ3:} What plays a significant role in the performance of LLMs on discrete optimization problems?

To make the performance of these problems clear in the disordered datasets as shown in this table, these problems are divided into four types according to the difficulties in understanding them. As is given by LLMs, Steiner Tree, UBQP, 2D-Cutting Packing-Constrained Guillotine, MDVRP and PVRP are more difficult to understand; CVRP, Generalised Assignment problems and Aircraft Landing Problems are moderately difficult to understand; crew scheduling, multi-dimensional knapsack, capacitated warehouse location, assignment problems and 1D-Binpacking are easier to understand. Surprisingly, a significant consistency between the above classifications and Table \ref{tab:problem_changes} is observed. This suggests that the degree of difference in performance on datasets in disorder compared to the original dataset can be used to quantify LLMs’ understanding difficulty of this type of problem. Therefore, for an unknown problem, we can first predict the performance difference between disordered datasets and original ones. Based on the  definition methods presented in this paper, for problems that are easier to understand, we can enhance LLMs’ problem-solving capabilities using a method similar to that of disorder. Conversely, for problems that are more difficult to understand, the Chain of Thought (CoT) approach can significantly improve LLMs’ problem-solving abilities. A notable example is that by analyzing Table \ref{PR}.\ref{AR}.\ref{MAPE}, it shows that difficult-to-understand problems have improved their problem-solving capabilities through CoT, while for easier problems, CoT may sometimes negatively impact problem-solving performance. Thus, when LLMs try to deal with discrete optimization problems, their difficulties in understanding the problems will have a huge impact on the performance. 

\textbf{RQ4:} What strategy should I adopt?

There are three strategies:

Firstly, weak models are recommended to use no-CoT technique with original datasets while stronger models are recommended to use CoT technique, then with disordered datasets. 

Secondly, the guarantee of quality of solution under any condition requires CoT while the pursuit of optimal solution requires disordered datasets. 

Thirdly, if the difficulty of the problems has been figured out and ranked based on Table \ref{tab:problem_changes}, disordered datasets are recommended for difficult-to-understand problems with instability of that on easier problems. If none of these strategies takes effect, you need to consider alternative methods beyond prompt optimization techniques (non-PoT approaches). However, this extension lies outside the scope of the current study and will be explored in our future research directions.

\begin{table}[ht]
    \centering
    \caption{Performance Effectiveness of Different Problems}
    \footnotesize
    \begin{tabular}{@{\extracolsep{\fill}} l p{5cm} p{5cm} @{}}
        \toprule
        \textbf{Status} & \textbf{Status Descriptions}& \textbf{Problems} \\ 
        \midrule
        Worsened & Disordered datasets are slightly inferior &1D-Binpacking, Crew-scheduling, CVRP \\ \midrule
        Significantly Worsened & Disordered datasets are greatly inferior &Steiner, UBQP \\ \midrule
        Improved & Disordered datasets are slightly superior &Assignment, MDVRP, Aircraft Landing, Multi-dimensional Knapsack \\ \midrule
        Significantly Improved & Disordered datasets are greatly superior  &PVRP, Generalised Assignment, Capacitated Warehouse Location, 2D-Cutting Packing-Constrained Guillotine \\ 
        \bottomrule
    \end{tabular}
    \label{tab:problem_changes}
\end{table}

\clearpage
\paragraph{Data Availability}
The datasets generated and/or analysed during the current study are available in the Figshare repository at \url{https://doi.org/10.6084/m9.figshare.29376404.v1}.

\bibliographystyle{plainnat}
% 参考文献样式可根据不同刊物要求进行更改
% 如IEEEtran, plain, unsrt， alpha, acm等。

\bibliography{cas-refs}

\newpage

\Huge \textbf{Appendix}

\appendix
\normalsize
\renewcommand{\thesection}{\Alph{section}} % 将子章节编号格式更改为字母
\renewcommand{\thesubsection}{\Alph{section}\arabic{subsection}} % 将子章节编号设置为 1.1, 1.2 的格式

\setcounter{section}{0} % 重置附录的节计数器

\section{Different Problems}\label{AppendixA}

\subsection{Assignment}\label{Assignment}

\begin{table}[ht]
\centering
\caption{Assignment Problems Performance Metrics}
\label{tab2}
\resizebox{0.9\textwidth}{!}{
\begin{tabular}{llcccccccc}
\toprule
\multirow{2}{*}{Error Type} & \multirow{2}{*}{Dataset} & \multicolumn{2}{c}{GPT-4o-mini} & \multicolumn{2}{c}{LLAMA3-8B} & \multicolumn{2}{c}{ORLM} & \multicolumn{2}{c}{DeepSeek-R1}\\
\cmidrule(lr){3-4} \cmidrule(lr){5-6}\cmidrule(lr){7-8}\cmidrule(lr){9-10}
 & & PoT & PoT+CoT & PoT & PoT+CoT &PoT & PoT+CoT &PoT & PoT+CoT \\
\midrule

\multirow{3}{*}{IndexError} 
& Explicit-origin & \textbackslash{}  & 0.00 & \textbackslash{}   &  \textbackslash{} &  \textbackslash{} &  \textbackslash{} &  \textbackslash{} &  \textbackslash{}\\
& Implicit-origin & \textbf{24.24}  & 19.57 & 2.22  & 9.80 & 1.03 & 3.32 & 0.00 & 9.09 \\
& Implicit-disorder & \textbackslash{}  & 0.00 & \textbf{22.22}  & 3.70 & 2.69  & 2.85 & \textbackslash{}  & \textbackslash{}  \\
\cmidrule(lr){1-10}

\multirow{3}{*}{FileNotFoundError}
& Explicit-origin & \textbackslash{}  & \textbf{9.52} & \textbackslash{}  & \textbackslash{} & \textbackslash{} & \textbackslash{} & \textbackslash{} & \textbackslash{}  \\
& Implicit-origin & \textbf{3.03}  & 2.17 & 0.00  & 0.00 & 0.69 & 1.85 & 0.00 & 0.00  \\
& Implicit-disorder & \textbackslash{}  & 0.00 & \textbf{3.03}  & 0.00 & 2.31  & 1.63 & \textbackslash{}  & \textbackslash{} \\
\cmidrule(lr){1-10}

\multirow{3}{*}{OverflowError}
& Explicit-origin & \textbackslash{}  & 0.00 & \textbackslash{}  & \textbackslash{} & \textbackslash{} & \textbackslash{} & \textbackslash{} & \textbackslash{}  \\
& Implicit-origin & 0.00  & 0.00 & 0.00  & \textbf{1.96} & 0.00 & 0.00 & 0.00 & 0.00  \\
& Implicit-disorder & \textbackslash{}  & 0.00 & \textbf{1.01}  & 0.00 & 0.00  & 0.20 & \textbackslash{}  & \textbackslash{}  \\
\cmidrule(lr){1-10}

\multirow{3}{*}{NameError}
& Explicit-origin & \textbackslash{}  & 0.00 & \textbackslash{}  & \textbackslash{} & \textbackslash{} & \textbackslash{} & \textbackslash{} & \textbackslash{}  \\
& Implicit-origin & 0.00  & 0.00 & \textbf{4.44}  & 0.00 & 3.10 & 3.32 & 0.00 & 0.00  \\
& Implicit-disorder & \textbackslash{}  & 0.00 & \textbf{5.05}  & 3.70 & 3.85  & 3.25 & \textbackslash{}  & \textbackslash{}  \\
\cmidrule(lr){1-10}

\multirow{3}{*}{ValueError}
& Explicit-origin & \textbackslash{}  & $\textbf{80.95}$ & \textbackslash{}  & \textbackslash{} & \textbackslash{} & \textbackslash{} & \textbackslash{} & \textbackslash{}  \\
& Implicit-origin & 57.58  & $\textbf{67.39}$ & 53.33  & 35.29 & 5.52 & 8.86 & 50.00 & 50.00  \\
& Implicit-disorder & \textbackslash{}  & 28.89 & 53.54  & $\textbf{55.56}$ & 5.96  & 8.33 & \textbackslash{}  & \textbackslash{}  \\
\cmidrule(lr){1-10}

\multirow{3}{*}{AttributeError}
& Explicit-origin & \textbackslash{}  & 0.00 & \textbackslash{}  & \textbackslash{} & \textbackslash{} & \textbackslash{} & \textbackslash{} & \textbackslash{}  \\
& Implicit-origin & 0.00  & 0.00 & \textbf{4.44}  & 0.00 & 3.10 & 4.06 & 0.00 & 0.00  \\
& Implicit-disorder & \textbackslash{}  & 0.00 & 1.01  & 1.23 & \textbf{4.62}  & 3.05 & \textbackslash{}  & \textbackslash{}  \\
\cmidrule(lr){1-10}

\multirow{3}{*}{KeyError}
& Explicit-origin & \textbackslash{}  & 4.76 & \textbackslash{}  & \textbackslash{} & \textbackslash{} & \textbackslash{} & \textbackslash{} & \textbackslash{}  \\
& Implicit-origin & 0.00  & 0.00 & 0.00  & 0.00 & \textbf{0.69} & 0.37 & 0.00 & 0.00  \\
& Implicit-disorder & \textbackslash{}  & 0.00 & 0.00  & 0.00 & \textbf{0.38}  & 0.00 & \textbackslash{}  & \textbackslash{}  \\
\cmidrule(lr){1-10}

\multirow{3}{*}{SyntaxError}
& Explicit-origin & \textbackslash{}  & 4.76 & \textbackslash{}  & \textbackslash{} & \textbackslash{} & \textbackslash{} & \textbackslash{} & \textbackslash{}  \\
& Implicit-origin & \textbf{12.12}  & 10.87 & 4.44  & 3.92 & 7.24 & 9.59 & 0.00 & 0.00  \\
& Implicit-disorder & \textbackslash{}  & 0.00 & 3.03  & 3.70 & \textbf{7.88}  & 6.71 & \textbackslash{}  & \textbackslash{}  \\
\cmidrule(lr){1-10}

\multirow{3}{*}{TypeError}
& Explicit-origin & \textbackslash{}  & 0.00 & \textbackslash{}  & \textbackslash{} & \textbackslash{} & \textbackslash{} & \textbackslash{} & \textbackslash{}  \\
& Implicit-origin & 3.03  & 0.00 & \textbf{8.89}  & 5.88 & 1.03 & 4.06 & 0.00 & 0.00  \\
& Implicit-disorder & \textbackslash{}  & 2.22 & 4.04  & $\textbf{12.35}$ & 1.92  & 3.05 & \textbackslash{}  & \textbackslash{}  \\
\cmidrule(lr){1-10}

\multirow{3}{*}{ImportError}
& Explicit-origin & \textbackslash{}  & 0.00 & \textbackslash{}  & \textbackslash{} & \textbackslash{} & \textbackslash{} & \textbackslash{} & \textbackslash{}  \\
& Implicit-origin & 0.00  & 0.00 & 0.00  & 0.00 & 0.34 & \textbf{0.74} & 0.00 & 0.00  \\
& Implicit-disorder & \textbackslash{}  & 0.00 & 0.00  & \textbf{1.23} & 0.58  & 0.41 & \textbackslash{}  & \textbackslash{}  \\
\cmidrule(lr){1-10}

\multirow{3}{*}{PulpError}
& Explicit-origin & \textbackslash{}  & 0.00 & \textbackslash{}  & \textbackslash{} & \textbackslash{} & \textbackslash{} & \textbackslash{} & \textbackslash{}  \\
& Implicit-origin & 0.00  & 0.00 & 0.00  & 0.00 & 0.00 & 0.00 & 0.00 & 0.00  \\
& Implicit-disorder & \textbackslash{}  & 0.00 & 0.00  & 0.00 & 0.00  & 0.00 & \textbackslash{}  & \textbackslash{}  \\

\bottomrule
\end{tabular}
}
\end{table}

Table \ref{tab2} gives the rate of each Error type, which shows that ValueError is the most common error and GPT-4o-mini will make more IndexError and SyntaxError than LLAMA3-8B. Besides, the rate of these errors decline with the help of CoT. From the datasets, the error rate goes down significantly but the trend still maintains the same.

\subsection{1D-Binpacking}

\begin{table}[ht]
\centering
\caption{1D-Binpacking Problems Performance Metrics}
\label{tab4}
\resizebox{0.85\textwidth}{!}{
\begin{tabular}{llcccccccc}
\toprule
\multirow{2}{*}{Error Type} & \multirow{2}{*}{Dataset} & \multicolumn{2}{c}{GPT-4o-mini} & \multicolumn{2}{c}{LLAMA3-8B} & \multicolumn{2}{c}{ORLM} & \multicolumn{2}{c}{DeepSeek-R1}\\
\cmidrule(lr){3-4} \cmidrule(lr){5-6}\cmidrule(lr){7-8}\cmidrule(lr){9-10}
 & & PoT & PoT+CoT & PoT & PoT+CoT &PoT & PoT+CoT &PoT & PoT+CoT \\
\midrule

\multirow{3}{*}{IndexError} 
& Explicit-origin & \textbackslash{}  & 0.00 & \textbackslash{}   &  \textbackslash{} &  \textbackslash{} &  \textbackslash{} &  \textbackslash{} &  \textbackslash{}\\
& Implicit-origin & 4.23  & 3.23 & 7.19  & 1.24 & 17.26 & 2.98 & 15.94 & $\textbf{29.23}$ \\
& Implicit-disorder & \textbackslash{}  & 8.47 & 1.70  & 1.40 & \textbf{14.50}  & 8.02 & \textbackslash{}  & \textbackslash{}  \\
\cmidrule(lr){1-10}

\multirow{3}{*}{FileNotFoundError}
& Explicit-origin & \textbackslash{}  & 0.00 & \textbackslash{}  & \textbackslash{} & \textbackslash{} & \textbackslash{} & \textbackslash{} & \textbackslash{}  \\
& Implicit-origin & 0.00  & 0.00 & 5.12  & 0.00 & \textbf{5.58} & 1.75 & 0.00 & 2.31  \\
& Implicit-disorder & \textbackslash{}  & 0.00 & 10.21  & 3.85 & 4.46  & $\textbf{31.28}$ & \textbackslash{}  & \textbackslash{} \\
\cmidrule(lr){1-10}

\multirow{3}{*}{OverflowError}
& Explicit-origin & \textbackslash{}  & 0.00 & \textbackslash{}  & \textbackslash{} & \textbackslash{} & \textbackslash{} & \textbackslash{} & \textbackslash{}  \\
& Implicit-origin & 0.00  & 0.00 & \textbf{0.12}  & 0.08 & 0.00 & 0.07 & 0.00 & 0.00  \\
& Implicit-disorder & \textbackslash{}  & 0.00 & 0.00  & 0.00 & 0.00  & 0.00 & \textbackslash{}  & \textbackslash{}  \\
\cmidrule(lr){1-10}

\multirow{3}{*}{NameError}
& Explicit-origin & \textbackslash{}  & 0.00 & \textbackslash{}  & \textbackslash{} & \textbackslash{} & \textbackslash{} & \textbackslash{} & \textbackslash{}  \\
& Implicit-origin & 0.39  & 0.00 & \textbf{11.81}  & 1.17 & 10.66 & 3.13 & 0.00 & 0.77  \\
& Implicit-disorder & \textbackslash{}  & 0.13 & 2.55  & 0.87 & $\textbf{12.27}$  & 9.63 & \textbackslash{}  & \textbackslash{}  \\
\cmidrule(lr){1-10}

\multirow{3}{*}{ValueError}
& Explicit-origin & \textbackslash{}  & $\textbf{80.95}$ & \textbackslash{}  & \textbackslash{} & \textbackslash{} & \textbackslash{} & \textbackslash{} & \textbackslash{}  \\
& Implicit-origin & 38.58  & $\textbf{93.55}$ & 59.44  & 56.45 & 28.43 & 4.66 & 28.87 & 22.31  \\
& Implicit-disorder & \textbackslash{}  & $\textbf{91.27}$ & 70.64  & 68.85 & 14.87  & 8.29 & \textbackslash{}  & \textbackslash{}  \\
\cmidrule(lr){1-10}

\multirow{3}{*}{AttributeError}
& Explicit-origin & \textbackslash{}  & \textbf{9.52} & \textbackslash{}  & \textbackslash{} & \textbackslash{} & \textbackslash{} & \textbackslash{} & \textbackslash{}  \\
& Implicit-origin & 0.16  & 0.00 & 2.07  & 0.86 & \textbf{12.69} & 3.13 & 0.23 & 0.77  \\
& Implicit-disorder & \textbackslash{}  & 0.00 & 2.13  & 1.40 & $\textbf{14.13}$  & 9.36 & \textbackslash{}  & \textbackslash{}  \\
\cmidrule(lr){1-10}

\multirow{3}{*}{KeyError}
& Explicit-origin & \textbackslash{}  & 0.00 & \textbackslash{}  & \textbackslash{} & \textbackslash{} & \textbackslash{} & \textbackslash{} & \textbackslash{}  \\
& Implicit-origin & 0.08  & 0.00 & \textbf{2.19}  & 0.47 & 0.00 & 0.15 & 0.69 & 0.00  \\
& Implicit-disorder & \textbackslash{}  & 0.00 & \textbf{1.70}  & 0.44 & 1.12  & 0.00 & \textbackslash{}  & \textbackslash{}  \\
\cmidrule(lr){1-10}

\multirow{3}{*}{SyntaxError}
& Explicit-origin & \textbackslash{}  & 0.00 & \textbackslash{}  & \textbackslash{} & \textbackslash{} & \textbackslash{} & \textbackslash{} & \textbackslash{}  \\
& Implicit-origin & 0.16  & 0.00 & 4.51  & 2.18 & \textbf{17.77} & 6.11 & 0.46 & 0.00  \\
& Implicit-disorder & \textbackslash{}  & 0.00 & 5.53  & 1.75 & $\textbf{30.48}$  & 27.54 & \textbackslash{}  & \textbackslash{}  \\
\cmidrule(lr){1-10}

\multirow{3}{*}{TypeError}
& Explicit-origin & \textbackslash{}  & \textbf{4.76} & \textbackslash{}  & \textbackslash{} & \textbackslash{} & \textbackslash{} & \textbackslash{} & \textbackslash{}  \\
& Implicit-origin & 1.49  & 3.23 & \textbf{7.31}  & 1.32 & 7.11 & 2.40 & 0.00 & 0.77  \\
& Implicit-disorder & \textbackslash{}  & 0.13 & 4.26  & 1.92 & \textbf{7.06}  & 4.01 & \textbackslash{}  & \textbackslash{}  \\
\cmidrule(lr){1-10}

\multirow{3}{*}{ImportError}
& Explicit-origin & \textbackslash{}  & 0.00 & \textbackslash{}  & \textbackslash{} & \textbackslash{} & \textbackslash{} & \textbackslash{} & \textbackslash{}  \\
& Implicit-origin & 0.39  & 0.00 & 0.24  & 0.16 & 0.51 & 0.44 & 0.00 & \textbf{0.77}  \\
& Implicit-disorder & \textbackslash{}  & 0.00 & 1.28  & 0.17 & 1.12  & \textbf{1.87} & \textbackslash{}  & \textbackslash{}  \\
\cmidrule(lr){1-10}

\multirow{3}{*}{PulpError}
& Explicit-origin & \textbackslash{}  & 0.00 & \textbackslash{}  & \textbackslash{} & \textbackslash{} & \textbackslash{} & \textbackslash{} & \textbackslash{}  \\
& Implicit-origin & 0.00  & 0.00 & 0.00  & 0.00 & 0.00 & 0.00 & 0.00 & 0.00  \\
& Implicit-disorder & \textbackslash{}  & 0.00 & 0.00  & 0.00 & 0.00  & 0.00 & \textbackslash{}  & \textbackslash{}  \\

\bottomrule
\end{tabular}
}
\end{table}

For the 1D-Binpacking problems, Table \ref{MAPE} shows that the MAPE is comparably low, indicating LLMs' ability to deal with 1D-Binpacking problems. From AR(\ref{AR}) and MAPE(\ref{MAPE}), some solutions degraded from the optimal solution to suboptimal ones, which might explain the simultaneous decrease in both MAPE and AR. The result is aligned with the instability of LLAMA3-8B and the low PR indicates that LLAMA3-8B is not suitable for dealing with problems that need iterative calculations or problems that involve conditional statements such as OR problems. 

From Table \ref{tab4}, ValueError still takes the lead similarly. The conclusion is close to that in the assignment problems. 

\subsection{Crew-scheduling}\label{Crewscheduling}

\begin{table}[ht]
\centering
\caption{Crew-scheduling Problems Performance Metrics}
\label{tab6}
\resizebox{0.85\textwidth}{!}{
\begin{tabular}{llcccccccc}
\toprule
\multirow{2}{*}{Error Type} & \multirow{2}{*}{Dataset} & \multicolumn{2}{c}{GPT-4o-mini} & \multicolumn{2}{c}{LLAMA3-8B} & \multicolumn{2}{c}{ORLM} & \multicolumn{2}{c}{DeepSeek-R1}\\
\cmidrule(lr){3-4} \cmidrule(lr){5-6}\cmidrule(lr){7-8}\cmidrule(lr){9-10}
 & & PoT & PoT+CoT & PoT & PoT+CoT &PoT & PoT+CoT &PoT & PoT+CoT \\
\midrule

\multirow{3}{*}{IndexError} 
& Explicit-origin & \textbackslash{}  & $\textbf{30.77}$ & \textbackslash{}   &  \textbackslash{} &  \textbackslash{} &  \textbackslash{} &  \textbackslash{} &  \textbackslash{}\\
& Implicit-origin & 46.40  & 48.06 & 19.61  & 30.47 & 11.76 & 7.05 & \textbf{59.78} & 32.89 \\
& Implicit-disorder & \textbackslash{}  & 33.67 & 28.46  & 22.89 & $\textbf{34.78}$  & 2.62 & \textbackslash{}  & \textbackslash{}  \\
\cmidrule(lr){1-10}

\multirow{3}{*}{FileNotFoundError}
& Explicit-origin & \textbackslash{}  & \textbf{14.62} & \textbackslash{}  & \textbackslash{} & \textbackslash{} & \textbackslash{} & \textbackslash{} & \textbackslash{}  \\
& Implicit-origin & 3.60  & \textbf{6.20} & 0.00  & 0.78 & 0.00 & 2.12 & 3.26 & 0.67  \\
& Implicit-disorder & \textbackslash{}  & 0.00 & 0.00  & 0.00 & 2.17  & \textbf{4.84} & \textbackslash{}  & \textbackslash{} \\
\cmidrule(lr){1-10}

\multirow{3}{*}{OverflowError}
& Explicit-origin & \textbackslash{}  & \textbf{2.31} & \textbackslash{}  & \textbackslash{} & \textbackslash{} & \textbackslash{} & \textbackslash{} & \textbackslash{}  \\
& Implicit-origin & 0.40  & 0.00 & 0.00  & \textbf{0.78} & 0.00 & 0.14 & 0.00 & 0.67  \\
& Implicit-disorder & \textbackslash{}  & 0.80 & 0.00  & 0.00 & \textbf{3.62}  & 0.00 & \textbackslash{}  & \textbackslash{}  \\
\cmidrule(lr){1-10}

\multirow{3}{*}{NameError}
& Explicit-origin & \textbackslash{}  & \textbf{7.69} & \textbackslash{}  & \textbackslash{} & \textbackslash{} & \textbackslash{} & \textbackslash{} & \textbackslash{}  \\
& Implicit-origin & 0.80  & 2.71 & 5.23  & 5.47 & $\textbf{17.65}$ & 7.90 & 0.00 & 0.67  \\
& Implicit-disorder & \textbackslash{}  & 0.80 & 4.88  & 3.21 & 0.72  & \textbf{5.00} & \textbackslash{}  & \textbackslash{}  \\
\cmidrule(lr){1-10}

\multirow{3}{*}{ValueError}
& Explicit-origin & \textbackslash{}  & 26.92 & \textbackslash{}  & \textbackslash{} & \textbackslash{} & \textbackslash{} & \textbackslash{} & \textbackslash{}  \\
& Implicit-origin & 35.20 & 31.78 & 25.49  & \textbf{38.28} & 11.76 & 9.31 & 1.09 & 0.00  \\
& Implicit-disorder & \textbackslash{}  & \textbf{35.46} & 32.52  & 35.34 & 0.72  & 3.44 & \textbackslash{}  & \textbackslash{}  \\
\cmidrule(lr){1-10}

\multirow{3}{*}{AttributeError}
& Explicit-origin & \textbackslash{}  & \textbf{3.85} & \textbackslash{}  & \textbackslash{} & \textbackslash{} & \textbackslash{} & \textbackslash{} & \textbackslash{}  \\
& Implicit-origin & 0.40  & 1.94 & 0.65  & 1.56 & \textbf{11.76} & 4.80 & 2.17 & 4.03  \\
& Implicit-disorder & \textbackslash{}  & 0.60 & \textbf{3.25}  & 0.00 & 0.72  & 2.95 & \textbackslash{}  & \textbackslash{}  \\
\cmidrule(lr){1-10}

\multirow{3}{*}{KeyError}
& Explicit-origin & \textbackslash{}  & \textbf{4.62} & \textbackslash{}  & \textbackslash{} & \textbackslash{} & \textbackslash{} & \textbackslash{} & \textbackslash{}  \\
& Implicit-origin & 0.00  & 0.39 & \textbf{3.92}  & 0.78 & 0.00 & 0.71 & 1.09 & 2.01  \\
& Implicit-disorder & \textbackslash{}  & 0.20 & \textbf{1.63}  & 1.61 & 0.72  & 0.25 & \textbackslash{}  & \textbackslash{}  \\
\cmidrule(lr){1-10}

\multirow{3}{*}{SyntaxError}
& Explicit-origin & \textbackslash{}  & \textbf{2.31} & \textbackslash{}  & \textbackslash{} & \textbackslash{} & \textbackslash{} & \textbackslash{} & \textbackslash{}  \\
& Implicit-origin & 4.40  & 6.20 & 9.80  & 16.41 & $\textbf{47.06}$ & 10.72 & 5.43 & 4.03  \\
& Implicit-disorder & \textbackslash{}  & 0.20 & \textbf{12.60}  & 12.05 & 2.90  & 9.34 & \textbackslash{}  & \textbackslash{}  \\
\cmidrule(lr){1-10}

\multirow{3}{*}{TypeError}
& Explicit-origin & \textbackslash{}  & \textbf{5.38} & \textbackslash{}  & \textbackslash{} & \textbackslash{} & \textbackslash{} & \textbackslash{} & \textbackslash{}  \\
& Implicit-origin & 8.00  & 2.33 & \textbf{11.76}  & 4.69 & 0.00 & 3.67 & 6.52 & 9.40  \\
& Implicit-disorder & \textbackslash{}  & 1.39 & 6.50  & \textbf{10.84} & 7.25  & 2.79 & \textbackslash{}  & \textbackslash{}  \\
\cmidrule(lr){1-10}

\multirow{3}{*}{ImportError}
& Explicit-origin & \textbackslash{}  & \textbf{0.77} & \textbackslash{}  & \textbackslash{} & \textbackslash{} & \textbackslash{} & \textbackslash{} & \textbackslash{}  \\
& Implicit-origin & 0.40  & 0.00 & 0.65  & 0.78 & 0.00 & 0.42 & \textbf{20.65} & 12.08  \\
& Implicit-disorder & \textbackslash{}  & 0.00 & 1.63  & 2.01 & \textbf{10.87}  & 0.49 & \textbackslash{}  & \textbackslash{}  \\
\cmidrule(lr){1-10}

\multirow{3}{*}{PulpError}
& Explicit-origin & \textbackslash{}  & \textbf{0.77} & \textbackslash{}  & \textbackslash{} & \textbackslash{} & \textbackslash{} & \textbackslash{} & \textbackslash{}  \\
& Implicit-origin & \textbf{0.40}  & 0.39 & 0.00  & 0.00 & 0.00 & 0.00 & 0.00 & 0.00  \\
& Implicit-disorder & \textbackslash{}  & 0.00 & 0.00  & 0.00 & 0.00  & 0.00 & \textbackslash{}  & \textbackslash{}  \\

\bottomrule
\end{tabular}
}
\end{table}

For Crew-scheduling problems, Table \ref{AR} shows that the highest AR value is just 1.33\%. This indicates that LLMs have significant limitations in solving problems. Namely, it's difficult for LLMs with PoT method to optimize such kind of problems and thus we try to deal with problems without PoT under the condition of provided with an initial solution. Still, for Crew-sheduling problems, datasets in disorder don't perform well, which may owe to its' intricacy of the problems compared to assignment and 1D-binpacking problems. 

From Table \ref{tab6}, IndexError takes the lead while the rate of ValueError remains high and IndexError has emerged, even surpassing ValueError. 

\subsection{Steiner}\label{Steiner}

\begin{table}[ht]
\centering
\caption{Steiner Problems Performance Metrics}
\label{tab8}
\resizebox{0.8\textwidth}{!}{
\begin{tabular}{llcccccccc}
\toprule
\multirow{2}{*}{Error Type} & \multirow{2}{*}{Dataset} 
  & \multicolumn{2}{c}{GPT-4o-mini} 
  & \multicolumn{2}{c}{LLAMA3-8B} 
  & \multicolumn{2}{c}{ORLM} 
  & \multicolumn{2}{c}{DeepSeek-R1} \\
\cmidrule(lr){3-4} \cmidrule(lr){5-6} \cmidrule(lr){7-8} \cmidrule(lr){9-10}
 & & PoT & PoT+CoT & PoT & PoT+CoT & PoT & PoT+CoT & PoT & PoT+CoT \\
\midrule

\multirow{3}{*}{IndexError} 
  & Explicit-origin        & \textbackslash{} & 0.00   & \textbackslash{} & \textbackslash{} & \textbackslash{} & \textbackslash{} & \textbackslash{} & \textbackslash{} \\
  & Implicit-origin        & 4.14            & \textbf{8.55}   & 7.03             & 4.51            & 1.39            & 6.16            & 0.23            & 1.35            \\
  & Implicit-disorder      & \textbackslash{} & 0.92   & \textbf{13.82}            & 11.09           & 1.24            & 1.68            & \textbackslash{} & \textbackslash{} \\
\cmidrule(lr){1-10}

\multirow{3}{*}{FileNotFoundError}
  & Explicit-origin        & \textbackslash{} & 0.00   & \textbackslash{} & \textbackslash{} & \textbackslash{} & \textbackslash{} & \textbackslash{} & \textbackslash{} \\
  & Implicit-origin        & \textbf{3.45}            & 1.32   & 0.41             & 0.16            & 0.00            & 0.47            & 0.00            & 1.35            \\
  & Implicit-disorder      & \textbackslash{} & 0.00   & \textbf{5.69}             & 0.33            & 0.59            & 0.66            & \textbackslash{} & \textbackslash{} \\
\cmidrule(lr){1-10}

\multirow{3}{*}{OverflowError}
  & Explicit-origin        & \textbackslash{} & 0.00   & \textbackslash{} & \textbackslash{} & \textbackslash{} & \textbackslash{} & \textbackslash{} & \textbackslash{} \\
  & Implicit-origin        & 0.00            & 0.00   & 0.00             & 0.00            & 0.00            & 0.00            & 0.00            & \textbf{1.35}            \\
  & Implicit-disorder      & \textbackslash{} & 0.00   & 0.00             & \textbf{0.76}            & 0.00            & 0.00            & \textbackslash{} & \textbackslash{} \\
\cmidrule(lr){1-10}

\multirow{3}{*}{NameError}
  & Explicit-origin        & \textbackslash{} & 0.00   & \textbackslash{} & \textbackslash{} & \textbackslash{} & \textbackslash{} & \textbackslash{} & \textbackslash{} \\
  & Implicit-origin        & 2.76            & 8.55   & 8.24             & 8.09            & 3.73            & \textbf{23.70}           & 1.16            & 2.70            \\
  & Implicit-disorder      & \textbackslash{} & 0.56   & 5.69             & \textbf{14.02}           & 2.49            & 4.52            & \textbackslash{} & \textbackslash{} \\
\cmidrule(lr){1-10}

\multirow{3}{*}{ValueError}
  & Explicit-origin        & \textbackslash{} & \textbf{85.00}  & \textbackslash{} & \textbackslash{} & \textbackslash{} & \textbackslash{} & \textbackslash{} & \textbackslash{} \\
  & Implicit-origin        & 12.87           & 7.89   & 8.92             & 7.78            & 3.36            & 11.85           & 2.55            & \textbf{16.22}           \\
  & Implicit-disorder      & \textbackslash{} & 1.34   & 8.94             & \textbf{14.24}           & 1.98            & 3.28            & \textbackslash{} & \textbackslash{} \\
\cmidrule(lr){1-10}

\multirow{3}{*}{AttributeError}
  & Explicit-origin        & \textbackslash{} & 0.00   & \textbackslash{} & \textbackslash{} & \textbackslash{} & \textbackslash{} & \textbackslash{} & \textbackslash{} \\
  & Implicit-origin        & 2.30            & 1.32   & 2.16             & 1.17            & 2.12            & \textbf{5.69}            & 0.00            & 2.70            \\
  & Implicit-disorder      & \textbackslash{} & 0.21   & 1.63             & \textbf{1.85}            & 1.39            & 1.60            & \textbackslash{} & \textbackslash{} \\
\cmidrule(lr){1-10}

\multirow{3}{*}{KeyError}
  & Explicit-origin        & \textbackslash{} & 0.00   & \textbackslash{} & \textbackslash{} & \textbackslash{} & \textbackslash{} & \textbackslash{} & \textbackslash{} \\
  & Implicit-origin        & \textbf{4.37}            & 2.63   & 0.41             & 3.42            & 0.29            & 1.42            & 0.00            & 0.00            \\
  & Implicit-disorder      & \textbackslash{} & 0.28   & 1.63             & \textbf{6.63}            & 0.15            & 0.22            & \textbackslash{} & \textbackslash{} \\
\cmidrule(lr){1-10}

\multirow{3}{*}{SyntaxError}
  & Explicit-origin        & \textbackslash{} & 0.00   & \textbackslash{} & \textbackslash{} & \textbackslash{} & \textbackslash{} & \textbackslash{} & \textbackslash{} \\
  & Implicit-origin        & 3.91            & 29.61  & 8.51             & 9.34            & 5.12            & 36.49           & 6.96            & \textbf{40.54}           \\
  & Implicit-disorder      & \textbackslash{} & 1.41   & \textbf{39.84}            & 13.48           & 8.71            & 8.97            & \textbackslash{} & \textbackslash{} \\
\cmidrule(lr){1-10}

\multirow{3}{*}{TypeError}
  & Explicit-origin        & \textbackslash{} & \textbf{15.00}  & \textbackslash{} & \textbackslash{} & \textbackslash{} & \textbackslash{} & \textbackslash{} & \textbackslash{} \\
  & Implicit-origin        & 12.18           & \textbf{40.13}  & 27.57            & 21.32           & 3.00            & 11.37           & 0.46            & 5.41            \\
  & Implicit-disorder      & \textbackslash{} & 3.39   & 22.76            & \textbf{37.28}           & 2.41            & 3.65            & \textbackslash{} & \textbackslash{} \\
\cmidrule(lr){1-10}

\multirow{3}{*}{ImportError}
  & Explicit-origin        & \textbackslash{} & 0.00   & \textbackslash{} & \textbackslash{} & \textbackslash{} & \textbackslash{} & \textbackslash{} & \textbackslash{} \\
  & Implicit-origin        & 0.92            & 0.00   & 0.14             & 0.00            & 1.10            & \textbf{2.84}            & 0.00            & 0.00            \\
  & Implicit-disorder      & \textbackslash{} & 0.07   & 0.00             & 0.33            & 0.73            & \textbf{0.95}            & \textbackslash{} & \textbackslash{} \\
\cmidrule(lr){1-10}

\multirow{3}{*}{PulpError}
  & Explicit-origin        & \textbackslash{} & 0.00   & \textbackslash{} & \textbackslash{} & \textbackslash{} & \textbackslash{} & \textbackslash{} & \textbackslash{} \\
  & Implicit-origin        & \textbf{0.23}            & 0.00   & 0.00             & 0.00            & 0.00            & 0.00            & 0.00            & 0.00            \\
  & Implicit-disorder      & \textbackslash{} & 0.00   & 0.00             & 0.00            & 0.00            & 0.00            & \textbackslash{} & \textbackslash{} \\

\bottomrule
\end{tabular}
}
\end{table}

For Steiner problems, Table \ref{PR} shows that GPT-4o-mini with PoT and CoT reaches 93.67\% in PR, indicating that Steiner problems can be solved automatically with GPT-4o-mini at ease, disregarding the quality of the solutions. Besides, although the AR is relatively high, the MAPE does not reflect this alignment. Namely, MAPE is not related to AR. And this time, we have seen a significant decline in the performance of datasets in disorder. While you may ascribe it to the assumption that LLMs rely on semantic information to deal with Steiner problems, the fact is that the codes generated by LLMs often cause read error occurrence. Besides, for Steiner problem, DeepSeek-R1 tends to generate the form \textbf{[result, [solution]]}, even when prompted, which causes low performance in PR.

Table \ref{tab8} shows that SyntaxError and TypeError take the lead this time. SyntaxError occurrences arise when LLMs use CoT and PoT techniques while ValueError and AttributeError occurrences arise when LLMs use just PoT technique. Specifically, for TypeError and AttributeError, LLMs usually mix up 'List' and 'Tuple' as well as 'NumPy'.

\subsection{UBQP}\label{UBQP}

\begin{table}[ht]
\centering
\caption{UBQP Problems Performance Metrics}
\label{tab10}
\resizebox{0.8\textwidth}{!}{
\begin{tabular}{llcccccccc}
\toprule
\multirow{2}{*}{Error Type} & \multirow{2}{*}{Expression} & \multicolumn{2}{c}{GPT-4o-mini} & \multicolumn{2}{c}{LLAMA3-8B} & \multicolumn{2}{c}{ORLM} & \multicolumn{2}{c}{DeepSeek-R1}\\
\cmidrule(lr){3-4} \cmidrule(lr){5-6}\cmidrule(lr){7-8}\cmidrule(lr){9-10}
 & & PoT & PoT+CoT & PoT & PoT+CoT &PoT & PoT+CoT &PoT & PoT+CoT \\
\midrule

\multirow{3}{*}{IndexError} 
& Explicit & \textbackslash{}  & \textbf{25.53} & \textbackslash{}   &  \textbackslash{} &  \textbackslash{} &  \textbackslash{} &  \textbackslash{} &  \textbackslash{}\\
& Implicit & \textbf{62.04}  & 47.83 & 18.18  & 17.39 &0.92&6.16& 34.74 &60.66 \\
& Implicit-disorder & \textbackslash{}  & \textbf{29.04} & 18.69  & 14.29& 1.99  & 3.63& \textbackslash{}  & \textbackslash{}  \\
\cmidrule(lr){1-10}

\multirow{3}{*}{FileNotFoundError}
& Explicit & \textbackslash{}  & \textbf{8.51} & \textbackslash{}  & \textbackslash{} & \textbackslash{} & \textbackslash{} & \textbackslash{} & \textbackslash{}  \\
& Implicit & 10.22  & $\textbf{17.39}$ & 0.00  & 0.00 & 0.15 &1.11& 0.00 &1.64  \\
& Implicit-disorder & \textbackslash{}  & 0.00 & \textbf{1.52}  & 1.38 & 0.27  & 1.24& \textbackslash{}  & \textbackslash{} \\
\cmidrule(lr){1-10}

\multirow{3}{*}{OverflowError}
& Explicit & \textbackslash{}  & 0.00 & \textbackslash{}  & \textbackslash{} & \textbackslash{} & \textbackslash{} & \textbackslash{} & \textbackslash{}  \\
& Implicit & 0.73  & 0.00 & 0.00  & 0.00 & 0.15 & 0.47& \textbf{5.26} &0.00  \\
& Implicit-disorder & \textbackslash{}  & 0.00 & \textbf{0.51}  & 0.00& 0.00  & 0.10& \textbackslash{}  & \textbackslash{}  \\
\cmidrule(lr){1-10}

\multirow{3}{*}{NameError}
& Explicit & \textbackslash{}  & \textbf{2.13} & \textbackslash{}  & \textbackslash{} & \textbackslash{} & \textbackslash{} & \textbackslash{} & \textbackslash{}  \\
& Implicit & 1.46  & 1.74 & 1.65  & 3.73&1.99&$\textbf{6.79}$& 0.00 &3.28  \\
& Implicit-disorder & \textbackslash{}  & 0.00 & 2.02  & 1.84& 1.72  & $\textbf{4.88}$& \textbackslash{}  & \textbackslash{}  \\
\cmidrule(lr){1-10}

\multirow{3}{*}{ValueError}
& Explicit & \textbackslash{}  & $\textbf{53.19}$ & \textbackslash{}  & \textbackslash{} & \textbackslash{} & \textbackslash{} & \textbackslash{} & \textbackslash{}  \\
& Implicit & 23.36  & 32.17 & \textbf{41.32}  & 28.57 & 3.21 & 16.59& 0.00 &0.00  \\
& Implicit-disorder & \textbackslash{}  & 0.60 & 56.06  & $\textbf{57.60}$& 3.70  & 9.85& \textbackslash{}  & \textbackslash{}  \\
\cmidrule(lr){1-10}

\multirow{3}{*}{AttributeError}
& Explicit & \textbackslash{}  & \textbf{2.13} & \textbackslash{}  & \textbackslash{} & \textbackslash{} & \textbackslash{} & \textbackslash{} & \textbackslash{}  \\
& Implicit & 0.73  & 0.87 & 0.83  & 2.48 & 0.76 & \textbf{4.27} & 0.00 &0.00  \\
& Implicit-disorder & \textbackslash{}  & \textbf{39.82} & 2.53  & 2.76& 1.17  & 4.11& \textbackslash{}  & \textbackslash{}  \\
\cmidrule(lr){1-10}

\multirow{3}{*}{KeyError}
& Explicit & \textbackslash{}  & 0.00 & \textbackslash{}  & \textbackslash{} & \textbackslash{} & \textbackslash{} & \textbackslash{} & \textbackslash{}  \\
& Implicit & 0.00  & 0.00 & \textbf{2.48}  & 0.00 & 0.00 & 0.32& 0.00 &0.00  \\
& Implicit-disorder & \textbackslash{}  & 0.30 & \textbf{0.51}  & 0.46& 0.09  & 0.48& \textbackslash{}  & \textbackslash{}  \\
\cmidrule(lr){1-10}

\multirow{3}{*}{SyntaxError}
& Explicit & \textbackslash{}  & \textbf{4.26} & \textbackslash{}  & \textbackslash{} & \textbackslash{} & \textbackslash{} & \textbackslash{} & \textbackslash{}  \\
& Implicit & 0.73  & 0.00 & 6.61  & 1.24 & 7.80 & 10.43& 1.05 &$\textbf{11.48}$  \\
& Implicit-disorder & \textbackslash{}  & 0.00 & 0.51  & 4.61& 6.68  & $\textbf{10.61}$& \textbackslash{}  & \textbackslash{}  \\
\cmidrule(lr){1-10}

\multirow{3}{*}{TypeError}
& Explicit & \textbackslash{}  & \textbf{2.13} & \textbackslash{}  & \textbackslash{} & \textbackslash{} & \textbackslash{} & \textbackslash{} & \textbackslash{}  \\
& Implicit & 0.73  & 0.00 & 12.40  & \textbf{14.29} & 1.38 & 6.00& 1.05 &0.00  \\
& Implicit-disorder & \textbackslash{}  & 0.30 & \textbf{8.08}  & 5.53& 1.36  & 3.35& \textbackslash{}  & \textbackslash{}  \\
\cmidrule(lr){1-10}

\multirow{3}{*}{ImportError}
& Explicit & \textbackslash{}  & \textbf{2.13} & \textbackslash{}  & \textbackslash{} & \textbackslash{} & \textbackslash{} & \textbackslash{} & \textbackslash{}  \\
& Implicit & 0.00  & 0.00 & 0.00  & 1.24 & 0.46 & \textbf{2.21}& 0.00 &0.00  \\
& Implicit-disorder & \textbackslash{}  & 0.30 & 1.01  & 0.46& 0.27  & $\textbf{1.43}$& \textbackslash{}  & \textbackslash{}  \\
\cmidrule(lr){1-10}

\multirow{3}{*}{PulpError}
& Explicit & \textbackslash{}  & 0.00 & \textbackslash{}  & \textbackslash{} & \textbackslash{} & \textbackslash{} & \textbackslash{} & \textbackslash{}  \\
& Implicit & 0.00  & 0.00 & 0.00  & 0.00 & 0.00 & 0.00& 0.00 &0.00  \\
& Implicit-disorder & \textbackslash{}  & 0.00 & 0.00  & 0.00& 0.00  & 0.00& \textbackslash{}  & \textbackslash{}  \\

\bottomrule
\end{tabular}
}
\end{table}

For UBQP problems, when checking the number of parameters, We found that the number of variables achieving optimal result tends to cluster around 50, 60, and 70. Moreover, the quantity does not align with the sample size obtained from data augmentation. Considering the problem, constructed artificially, is semantically accurate, indicating that different phrasing for the same issue can lead to varying results. Aside from modeling errors, different modeling methods, built-in solvers, and even the algorithms employed can all impact the results. The former may lead to infeasibility, while the latter can cause the solution to get stuck at a local optimum or result in timeout issues. Also, the number of variables does not necessarily affect the likelihood of finding the optimal solution. This inspires us to leverage LLMs to automatically adjust the structure of problems, making it more advantageous for solving complex OR issues. Compared to steiner problems, UBQP is indeed a semantic problem.

From Table \ref{tab10}, it is more likely for weaker models to raise a ValueError, while stronger models are more inclined to raise an IndexError.  

\subsection{CVRP, MDVRP and PVRP}\label{VRP}

% CVRP Problems Error Types Performance Table
\begin{table}[ht]
\centering
\caption{CVRP Performance Metrics}
\label{tab12}
\resizebox{0.8\textwidth}{!}{
\begin{tabular}{llcccccccc}
\toprule
\multirow{2}{*}{Error Type} & \multirow{2}{*}{Expression} & \multicolumn{2}{c}{GPT-4o-mini} & \multicolumn{2}{c}{LLAMA3-8B} & \multicolumn{2}{c}{ORLM} & \multicolumn{2}{c}{DeepSeek-R1}\\
\cmidrule(lr){3-4} \cmidrule(lr){5-6}\cmidrule(lr){7-8}\cmidrule(lr){9-10}
 & & PoT & PoT+CoT & PoT & PoT+CoT &PoT & PoT+CoT &PoT & PoT+CoT \\
\midrule

\multirow{3}{*}{IndexError} 
& Explicit & \textbackslash{}  & \textbf{10.29} & \textbackslash{}   &  \textbackslash{} &  \textbackslash{} &  \textbackslash{} &  \textbackslash{} &  \textbackslash{}\\
& Implicit & $\textbf{8.52}$  & 2.87 & 2.51  & 2.30 &3.20&3.17& 0.39 &1.55 \\
& Implicit-disorder & \textbackslash{}  & \textbf{4.98} & 4.31  & 2.96& 1.71  & 0.44& \textbackslash{}  & \textbackslash{}  \\
\cmidrule(lr){1-10}

\multirow{3}{*}{FileNotFoundError}
& Explicit & \textbackslash{}  & $\textbf{15.04}$ & \textbackslash{}  & \textbackslash{} & \textbackslash{} & \textbackslash{} & \textbackslash{} & \textbackslash{}  \\
& Implicit & 0.00  & 1.91 & 0.00  & 0.27 & 2.28 &\textbf{3.49}& 0.00 &0.62  \\
& Implicit-disorder & \textbackslash{}  & 0.09 & 0.49  & 0.10 & $\textbf{7.17}$  & 1.53& \textbackslash{}  & \textbackslash{} \\
\cmidrule(lr){1-10}

\multirow{3}{*}{OverflowError}
& Explicit & \textbackslash{}  & \textbf{1.06} & \textbackslash{}  & \textbackslash{} & \textbackslash{} & \textbackslash{} & \textbackslash{} & \textbackslash{}  \\
& Implicit & 0.14  & 0.64 & 0.00  & 0.00 & 0.00 & 0.00& 0.00 &$\textbf{2.17}$  \\
& Implicit-disorder & \textbackslash{}  & \textbf{0.27} & 0.10  & 0.00& 0.00  & 0.00& \textbackslash{}  & \textbackslash{}  \\
\cmidrule(lr){1-10}

\multirow{3}{*}{NameError}
& Explicit & \textbackslash{}  & \textbf{2.90} & \textbackslash{}  & \textbackslash{} & \textbackslash{} & \textbackslash{} & \textbackslash{} & \textbackslash{}  \\
& Implicit & 2.79  & 1.59 & 1.76  & 4.47&$\textbf{20.09}$&10.79& 0.59 &2.48  \\
& Implicit-disorder & \textbackslash{}  & 2.22 & 4.40  & 4.21& \textbf{17.75}  & 4.09& \textbackslash{}  & \textbackslash{}  \\
\cmidrule(lr){1-10}

\multirow{3}{*}{ValueError}
& Explicit & \textbackslash{}  & 7.65 & \textbackslash{}  & \textbackslash{} & \textbackslash{} & \textbackslash{} & \textbackslash{} & \textbackslash{}  \\
& Implicit & $\textbf{34.43}$  & 13.69 & 7.40  & 14.88 & 12.33 & 14.29& 10.39 &10.53  \\
& Implicit-disorder & \textbackslash{}  & \textbf{18.84} & 12.62  & 10.42& 10.92  & 2.04& \textbackslash{}  & \textbackslash{}  \\
\cmidrule(lr){1-10}

\multirow{3}{*}{AttributeError}
& Explicit & \textbackslash{}  & \textbf{12.93} & \textbackslash{}  & \textbackslash{} & \textbackslash{} & \textbackslash{} & \textbackslash{} & \textbackslash{}  \\
& Implicit & 3.21  & $\textbf{14.33}$ & 5.27  & 9.88 & 11.42 & 8.89& 9.41 &10.84  \\
& Implicit-disorder & \textbackslash{}  & 13.07 & 12.82  & \textbf{17.59} & 8.53  & 1.68& \textbackslash{}  & \textbackslash{}  \\
\cmidrule(lr){1-10}

\multirow{3}{*}{KeyError}
& Explicit & \textbackslash{}  & \textbf{13.46} & \textbackslash{}  & \textbackslash{} & \textbackslash{} & \textbackslash{} & \textbackslash{} & \textbackslash{}  \\
& Implicit & 19.97  & $\textbf{27.39}$ & 6.52  & 7.44 & 12.79 & 7.62& 20.98 &2.48  \\
& Implicit-disorder & \textbackslash{}  & \textbf{22.84} & 12.04  & 8.99& 6.48  & 1.17& \textbackslash{}  & \textbackslash{}  \\
\cmidrule(lr){1-10}

\multirow{3}{*}{SyntaxError}
& Explicit & \textbackslash{}  & \textbf{3.17} & \textbackslash{}  & \textbackslash{} & \textbackslash{} & \textbackslash{} & \textbackslash{} & \textbackslash{}  \\
& Implicit & 0.77  & 3.18 & 17.69  & 21.65 & 25.57 & $\textbf{42.54}$& 9.22 &12.07  \\
& Implicit-disorder & \textbackslash{}  & 3.47 & 19.57  & 24.86& \textbf{35.84}  & 12.92& \textbackslash{}  & \textbackslash{}  \\
\cmidrule(lr){1-10}

\multirow{3}{*}{TypeError}
& Explicit & \textbackslash{}  & \textbf{16.36} & \textbackslash{}  & \textbackslash{} & \textbackslash{} & \textbackslash{} & \textbackslash{} & \textbackslash{}  \\
& Implicit & 4.61  & 16.24 & 4.77  & 10.56 & 9.13 & 5.40& 24.71 & \textbf{31.89}  \\
& Implicit-disorder & \textbackslash{}  & 15.82 & 14.38  & \textbf{16.63}& 8.87  & 1.46& \textbackslash{}  & \textbackslash{}  \\
\cmidrule(lr){1-10}

\multirow{3}{*}{ImportError}
& Explicit & \textbackslash{}  & $\textbf{16.89}$ & \textbackslash{}  & \textbackslash{} & \textbackslash{} & \textbackslash{} & \textbackslash{} & \textbackslash{}  \\
& Implicit & 2.86  & $\textbf{18.15}$ & 0.50  & 4.06 & 3.20 & 3.81& 0.00 &0.00  \\
& Implicit-disorder & \textbackslash{}  & \textbf{18.40} & 3.03  & 2.87& 2.73  & 1.82& \textbackslash{}  & \textbackslash{}  \\
\cmidrule(lr){1-10}

\multirow{3}{*}{PulpError}
& Explicit & \textbackslash{}  & \textbf{0.26} & \textbackslash{}  & \textbackslash{} & \textbackslash{} & \textbackslash{} & \textbackslash{} & \textbackslash{}  \\
& Implicit & \textbf{0.21}  & 0.00 & 0.00  & 0.00 & 0.00 & 0.00& 0.00 &0.00  \\
& Implicit-disorder & \textbackslash{}  & 0.00 & 0.00  & 0.00& 0.00  & 0.00& \textbackslash{}  & \textbackslash{}  \\

\bottomrule
\end{tabular}
}
\end{table}

% MDVRP Problems Error Types Performance Table
\begin{table}[ht]
\centering
\caption{MDVRP Performance Metrics}
\label{tab14}
\resizebox{0.9\textwidth}{!}{
\begin{tabular}{llcccccccc}
\toprule
\multirow{2}{*}{Error Type} & \multirow{2}{*}{Expression} & \multicolumn{2}{c}{GPT-4o-mini} & \multicolumn{2}{c}{LLAMA3-8B} & \multicolumn{2}{c}{ORLM} & \multicolumn{2}{c}{DeepSeek-R1}\\
\cmidrule(lr){3-4} \cmidrule(lr){5-6}\cmidrule(lr){7-8}\cmidrule(lr){9-10}
 & & PoT & PoT+CoT & PoT & PoT+CoT &PoT & PoT+CoT &PoT & PoT+CoT \\
\midrule

\multirow{3}{*}{IndexError} 
& Explicit & \textbackslash{}  & $\textbf{29.02}$ & \textbackslash{}   &  \textbackslash{} &  \textbackslash{} &  \textbackslash{} &  \textbackslash{} &  \textbackslash{}\\
& Implicit & 4.05  & 14.12 & 7.48  & 0.95 &0.31&0.61& 6.54 &\textbf{39.61} \\
& Implicit-disorder & \textbackslash{}  & 5.05 & \textbf{7.81}  & 2.51& 0.92  & 0.31& \textbackslash{}  & \textbackslash{}  \\
\cmidrule(lr){1-10}

\multirow{3}{*}{FileNotFoundError}
& Explicit & \textbackslash{}  & \textbf{6.70} & \textbackslash{}  & \textbackslash{} & \textbackslash{} & \textbackslash{} & \textbackslash{} & \textbackslash{}  \\
& Implicit & \textbf{5.41}  & 4.71 & 0.00  & 0.00 & 0.15 &1.23& 1.87 &0.48  \\
& Implicit-disorder & \textbackslash{}  & 0.00 & 0.00  & 0.00 & 1.23  & $\textbf{1.34}$& \textbackslash{}  & \textbackslash{} \\
\cmidrule(lr){1-10}

\multirow{3}{*}{OverflowError}
& Explicit & \textbackslash{}  & 0.00 & \textbackslash{}  & \textbackslash{} & \textbackslash{} & \textbackslash{} & \textbackslash{} & \textbackslash{}  \\
& Implicit & 0.68  & 0.00 & 0.00  & 0.00 & 0.00 & 0.00& $\textbf{0.93}$ &0.00  \\
& Implicit-disorder & \textbackslash{}  & 0.00 & 0.00  & 0.00& 0.00  & 0.00& \textbackslash{}  & \textbackslash{}  \\
\cmidrule(lr){1-10}

\multirow{3}{*}{NameError}
& Explicit & \textbackslash{}  & \textbf{11.61} & \textbackslash{}  & \textbackslash{} & \textbackslash{} & \textbackslash{} & \textbackslash{} & \textbackslash{}  \\
& Implicit & 7.43  & 5.88 & $\textbf{14.97}$  & 10.48&2.76&3.53& 1.87 & 1.45  \\
& Implicit-disorder & \textbackslash{}  & 7.41 & 9.90  & \textbf{11.06}& 2.66  & 2.99& \textbackslash{}  & \textbackslash{}  \\
\cmidrule(lr){1-10}

\multirow{3}{*}{ValueError}
& Explicit & \textbackslash{}  & \textbf{16.07} & \textbackslash{}  & \textbackslash{} & \textbackslash{} & \textbackslash{} & \textbackslash{} & \textbackslash{}  \\
& Implicit & 10.81  & 13.53 & 5.44  & 5.24 & 1.07 & 1.07& $\textbf{29.91}$ &21.26  \\
& Implicit-disorder & \textbackslash{}  & 8.75 & \textbf{23.44}  & 17.59 & 1.13  & 1.44& \textbackslash{}  & \textbackslash{}  \\
\cmidrule(lr){1-10}

\multirow{3}{*}{AttributeError}
& Explicit & \textbackslash{}  & \textbf{9.38} & \textbackslash{}  & \textbackslash{} & \textbackslash{} & \textbackslash{} & \textbackslash{} & \textbackslash{}  \\
& Implicit & 8.78  & 0.00 & 5.44  & 7.14 & 0.46 & 1.07& 5.61 & \textbf{9.66}  \\
& Implicit-disorder & \textbackslash{}  & 0.67 & 6.77  & \textbf{7.54} & 1.13  & 0.93& \textbackslash{}  & \textbackslash{}  \\
\cmidrule(lr){1-10}

\multirow{3}{*}{KeyError}
& Explicit & \textbackslash{}  & 6.25 & \textbackslash{}  & \textbackslash{} & \textbackslash{} & \textbackslash{} & \textbackslash{} & \textbackslash{}  \\
& Implicit & $\textbf{10.14}$  & 4.71 & 5.44  & 1.43 & 0.31 & 0.92& 2.80 &3.38  \\
& Implicit-disorder & \textbackslash{}  & 2.69 & \textbf{5.21}  & 4.02& 0.20  & 0.31& \textbackslash{}  & \textbackslash{}  \\
\cmidrule(lr){1-10}

\multirow{3}{*}{SyntaxError}
& Explicit & \textbackslash{}  & \textbf{4.91} & \textbackslash{}  & \textbackslash{} & \textbackslash{} & \textbackslash{} & \textbackslash{} & \textbackslash{}  \\
& Implicit & 11.49  & $\textbf{23.53}$ & 23.81  & 12.86 & 6.43 & 6.29& 9.35 &11.59  \\
& Implicit-disorder & \textbackslash{}  & 2.36 & 15.63  & \textbf{24.12}& 6.45  & 5.57& \textbackslash{}  & \textbackslash{}  \\
\cmidrule(lr){1-10}

\multirow{3}{*}{TypeError}
& Explicit & \textbackslash{}  & \textbf{9.38} & \textbackslash{}  & \textbackslash{} & \textbackslash{} & \textbackslash{} & \textbackslash{} & \textbackslash{}  \\
& Implicit & $\textbf{37.16}$  & 30.59 & 3.40  & 9.52 & 1.99 & 3.53& 13.08 & 12.56  \\
& Implicit-disorder & \textbackslash{}  & \textbf{19.53} & 14.58  & 14.57& 2.46  & 2.68& \textbackslash{}  & \textbackslash{}  \\
\cmidrule(lr){1-10}

\multirow{3}{*}{ImportError}
& Explicit & \textbackslash{}  & \textbf{6.70} & \textbackslash{}  & \textbackslash{} & \textbackslash{} & \textbackslash{} & \textbackslash{} & \textbackslash{}  \\
& Implicit & \textbf{4.05}  & 1.76 & 0.68  & 1.90 & 0.61 & 0.92& 0.00 &0.00  \\
& Implicit-disorder & \textbackslash{}  & 2.69 & \textbf{5.21}  & 2.01& 0.31  & 0.52& \textbackslash{}  & \textbackslash{}  \\
\cmidrule(lr){1-10}

\multirow{3}{*}{PulpError}
& Explicit & \textbackslash{}  & 0.00 & \textbackslash{}  & \textbackslash{} & \textbackslash{} & \textbackslash{} & \textbackslash{} & \textbackslash{}  \\
& Implicit & 0.00  & $\textbf{1.18}$ & 0.00  & 0.00 & 0.00 & 0.00& 0.00 &0.00  \\
& Implicit-disorder & \textbackslash{}  & \textbf{0.67} & 0.00  & 0.00& 0.00  & 0.00& \textbackslash{}  & \textbackslash{}  \\

\bottomrule
\end{tabular}
}
\end{table}

\clearpage

% PVRP Problems Error Types Performance Table
\begin{table}[ht]
\centering
\caption{PVRP Performance Metrics}
\label{tab16}
\resizebox{0.9\textwidth}{!}{
\begin{tabular}{llcccccccc}
\toprule
\multirow{2}{*}{Error Type} & \multirow{2}{*}{Expression} & \multicolumn{2}{c}{GPT-4o-mini} & \multicolumn{2}{c}{LLAMA3-8B} & \multicolumn{2}{c}{ORLM} & \multicolumn{2}{c}{DeepSeek-R1}\\
\cmidrule(lr){3-4} \cmidrule(lr){5-6}\cmidrule(lr){7-8}\cmidrule(lr){9-10}
 & & PoT & PoT+CoT & PoT & PoT+CoT &PoT & PoT+CoT &PoT & PoT+CoT \\
\midrule

\multirow{3}{*}{IndexError} 
& Explicit & \textbackslash{}  & \textbf{2.14} & \textbackslash{}   &  \textbackslash{} &  \textbackslash{} &  \textbackslash{} &  \textbackslash{} &  \textbackslash{}\\
& Implicit & $\textbf{7.97}$  & 5.47 & 1.66  & 2.11 &0.16&0.64& 0.00 & 1.14 \\
& Implicit-disorder & \textbackslash{}  & 3.02 & \textbf{5.09}  & 3.65& 0.29  & 0.68& \textbackslash{}  & \textbackslash{}  \\
\cmidrule(lr){1-10}

\multirow{3}{*}{FileNotFoundError}
& Explicit & \textbackslash{}  & 0.00 & \textbackslash{}  & \textbackslash{} & \textbackslash{} & \textbackslash{} & \textbackslash{} & \textbackslash{}  \\
& Implicit & \textbf{3.62}  & 2.34 & 0.00  & 0.00 & 0.48 &0.48& 0.00 &0.00  \\
& Implicit-disorder & \textbackslash{}  & 0.00 & 0.00  & 0.00 & $\textbf{0.87}$  & 0.39& \textbackslash{}  & \textbackslash{} \\
\cmidrule(lr){1-10}

\multirow{3}{*}{OverflowError}
& Explicit & \textbackslash{}  & 0.00 & \textbackslash{}  & \textbackslash{} & \textbackslash{} & \textbackslash{} & \textbackslash{} & \textbackslash{}  \\
& Implicit & 0.00  & 0.00 & 0.00  & 0.00 & 0.00 & 0.00& 0.00 &$\textbf{1.14}$  \\
& Implicit-disorder & \textbackslash{}  & \textbf{0.43} & 0.00  & 0.00& 0.00  & 0.00& \textbackslash{}  & \textbackslash{}  \\
\cmidrule(lr){1-10}

\multirow{3}{*}{NameError}
& Explicit & \textbackslash{}  & \textbf{8.02} & \textbackslash{}  & \textbackslash{} & \textbackslash{} & \textbackslash{} & \textbackslash{} & \textbackslash{}  \\
& Implicit & 9.42  & \textbf{10.94} & 5.52  & 6.75&2.75&2.56& 4.59 &  3.41  \\
& Implicit-disorder & \textbackslash{}  & 5.17 & 12.04  & \textbf{12.33}& 2.50  & 3.29& \textbackslash{}  & \textbackslash{}  \\
\cmidrule(lr){1-10}

\multirow{3}{*}{ValueError}
& Explicit & \textbackslash{}  & \textbf{3.74} & \textbackslash{}  & \textbackslash{} & \textbackslash{} & \textbackslash{} & \textbackslash{} & \textbackslash{}  \\
& Implicit & 10.87  & $\textbf{33.59}$ & 3.31  & 2.95 & 0.00 & 0.64& 24.77 & 14.77  \\
& Implicit-disorder & \textbackslash{}  & 10.78 & 10.19  & \textbf{11.87} & 0.77  & 0.58& \textbackslash{}  & \textbackslash{}  \\
\cmidrule(lr){1-10}

\multirow{3}{*}{AttributeError}
& Explicit & \textbackslash{}  & \textbf{5.88} & \textbackslash{}  & \textbackslash{} & \textbackslash{} & \textbackslash{} & \textbackslash{} & \textbackslash{}  \\
& Implicit & 5.07  & 2.34 & 2.76  & \textbf{5.91} & 0.48 & 1.76& 5.50 & 1.14  \\
& Implicit-disorder & \textbackslash{}  & 1.72 & \textbf{9.72}  & 5.48& 1.06  & 1.06& \textbackslash{}  & \textbackslash{}  \\
\cmidrule(lr){1-10}

\multirow{3}{*}{KeyError}
& Explicit & \textbackslash{}  & $\textbf{32.09}$ & \textbackslash{}  & \textbackslash{} & \textbackslash{} & \textbackslash{} & \textbackslash{} & \textbackslash{}  \\
& Implicit & 18.84  & 21.88 & 4.97  & 3.38 & 0.16 & 0.48& $\textbf{4.59}$ &4.55  \\
& Implicit-disorder & \textbackslash{}  & 5.60 & 6.48  & \textbf{8.22} & 0.58  & 0.97& \textbackslash{}  & \textbackslash{}  \\
\cmidrule(lr){1-10}

\multirow{3}{*}{SyntaxError}
& Explicit & \textbackslash{}  & \textbf{3.21} & \textbackslash{}  & \textbackslash{} & \textbackslash{} & \textbackslash{} & \textbackslash{} & \textbackslash{}  \\
& Implicit & 18.11  & 7.03 & \textbf{22.10}  & 15.61 & 3.39 & 4.33& 10.09 & 15.91  \\
& Implicit-disorder & \textbackslash{}  & 4.31 & \textbf{25.00}  & 22.37& 6.65  & 6.57& \textbackslash{}  & \textbackslash{}  \\
\cmidrule(lr){1-10}

\multirow{3}{*}{TypeError}
& Explicit & \textbackslash{}  & \textbf{5.35} & \textbackslash{}  & \textbackslash{} & \textbackslash{} & \textbackslash{} & \textbackslash{} & \textbackslash{}  \\
& Implicit & 18.12  & 11.72 & 4.97  & 8.44 & 0.81 & 2.72& $\textbf{16.51}$ & 10.23  \\
& Implicit-disorder & \textbackslash{}  & 11.64 & \textbf{17.59}  & 12.79& 2.02  & 1.93& \textbackslash{}  & \textbackslash{}  \\
\cmidrule(lr){1-10}

\multirow{3}{*}{ImportError}
& Explicit & \textbackslash{}  & \textbf{6.42} & \textbackslash{}  & \textbackslash{} & \textbackslash{} & \textbackslash{} & \textbackslash{} & \textbackslash{}  \\
& Implicit & $\textbf{7.97}$  & 3.13 & 1.10  & 1.69 & 0.81 & 0.80& 0.00 &0.00  \\
& Implicit-disorder & \textbackslash{}  & \textbf{5.60} & 0.46  & 0.46& 0.48  & 0.19& \textbackslash{}  & \textbackslash{}  \\
\cmidrule(lr){1-10}

\multirow{3}{*}{PulpError}
& Explicit & \textbackslash{}  & $\textbf{2.67}$ & \textbackslash{}  & \textbackslash{} & \textbackslash{} & \textbackslash{} & \textbackslash{} & \textbackslash{}  \\
& Implicit & 0.00  & \textbf{1.56} & 0.00  & 0.00 & 0.00 & 0.00& 0.00 &0.00  \\
& Implicit-disorder & \textbackslash{}  & 0.43 & $\textbf{0.46}$  & 0.00& 0.00  & 0.00& \textbackslash{}  & \textbackslash{}  \\

\bottomrule
\end{tabular}
}
\end{table}

For the VRP problems, Table \ref{tab12}.\ref{tab14}.\ref{tab16} indicate that LLAMA3-8B is not good at this series of problems, with frequent SyntaxError occurrences. We suspect that the issues may be due to insufficient training on this type of problem. In comparison, GPT-4o-mini tends to produce more KeyErrors. 

This time, weaker models still tend to raise a SyntaxError while stronger models tend to raise KeyError and TypeError.

\clearpage

\subsection{Aircraft Landing}

\begin{table}[ht]
\centering
\caption{Aircraft Landing Problems Performance Metrics}
\label{tab18}
\resizebox{0.9\textwidth}{!}{
\begin{tabular}{llcccccccc}
\toprule
\multirow{2}{*}{Error Type} & \multirow{2}{*}{Dataset}
  & \multicolumn{2}{c}{GPT-4o-mini}
  & \multicolumn{2}{c}{LLAMA3-8B}
  & \multicolumn{2}{c}{ORLM}
  & \multicolumn{2}{c}{DeepSeek-R1}\\
\cmidrule(lr){3-4} \cmidrule(lr){5-6} \cmidrule(lr){7-8} \cmidrule(lr){9-10}
 & & PoT & PoT+CoT & PoT & PoT+CoT & PoT & PoT+CoT & PoT & PoT+CoT \\
\midrule

\multirow{3}{*}{IndexError}
  & Explicit-origin        & \textbackslash{} & \textbf{13.00}   & \textbackslash{} & \textbackslash{} & \textbackslash{} & \textbackslash{} & \textbackslash{} & \textbackslash{} \\
  & Implicit-origin        & 3.33            & 2.60    & 2.38             & 1.37            & 0.93            & 0.00            & 9.52            & \textbf{50.00}           \\
  & Implicit-disorder      & \textbackslash{} & 0.74    & \textbf{5.56}             & 3.57            & 2.30            & 0.70            & \textbackslash{} & \textbackslash{} \\
\cmidrule(lr){1-10}

\multirow{3}{*}{FileNotFoundError}
  & Explicit-origin        & \textbackslash{} & \textbf{4.00}    & \textbackslash{} & \textbackslash{} & \textbackslash{} & \textbackslash{} & \textbackslash{} & \textbackslash{} \\
  & Implicit-origin        & 0.00            & 0.00    & 0.00             & 0.00            & \textbf{1.40}            & 0.00            & 0.00            & 0.00            \\
  & Implicit-disorder      & \textbackslash{} & 0.00    & 0.00             & 0.00            & 1.15            & \textbf{7.96}            & \textbackslash{} & \textbackslash{} \\
\cmidrule(lr){1-10}

\multirow{3}{*}{OverflowError}
  & Explicit-origin        & \textbackslash{} & 0.00    & \textbackslash{} & \textbackslash{} & \textbackslash{} & \textbackslash{} & \textbackslash{} & \textbackslash{} \\
  & Implicit-origin        & 0.00            & 0.00    & 0.00             & 0.00            & 0.00            & 0.00            & 0.00            & 0.00            \\
  & Implicit-disorder      & \textbackslash{} & 0.00    & 0.00             & 0.00            & 0.00            & 0.00            & \textbackslash{} & \textbackslash{} \\
\cmidrule(lr){1-10}

\multirow{3}{*}{NameError}
  & Explicit-origin        & \textbackslash{} & \textbf{1.00}    & \textbackslash{} & \textbackslash{} & \textbackslash{} & \textbackslash{} & \textbackslash{} & \textbackslash{} \\
  & Implicit-origin        & 3.33            & \textbf{6.49}    & 2.38             & 2.74            & 2.80            & 4.65            & 0.00            & 0.00            \\
  & Implicit-disorder      & \textbackslash{} & 2.22    & \textbf{9.72}             & 7.14            & 2.99            & 3.04            & \textbackslash{} & \textbackslash{} \\
\cmidrule(lr){1-10}

\multirow{3}{*}{ValueError}
  & Explicit-origin        & \textbackslash{} & \textbf{37.00}   & \textbackslash{} & \textbackslash{} & \textbackslash{} & \textbackslash{} & \textbackslash{} & \textbackslash{} \\
  & Implicit-origin        & 30.00           & 38.96   & \textbf{59.52}            & 31.51           & 4.21            & 1.86            & 40.48           & 37.50           \\
  & Implicit-disorder      & \textbackslash{} & 34.81   & \textbf{55.56}            & 40.48           & 3.45            & 1.87            & \textbackslash{} & \textbackslash{} \\
\cmidrule(lr){1-10}

\multirow{3}{*}{AttributeError}
  & Explicit-origin        & \textbackslash{} & \textbf{9.00}    & \textbackslash{} & \textbackslash{} & \textbackslash{} & \textbackslash{} & \textbackslash{} & \textbackslash{} \\
  & Implicit-origin        & 0.00            & 0.00    & 0.00             & 0.00            & 1.40            & \textbf{2.33}            & 0.00            & 0.00            \\
  & Implicit-disorder      & \textbackslash{} & 0.74    & \textbf{2.78}             & 1.19            & 1.38            & 0.23            & \textbackslash{} & \textbackslash{} \\
\cmidrule(lr){1-10}

\multirow{3}{*}{KeyError}
  & Explicit-origin        & \textbackslash{} & 0.00    & \textbackslash{} & \textbackslash{} & \textbackslash{} & \textbackslash{} & \textbackslash{} & \textbackslash{} \\
  & Implicit-origin        & 0.00            & 0.00    & 0.00             & 1.37            & 0.93            & \textbf{1.40}            & 0.00            & 0.00            \\
  & Implicit-disorder      & \textbackslash{} & 0.00    & 1.39             & \textbf{7.14}            & 1.15            & 0.23            & \textbackslash{} & \textbackslash{} \\
\cmidrule(lr){1-10}

\multirow{3}{*}{SyntaxError}
  & Explicit-origin        & \textbackslash{} & 1.00    & \textbackslash{} & \textbackslash{} & \textbackslash{} & \textbackslash{} & \textbackslash{} & \textbackslash{} \\
  & Implicit-origin        & 10.00           & 11.69   & 4.76             & 2.74            & 8.41            & 6.05            & \textbf{16.67}           & 10.94           \\
  & Implicit-disorder      & \textbackslash{} & 0.00    & 2.78             & \textbf{14.29}           & 7.36            & 3.98            & \textbackslash{} & \textbackslash{} \\
\cmidrule(lr){1-10}

\multirow{3}{*}{TypeError}
  & Explicit-origin        & \textbackslash{} & \textbf{34.00}   & \textbackslash{} & \textbackslash{} & \textbackslash{} & \textbackslash{} & \textbackslash{} & \textbackslash{} \\
  & Implicit-origin        & \textbf{53.33}           & 38.96   & 7.14             & 9.59            & 4.21            & 3.72            & 7.14            & 1.56            \\
  & Implicit-disorder      & \textbackslash{} & 10.37   & 13.89            & \textbf{15.48}           & 1.38            & 0.94            & \textbackslash{} & \textbackslash{} \\
\cmidrule(lr){1-10}

\multirow{3}{*}{ImportError}
  & Explicit-origin        & \textbackslash{} & 0.00    & \textbackslash{} & \textbackslash{} & \textbackslash{} & \textbackslash{} & \textbackslash{} & \textbackslash{} \\
  & Implicit-origin        & 0.00            & 0.00    & 0.00             & 0.00            & \textbf{0.47}            & \textbf{0.47}            & 0.00            & 0.00            \\
  & Implicit-disorder      & \textbackslash{} & 0.00    & \textbf{2.78}             & 1.19            & 0.46            & 0.00            & \textbackslash{} & \textbackslash{} \\
\cmidrule(lr){1-10}

\multirow{3}{*}{PulpError}
  & Explicit-origin        & \textbackslash{} & 1.00    & \textbackslash{} & \textbackslash{} & \textbackslash{} & \textbackslash{} & \textbackslash{} & \textbackslash{} \\
  & Implicit-origin        & 0.00            & 0.00    & 0.00             & 0.00            & 0.00            & 0.00            & 0.00            & 0.00            \\
  & Implicit-disorder      & \textbackslash{} & \textbf{0.74}    & 0.00             & 0.00            & 0.00            & 0.00            & \textbackslash{} & \textbackslash{} \\

\bottomrule
\end{tabular}
}
\end{table}

For the Aircraft Landing problems, Table \ref{PR}.\ref{AR}.\ref{MAPE} still shows that LLAMA3-8B is not good at dealing with such kind of problems. Besides, There is a strange phenomenon that AR is high while MAPE is unavailable. To find out what happened, we refer to the original datasets and check out the statistic on the testing result. According to Section \ref{statistic}, the variables of problems, that are dealt with accurately, are [15,15,20,20,30] while two samples are only used to construct datasets in the form of natural language in this type of problems. Namely, for some data, it's easier to reach the optimal result while others may encounter obstacles, depending on the style of description in natural language.

From Table \ref{tab18}, ValueError is raised frequently and there is no difference between weeker or stronger models. And, ORLM still tends to raise SyntaxError.

\subsection{Generalised Assignment}

\begin{table}[ht]
\centering
\caption{Generalised Assignment Problems Performance Metrics}
\label{tab20}
\resizebox{0.9\textwidth}{!}{
\begin{tabular}{llcccccccc}
\toprule
\multirow{2}{*}{Error Type} & \multirow{2}{*}{Dataset}
  & \multicolumn{2}{c}{GPT-4o-mini}
  & \multicolumn{2}{c}{LLAMA3-8B}
  & \multicolumn{2}{c}{ORLM}
  & \multicolumn{2}{c}{DeepSeek-R1} \\
\cmidrule(lr){3-4} \cmidrule(lr){5-6} \cmidrule(lr){7-8} \cmidrule(lr){9-10}
 & & PoT & PoT+CoT & PoT & PoT+CoT & PoT & PoT+CoT & PoT & PoT+CoT \\
\midrule

\multirow{3}{*}{IndexError}
  & Explicit-origin        & \textbackslash{} & \textbf{60.80} & \textbackslash{} & \textbackslash{} & \textbackslash{} & \textbackslash{} & \textbackslash{} & \textbackslash{} \\
  & Implicit-origin        & 12.98            & 9.15  & 13.41            & 17.95           & 1.15            & 2.31            & \textbf{41.67}           & 37.80           \\
  & Implicit-disorder      & \textbackslash{} & 3.85  & 9.25             & \textbf{12.80}           & 3.21            & 4.24            & \textbackslash{} & \textbackslash{} \\
\cmidrule(lr){1-10}

\multirow{3}{*}{FileNotFoundError}
  & Explicit-origin        & \textbackslash{} & \textbf{8.52}  & \textbackslash{} & \textbackslash{} & \textbackslash{} & \textbackslash{} & \textbackslash{} & \textbackslash{} \\
  & Implicit-origin        & 2.67             & 5.23  & \textbf{8.94}             & 1.03            & 0.46            & 0.93            & 0.00            & 0.00            \\
  & Implicit-disorder      & \textbackslash{} & 3.40  & \textbf{8.90}             & 7.61            & 4.49            & 7.63            & \textbackslash{} & \textbackslash{} \\
\cmidrule(lr){1-10}

\multirow{3}{*}{OverflowError}
  & Explicit-origin        & \textbackslash{} & 0.00  & \textbackslash{} & \textbackslash{} & \textbackslash{} & \textbackslash{} & \textbackslash{} & \textbackslash{} \\
  & Implicit-origin        & \textbf{0.38}             & 0.00  & 0.00             & 0.00            & 0.00            & 0.00            & 0.00            & 0.00            \\
  & Implicit-disorder      & \textbackslash{} & 0.00  & 0.00             & 0.00            & 0.00            & 0.00            & \textbackslash{} & \textbackslash{} \\
\cmidrule(lr){1-10}

\multirow{3}{*}{NameError}
  & Explicit-origin        & \textbackslash{} & 5.11  & \textbackslash{} & \textbackslash{} & \textbackslash{} & \textbackslash{} & \textbackslash{} & \textbackslash{} \\
  & Implicit-origin        & \textbf{15.27}            & 14.38 & 1.68             & 5.13            & 4.70            & 5.09            & 0.00            & 0.00            \\
  & Implicit-disorder      & \textbackslash{} & \textbf{20.41} & 7.12             & 6.92            & 19.23           & 19.49           & \textbackslash{} & \textbackslash{} \\
\cmidrule(lr){1-10}

\multirow{3}{*}{ValueError}
  & Explicit-origin        & \textbackslash{} & \textbf{7.95}  & \textbackslash{} & \textbackslash{} & \textbackslash{} & \textbackslash{} & \textbackslash{} & \textbackslash{} \\
  & Implicit-origin        & 16.79            & 29.74 & \textbf{36.87}            & 24.62           & 4.36            & 3.70            & 15.15           & 8.54            \\
  & Implicit-disorder      & \textbackslash{} & 19.50 & \textbf{44.48}            & 31.14           & 14.74           & 16.10           & \textbackslash{} & \textbackslash{} \\
\cmidrule(lr){1-10}

\multirow{3}{*}{AttributeError}
  & Explicit-origin        & \textbackslash{} & 3.98  & \textbackslash{} & \textbackslash{} & \textbackslash{} & \textbackslash{} & \textbackslash{} & \textbackslash{} \\
  & Implicit-origin        & 0.38             & 1.30  & \textbf{5.03}             & 4.10            & 3.90            & 3.24            & 0.00            & 0.00            \\
  & Implicit-disorder      & \textbackslash{} & 2.27  & 3.91             & 4.84            & 9.94            & \textbf{10.59}           & \textbackslash{} & \textbackslash{} \\
\cmidrule(lr){1-10}

\multirow{3}{*}{KeyError}
  & Explicit-origin        & \textbackslash{} & \textbf{7.39}  & \textbackslash{} & \textbackslash{} & \textbackslash{} & \textbackslash{} & \textbackslash{} & \textbackslash{} \\
  & Implicit-origin        & 1.15             & \textbf{15.03} & 6.15             & 4.62            & 0.46            & 0.46            & 0.76            & 0.00            \\
  & Implicit-disorder      & \textbackslash{} & 0.00  & 1.42             & \textbf{4.50}            & 1.28            & 0.00            & \textbackslash{} & \textbackslash{} \\
\cmidrule(lr){1-10}

\multirow{3}{*}{SyntaxError}
  & Explicit-origin        & \textbackslash{} & 0.57  & \textbackslash{} & \textbackslash{} & \textbackslash{} & \textbackslash{} & \textbackslash{} & \textbackslash{} \\
  & Implicit-origin        & 3.82             & 2.29  & 15.08            & \textbf{15.38}           & 3.21            & 10.07           & 2.27            & 0.00            \\
  & Implicit-disorder      & \textbackslash{} & 2.04  & 12.10            & 16.96           & 31.41           & \textbf{32.20}           & \textbackslash{} & \textbackslash{} \\
\cmidrule(lr){1-10}

\multirow{3}{*}{TypeError}
  & Explicit-origin        & \textbackslash{} & \textbf{2.27}  & \textbackslash{} & \textbackslash{} & \textbackslash{} & \textbackslash{} & \textbackslash{} & \textbackslash{} \\
  & Implicit-origin        & \textbf{44.27}            & 21.57 & 3.35             & 5.64            & 4.36            & 2.66            & 40.15           & 1.22            \\
  & Implicit-disorder      & \textbackslash{} & 3.17  & 6.05             & 7.96            & 13.78           & 8.47            & \textbackslash{} & \textbackslash{} \\
\cmidrule(lr){1-10}

\multirow{3}{*}{ImportError}
  & Explicit-origin        & \textbackslash{} & 2.27  & \textbackslash{} & \textbackslash{} & \textbackslash{} & \textbackslash{} & \textbackslash{} & \textbackslash{} \\
  & Implicit-origin        & 0.38             & \textbf{1.31}  & 0.00             & 0.51            & 0.69            & 0.69            & 0.00            & 0.00            \\
  & Implicit-disorder      & \textbackslash{} & 0.23  & 0.36             & 0.00            & \textbf{1.92}            & 1.27            & \textbackslash{} & \textbackslash{} \\
\cmidrule(lr){1-10}

\multirow{3}{*}{PulpError}
  & Explicit-origin        & \textbackslash{} & \textbf{1.14}  & \textbackslash{} & \textbackslash{} & \textbackslash{} & \textbackslash{} & \textbackslash{} & \textbackslash{} \\
  & Implicit-origin        & \textbf{1.91}             & 0.33  & 0.00             & 0.00            & 0.00            & 0.00            & 0.00            & 0.00            \\
  & Implicit-disorder      & \textbackslash{} & \textbf{0.91}  & 0.00             & 0.00            & 0.00            & 0.00            & \textbackslash{} & \textbackslash{} \\

\bottomrule
\end{tabular}
}
\end{table}

For Generalised Assignment problems, Table \ref{AR}.\ref{MAPE} provides evidence that datasets in disorder achieve a significant performance boost, far surpassing the original datasets. Actually, the trend is similar to Assignment problems. A possible explanation is that the datasets in disorder input the task requirements in advance to enable LLMs to address problems more effectively. 

Fron Table \ref{tab20}, DeepSeek-R1 tends to raise IndexError while other models except ORLM tend to raise ValueError. Similarly, ORLM tends to raise SyntaxError.

\subsection{Multi-dimensional Knapsack}

\begin{table}[ht]
\centering
\caption{Multi-dimensional Knapsack Problems Performance Metrics}
\label{tab22}
\resizebox{0.9\textwidth}{!}{
\begin{tabular}{llcccccccc}
\toprule
\multirow{2}{*}{Error Type} & \multirow{2}{*}{Dataset}
  & \multicolumn{2}{c}{GPT-4o-mini}
  & \multicolumn{2}{c}{LLAMA3-8B}
  & \multicolumn{2}{c}{ORLM}
  & \multicolumn{2}{c}{DeepSeek-R1} \\
\cmidrule(lr){3-4} \cmidrule(lr){5-6} \cmidrule(lr){7-8} \cmidrule(lr){9-10}
 & & PoT & PoT+CoT & PoT & PoT+CoT & PoT & PoT+CoT & PoT & PoT+CoT \\
\midrule

\multirow{3}{*}{IndexError}
  & Explicit-origin        & \textbackslash{} & \textbf{37.62} & \textbackslash{} & \textbackslash{} & \textbackslash{} & \textbackslash{} & \textbackslash{} & \textbackslash{} \\
  & Implicit-origin        & \textbf{26.28}            & 22.61 & 6.58             & 6.41            & 3.38            & 4.22            & 15.88           & 16.22           \\
  & Implicit-disorder      & \textbackslash{} & \textbf{17.59} & 4.16             & 4.87            & 3.66            & 2.97            & \textbackslash{} & \textbackslash{} \\
\cmidrule(lr){1-10}

\multirow{3}{*}{FileNotFoundError}
  & Explicit-origin        & \textbackslash{} & \textbf{0.99}  & \textbackslash{} & \textbackslash{} & \textbackslash{} & \textbackslash{} & \textbackslash{} & \textbackslash{} \\
  & Implicit-origin        & 1.53             & 1.01  & 0.12             & 0.38            & \textbf{6.39}            & 5.84            & 0.00            & 0.00            \\
  & Implicit-disorder      & \textbackslash{} & 0.08  & 0.13             & 0.23            & \textbf{6.50}            & 0.10            & \textbackslash{} & \textbackslash{} \\
\cmidrule(lr){1-10}

\multirow{3}{*}{OverflowError}
  & Explicit-origin        & \textbackslash{} & 0.00  & \textbackslash{} & \textbackslash{} & \textbackslash{} & \textbackslash{} & \textbackslash{} & \textbackslash{} \\
  & Implicit-origin        & 0.00             & 0.00  & 0.00             & 0.00            & 0.00            & 0.00            & 0.00            & 0.00            \\
  & Implicit-disorder      & \textbackslash{} & 0.00  & 0.00             & 0.00            & 0.00            & 0.00            & \textbackslash{} & \textbackslash{} \\
\cmidrule(lr){1-10}

\multirow{3}{*}{NameError}
  & Explicit-origin        & \textbackslash{} & \textbf{2.31}  & \textbackslash{} & \textbackslash{} & \textbackslash{} & \textbackslash{} & \textbackslash{} & \textbackslash{} \\
  & Implicit-origin        & 2.04             & 2.76  & 1.85             & 2.14            & \textbf{16.54}           & 9.42            & 0.00            & 0.00            \\
  & Implicit-disorder      & \textbackslash{} & 2.65  & 0.76             & 0.93            & \textbf{13.41}           & 4.21            & \textbackslash{} & \textbackslash{} \\
\cmidrule(lr){1-10}

\multirow{3}{*}{ValueError}
  & Explicit-origin        & \textbackslash{} & 42.24 & \textbackslash{} & \textbackslash{} & \textbackslash{} & \textbackslash{} & \textbackslash{} & \textbackslash{} \\
  & Implicit-origin        & 52.04            & 52.26 & \textbf{75.98}            & 73.99           & 32.71           & 31.82           & 29.79           & 29.73           \\
  & Implicit-disorder      & \textbackslash{} & 44.28 & 48.42            & \textbf{54.87}           & 18.29           & 10.44           & \textbackslash{} & \textbackslash{} \\
\cmidrule(lr){1-10}

\multirow{3}{*}{AttributeError}
  & Explicit-origin        & \textbackslash{} & 0.33  & \textbackslash{} & \textbackslash{} & \textbackslash{} & \textbackslash{} & \textbackslash{} & \textbackslash{} \\
  & Implicit-origin        & 0.00             & 0.98  & 2.08             & 3.77            & \textbf{7.14}            & 5.52            & 3.60            & 0.00            \\
  & Implicit-disorder      & \textbackslash{} & 0.30  & 2.40             & 2.09            & 10.57           & \textbf{17.62}           & \textbackslash{} & \textbackslash{} \\
\cmidrule(lr){1-10}

\multirow{3}{*}{KeyError}
  & Explicit-origin        & \textbackslash{} & 0.00  & \textbackslash{} & \textbackslash{} & \textbackslash{} & \textbackslash{} & \textbackslash{} & \textbackslash{} \\
  & Implicit-origin        & 0.00             & 0.00  & 0.23             & 0.25            & 0.75            & 0.65            & \textbf{12.60}           & 0.00            \\
  & Implicit-disorder      & \textbackslash{} & 0.23  & 0.13             & 0.58            & 0.81            & \textbf{8.91}            & \textbackslash{} & \textbackslash{} \\
\cmidrule(lr){1-10}

\multirow{3}{*}{SyntaxError}
  & Explicit-origin        & \textbackslash{} & \textbf{10.56} & \textbackslash{} & \textbackslash{} & \textbackslash{} & \textbackslash{} & \textbackslash{} & \textbackslash{} \\
  & Implicit-origin        & 0.77             & 1.26  & 8.31             & 6.66            & 21.05           & \textbf{33.77}           & 1.31            & 0.34            \\
  & Implicit-disorder      & \textbackslash{} & 0.76  & 10.47            & 6.26            & \textbf{38.62}           & 24.81           & \textbackslash{} & \textbackslash{} \\
\cmidrule(lr){1-10}

\multirow{3}{*}{TypeError}
  & Explicit-origin        & \textbackslash{} & 5.61  & \textbackslash{} & \textbackslash{} & \textbackslash{} & \textbackslash{} & \textbackslash{} & \textbackslash{} \\
  & Implicit-origin        & 17.09            & \textbf{19.10} & 4.73             & 6.03            & 9.77            & 6.17            & 3.27            & 0.68            \\
  & Implicit-disorder      & \textbackslash{} & 16.53 & 2.90             & 6.38            & 6.91            & \textbf{16.67}           & \textbackslash{} & \textbackslash{} \\
\cmidrule(lr){1-10}

\multirow{3}{*}{ImportError}
  & Explicit-origin        & \textbackslash{} & 0.33  & \textbackslash{} & \textbackslash{} & \textbackslash{} & \textbackslash{} & \textbackslash{} & \textbackslash{} \\
  & Implicit-origin        & 0.26             & 0.25  & 0.12             & 0.25            & 2.26            & \textbf{2.60}            & 0.00            & 0.00            \\
  & Implicit-disorder      & \textbackslash{} & 0.00  & 0.00             & 0.35            & 1.22            & \textbf{2.87}            & \textbackslash{} & \textbackslash{} \\
\cmidrule(lr){1-10}

\multirow{3}{*}{PulpError}
  & Explicit-origin        & \textbackslash{} & 0.00  & \textbackslash{} & \textbackslash{} & \textbackslash{} & \textbackslash{} & \textbackslash{} & \textbackslash{} \\
  & Implicit-origin        & 0.00             & 0.00  & 0.00             & 0.00            & 0.00            & 0.00            & 0.00            & 0.00            \\
  & Implicit-disorder      & \textbackslash{} & 0.00  & 0.00             & 0.00            & 0.00            & 0.00            & \textbackslash{} & \textbackslash{} \\

\bottomrule
\end{tabular}
}
\end{table}

For the Multi-dimensional Knapsack problems, while ORLM still follows the same rule, other models raise ValueError more frequently. The complete information is given in Table \ref{tab22}.
\clearpage

\subsection{Capacitated warehouse location}

\begin{table}[ht]
\centering
\caption{Capacitated Warehouse Location Problems Performance Metrics}
\label{tab24}
\resizebox{0.9\textwidth}{!}{
\begin{tabular}{llcccccccc}
\toprule
\multirow{2}{*}{Error Type} & \multirow{2}{*}{Dataset}
  & \multicolumn{2}{c}{GPT-4o-mini}
  & \multicolumn{2}{c}{LLAMA3-8B}
  & \multicolumn{2}{c}{ORLM}
  & \multicolumn{2}{c}{DeepSeek-R1} \\
\cmidrule(lr){3-4} \cmidrule(lr){5-6} \cmidrule(lr){7-8} \cmidrule(lr){9-10}
 & & PoT & PoT+CoT & PoT & PoT+CoT & PoT & PoT+CoT & PoT & PoT+CoT \\
\midrule

\multirow{3}{*}{IndexError}
  & Explicit-origin        & \textbackslash{} & 43.98 & \textbackslash{} & \textbackslash{} & \textbackslash{} & \textbackslash{} & \textbackslash{} & \textbackslash{} \\
  & Implicit-origin        & 23.51            & 26.09 & 0.00             & 6.29            & 1.03            & 1.21            & \textbf{31.47}           & 17.88           \\
  & Implicit-disorder      & \textbackslash{} & \textbf{21.05} & 1.81             & 2.94            & 1.21            & 1.19            & \textbackslash{} & \textbackslash{} \\
\cmidrule(lr){1-10}

\multirow{3}{*}{FileNotFoundError}
  & Explicit-origin        & \textbackslash{} & 4.71  & \textbackslash{} & \textbackslash{} & \textbackslash{} & \textbackslash{} & \textbackslash{} & \textbackslash{} \\
  & Implicit-origin        & 0.37             & \textbf{3.56}  & 0.00             & 0.00            & 1.23            & 0.60            & 0.00            & 1.12            \\
  & Implicit-disorder      & \textbackslash{} & 1.91  & 0.00             & 0.00            & 1.21            & \textbf{2.27}            & \textbackslash{} & \textbackslash{} \\
\cmidrule(lr){1-10}

\multirow{3}{*}{OverflowError}
  & Explicit-origin        & \textbackslash{} & 0.52  & \textbackslash{} & \textbackslash{} & \textbackslash{} & \textbackslash{} & \textbackslash{} & \textbackslash{} \\
  & Implicit-origin        & 0.00             & 0.00  & 0.00             & 0.00            & 0.00            & 0.00            & 0.00            & 0.00            \\
  & Implicit-disorder      & \textbackslash{} & 0.00  & 0.00             & 0.00            & 0.00            & 0.00            & \textbackslash{} & \textbackslash{} \\
\cmidrule(lr){1-10}

\multirow{3}{*}{NameError}
  & Explicit-origin        & \textbackslash{} & 6.81  & \textbackslash{} & \textbackslash{} & \textbackslash{} & \textbackslash{} & \textbackslash{} & \textbackslash{} \\
  & Implicit-origin        & 3.36             & \textbf{5.53}  & 1.89             & 2.52            & 4.72            & 2.42            & 2.10            & 4.47            \\
  & Implicit-disorder      & \textbackslash{} & \textbf{3.83}  & 1.81             & 2.35            & 3.74            & 3.22            & \textbackslash{} & \textbackslash{} \\
\cmidrule(lr){1-10}

\multirow{3}{*}{ValueError}
  & Explicit-origin        & \textbackslash{} & \textbf{17.28} & \textbackslash{} & \textbackslash{} & \textbackslash{} & \textbackslash{} & \textbackslash{} & \textbackslash{} \\
  & Implicit-origin        & \textbf{66.79}            & 55.73 & 64.15            & 29.56           & 8.42            & 5.04            & 34.97           & 12.85           \\
  & Implicit-disorder      & \textbackslash{} & 48.09 & \textbf{59.64}            & 56.47           & 4.47            & 3.34            & \textbackslash{} & \textbackslash{} \\
\cmidrule(lr){1-10}

\multirow{3}{*}{AttributeError}
  & Explicit-origin        & \textbackslash{} & 0.00  & \textbackslash{} & \textbackslash{} & \textbackslash{} & \textbackslash{} & \textbackslash{} & \textbackslash{} \\
  & Implicit-origin        & 0.37             & 0.75  & \textbf{3.77}             & \textbf{3.77}            & 1.64            & 2.22            & 0.00            & 0.00            \\
  & Implicit-disorder      & \textbackslash{} & 0.48  & \textbf{3.01}             & 2.35            & 2.66            & 2.27            & \textbackslash{} & \textbackslash{} \\
\cmidrule(lr){1-10}

\multirow{3}{*}{KeyError}
  & Explicit-origin        & \textbackslash{} & \textbf{12.57} & \textbackslash{} & \textbackslash{} & \textbackslash{} & \textbackslash{} & \textbackslash{} & \textbackslash{} \\
  & Implicit-origin        & 0.00             & 0.40  & 0.94             & 1.26            & 0.41            & 0.00            & \textbf{1.40}            & 0.00            \\
  & Implicit-disorder      & \textbackslash{} & 0.00  & \textbf{1.20}             & 1.18            & 0.24            & 0.36            & \textbackslash{} & \textbackslash{} \\
\cmidrule(lr){1-10}

\multirow{3}{*}{SyntaxError}
  & Explicit-origin        & \textbackslash{} & \textbf{8.90}  & \textbackslash{} & \textbackslash{} & \textbackslash{} & \textbackslash{} & \textbackslash{} & \textbackslash{} \\
  & Implicit-origin        & 0.37             & 1.19  & 10.38            & \textbf{14.47}           & 3.08            & 7.66            & 4.90            & 0.00            \\
  & Implicit-disorder      & \textbackslash{} & 0.24  & 18.67            & \textbf{24.12}           & 10.02           & 5.85            & \textbackslash{} & \textbackslash{} \\
\cmidrule(lr){1-10}

\multirow{3}{*}{TypeError}
  & Explicit-origin        & \textbackslash{} & 3.14  & \textbackslash{} & \textbackslash{} & \textbackslash{} & \textbackslash{} & \textbackslash{} & 18.44           \\
  & Implicit-origin        & 5.22             & 4.74  & 1.89             & 3.77            & 3.29            & 1.81            & \textbf{25.17}           & \textbackslash{} \\
  & Implicit-disorder      & \textbackslash{} & 3.11  & \textbf{7.23}             & 5.29            & 1.33            & 0.48            & \textbackslash{} & \textbackslash{} \\
\cmidrule(lr){1-10}

\multirow{3}{*}{ImportError}
  & Explicit-origin        & \textbackslash{} & 1.57  & \textbackslash{} & \textbackslash{} & \textbackslash{} & \textbackslash{} & \textbackslash{} & \textbackslash{} \\
  & Implicit-origin        & 0.00             & 0.40  & 0.00             & 0.00            & \textbf{1.03}            & 0.40            & 0.00            & 0.00            \\
  & Implicit-disorder      & \textbackslash{} & 0.24  & 0.00             & 0.59            & \textbf{0.60}            & 0.48            & \textbackslash{} & \textbackslash{} \\
\cmidrule(lr){1-10}

\multirow{3}{*}{PulpError}
  & Explicit-origin        & \textbackslash{} & \textbf{0.52}  & \textbackslash{} & \textbackslash{} & \textbackslash{} & \textbackslash{} & \textbackslash{} & \textbackslash{} \\
  & Implicit-origin        & 0.00             & 0.00  & 0.00             & 0.00            & 0.00            & 0.00            & 0.00            & 0.00            \\
  & Implicit-disorder      & \textbackslash{} & 0.00  & 0.00             & 0.00            & 0.00            & 0.00            & \textbackslash{} & \textbackslash{} \\

\bottomrule
\end{tabular}
}
\end{table}

For Capacitated warehouse location problems, it shows a similar trend as Generalised Assignment problems. Datasets in disorder perform better than those in origin, which suggests that providing the objective at the beginning of the prompt can enhance the problem-solving ability of LLMs. 

From Table \ref{tab24}, the trend is similar to that of Aircraft Landing Problems.

\clearpage

\subsection{2D-cutting packing-constrained guillotine}\label{2D-Binpacking}

\begin{table}[ht]
\centering
\caption{2D-cutting packing-constrained guillotine Problems Performance Metrics}
\label{tab26}
\resizebox{0.9\textwidth}{!}{
\begin{tabular}{llcccccccc}
\toprule
\multirow{2}{*}{Error Type} & \multirow{2}{*}{Dataset}
  & \multicolumn{2}{c}{GPT-4o-mini}
  & \multicolumn{2}{c}{LLAMA3-8B}
  & \multicolumn{2}{c}{ORLM}
  & \multicolumn{2}{c}{DeepSeek-R1} \\
\cmidrule(lr){3-4} \cmidrule(lr){5-6} \cmidrule(lr){7-8} \cmidrule(lr){9-10}
 & & PoT & PoT+CoT & PoT & PoT+CoT & PoT & PoT+CoT & PoT & PoT+CoT \\
\midrule

\multirow{3}{*}{IndexError}
  & Explicit-origin        & \textbackslash{} & 0.00   & \textbackslash{} & \textbackslash{} & \textbackslash{} & \textbackslash{} & \textbackslash{} & \textbackslash{} \\
  & Implicit-origin        & \textbf{16.67}            & 10.00  & 2.51             & 0.00            & 0.00            & 0.00            & 15.91           & 3.32            \\
  & Implicit-disorder      & \textbackslash{} & 3.23   & \textbf{6.90}             & 0.00            & 0.00            & \textbf{6.90}            & \textbackslash{} & \textbackslash{} \\
\cmidrule(lr){1-10}

\multirow{3}{*}{FileNotFoundError}
  & Explicit-origin        & \textbackslash{} & 0.00   & \textbackslash{} & \textbackslash{} & \textbackslash{} & \textbackslash{} & \textbackslash{} & \textbackslash{} \\
  & Implicit-origin        & 0.00             & 0.00   & 0.00             & 0.00            & 1.18            & 1.18            & 0.00            & \textbf{1.85}            \\
  & Implicit-disorder      & \textbackslash{} & 0.00   & 0.00             & 0.00            & \textbf{0.70}            & 0.00            & \textbackslash{} & \textbackslash{} \\
\cmidrule(lr){1-10}

\multirow{3}{*}{OverflowError}
  & Explicit-origin        & \textbackslash{} & 0.00   & \textbackslash{} & \textbackslash{} & \textbackslash{} & \textbackslash{} & \textbackslash{} & \textbackslash{} \\
  & Implicit-origin        & 0.00             & 0.00   & 0.00             & 0.00            & 0.00            & 0.00            & 0.00            & 0.00            \\
  & Implicit-disorder      & \textbackslash{} & 0.00   & 0.00             & 0.00            & 0.00            & 0.00            & \textbackslash{} & \textbackslash{} \\
\cmidrule(lr){1-10}

\multirow{3}{*}{NameError}
  & Explicit-origin        & \textbackslash{} & 0.00   & \textbackslash{} & \textbackslash{} & \textbackslash{} & \textbackslash{} & \textbackslash{} & \textbackslash{} \\
  & Implicit-origin        & 0.00             & \textbf{10.00}  & 1.76             & 4.35            & 1.18            & 1.18            & 0.00            & 3.32            \\
  & Implicit-disorder      & \textbackslash{} & 0.00   & \textbf{6.90}             & 6.06            & 0.70            & \textbf{6.90}            & \textbackslash{} & \textbackslash{} \\
\cmidrule(lr){1-10}

\multirow{3}{*}{ValueError}
  & Explicit-origin        & \textbackslash{} & \textbf{57.14}  & \textbackslash{} & \textbackslash{} & \textbackslash{} & \textbackslash{} & \textbackslash{} & \textbackslash{} \\
  & Implicit-origin        & 33.33            & \textbf{70.00}  & 7.40             & 8.70            & 0.00            & 3.53            & 22.73           & 8.86            \\
  & Implicit-disorder      & \textbackslash{} & 19.35  & \textbf{58.62}            & 39.39           & 4.90            & \textbf{58.62}           & \textbackslash{} & \textbackslash{} \\
\cmidrule(lr){1-10}

\multirow{3}{*}{AttributeError}
  & Explicit-origin        & \textbackslash{} & 0.00   & \textbackslash{} & \textbackslash{} & \textbackslash{} & \textbackslash{} & \textbackslash{} & \textbackslash{} \\
  & Implicit-origin        & 0.00             & 2.37   & 5.27             & \textbf{13.04}           & 0.00            & 0.00            & 0.00            & 4.06            \\
  & Implicit-disorder      & \textbackslash{} & 0.00   & \textbf{6.90}             & 6.06            & 3.50            & \textbf{6.90}            & \textbackslash{} & \textbackslash{} \\
\cmidrule(lr){1-10}

\multirow{3}{*}{KeyError}
  & Explicit-origin        & \textbackslash{} & 0.00   & \textbackslash{} & \textbackslash{} & \textbackslash{} & \textbackslash{} & \textbackslash{} & \textbackslash{} \\
  & Implicit-origin        & 0.00             & 0.00   & \textbf{6.52}             & 0.00            & 1.18            & 0.00            & 0.00            & 0.37            \\
  & Implicit-disorder      & \textbackslash{} & 0.00   & \textbf{3.45}             & 0.00            & 0.70            & \textbf{3.45}            & \textbackslash{} & \textbackslash{} \\
\cmidrule(lr){1-10}

\multirow{3}{*}{SyntaxError}
  & Explicit-origin        & \textbackslash{} & 0.00   & \textbackslash{} & \textbackslash{} & \textbackslash{} & \textbackslash{} & \textbackslash{} & \textbackslash{} \\
  & Implicit-origin        & 16.67            & 0.00   & \textbf{17.69}            & 0.00            & 1.18            & 4.71            & 2.27            & 9.59            \\
  & Implicit-disorder      & \textbackslash{} & 0.00   & 0.00             & 6.06            & \textbf{6.29}            & 0.00            & \textbackslash{} & \textbackslash{} \\
\cmidrule(lr){1-10}

\multirow{3}{*}{TypeError}
  & Explicit-origin        & \textbackslash{} & 42.86  & \textbackslash{} & \textbackslash{} & \textbackslash{} & \textbackslash{} & \textbackslash{} & \textbackslash{} \\
  & Implicit-origin        & \textbf{33.33}            & 0.00   & 4.77             & 17.39           & 2.35            & 1.18            & 40.91           & 4.06            \\
  & Implicit-disorder      & \textbackslash{} & 0.00   & 13.79             & \textbf{36.36}           & 2.10            & 13.79           & \textbackslash{} & \textbackslash{} \\
\cmidrule(lr){1-10}

\multirow{3}{*}{ImportError}
  & Explicit-origin        & \textbackslash{} & 0.00   & \textbackslash{} & \textbackslash{} & \textbackslash{} & \textbackslash{} & \textbackslash{} & \textbackslash{} \\
  & Implicit-origin        & 0.00             & 0.00   & 0.50             & 0.00            & 0.00            & 0.00            & 0.00            & \textbf{0.74}            \\
  & Implicit-disorder      & \textbackslash{} & 0.00   & 0.00             & 0.00            & 0.00            & 0.00            & \textbackslash{} & \textbackslash{} \\
\cmidrule(lr){1-10}

\multirow{3}{*}{PulpError}
  & Explicit-origin        & \textbackslash{} & 0.00   & \textbackslash{} & \textbackslash{} & \textbackslash{} & \textbackslash{} & \textbackslash{} & \textbackslash{} \\
  & Implicit-origin        & 0.00             & 0.00   & 0.00             & 0.00            & 0.00            & 0.00            & 0.00            & 0.00            \\
  & Implicit-disorder      & \textbackslash{} & 0.00   & 0.00             & 0.00            & 0.00            & 0.00            & \textbackslash{} & \textbackslash{} \\

\bottomrule
\end{tabular}
}
\end{table}

For 2D-cutting packing-constrained guillotine problems, Table \ref{AR}.\ref{MAPE} shows that the performance on datasets in disorder goes down. However, it is unfortunate that we do not have a clear understanding of why this phenomenon occurred. This issue necessitates further investigation in subsequent studies.

From Table \ref{tab26}, ValueError takes the lead. Noticeably, the CoT method or the absence of the CoT method results in an interchange of error types between TypeError and SyntaxError for models such as Deepseek-R1 and LLAMA3. This indicates that if the method without CoT frequently raises a TypeError, the CoT method significantly reduces the occurrence of this error type while simultaneously increasing the occurrence of SyntaxError. Conversely, the opposite trend is also observed.

\section{Statistic}\label{statistic}

\begin{figure}[htbp!] % [h]表示图片尽量放在此处
    \centering % 使图片居中
    \includegraphics[width=0.75\textwidth,keepaspectratio]{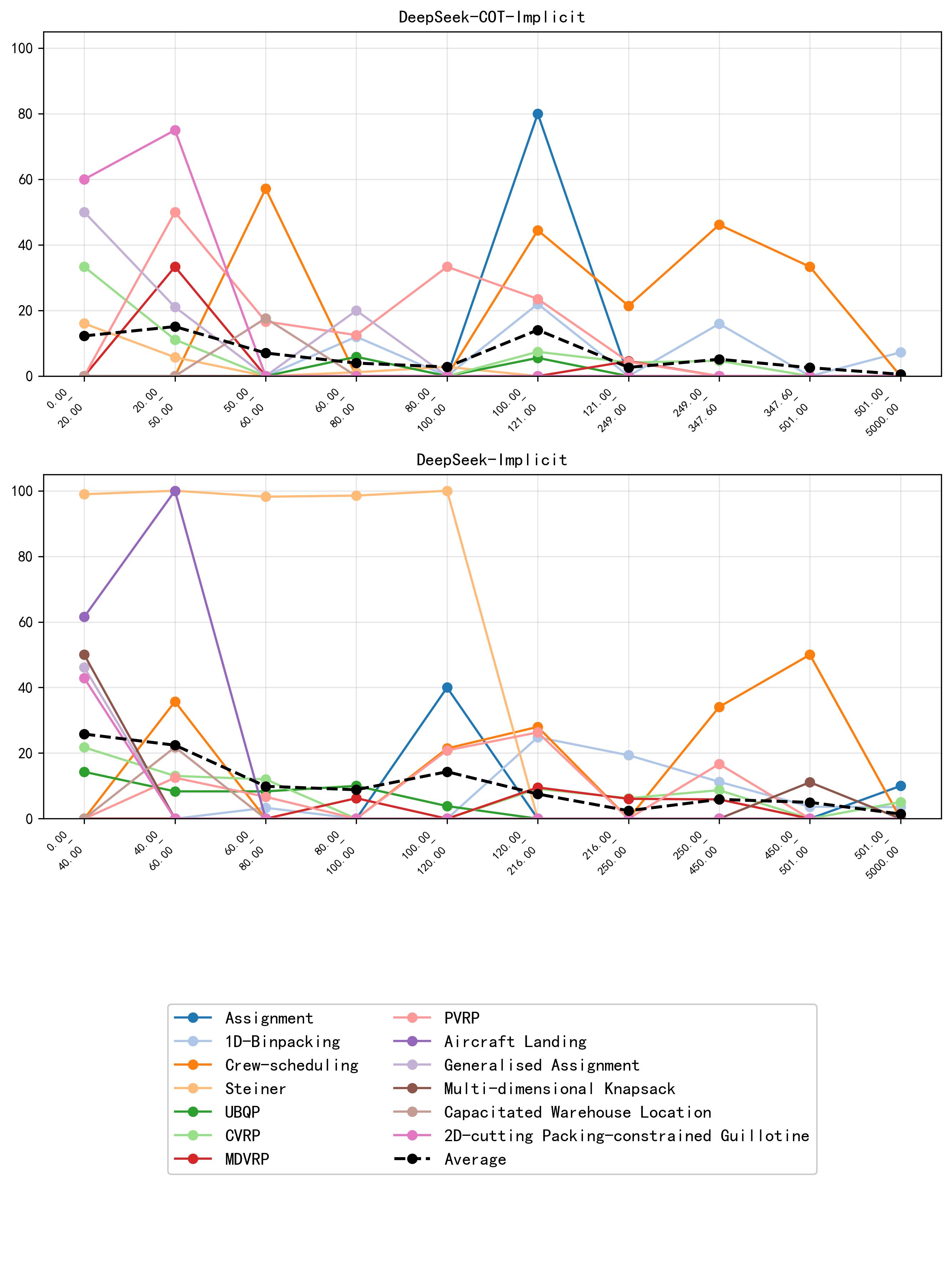} % 插入图片，文件名不需要后缀
    \caption{AR/PR Ratio Across Specified Parameter Range with DeepSeek-Implicit on Different Problems} % 图片标题
    \label{model2} % 图片标签，用于引用
\end{figure}

\begin{figure}[ht] % [h]表示图片尽量放在此处
    \centering % 使图片居中
    \includegraphics[width=0.65\textwidth,keepaspectratio]{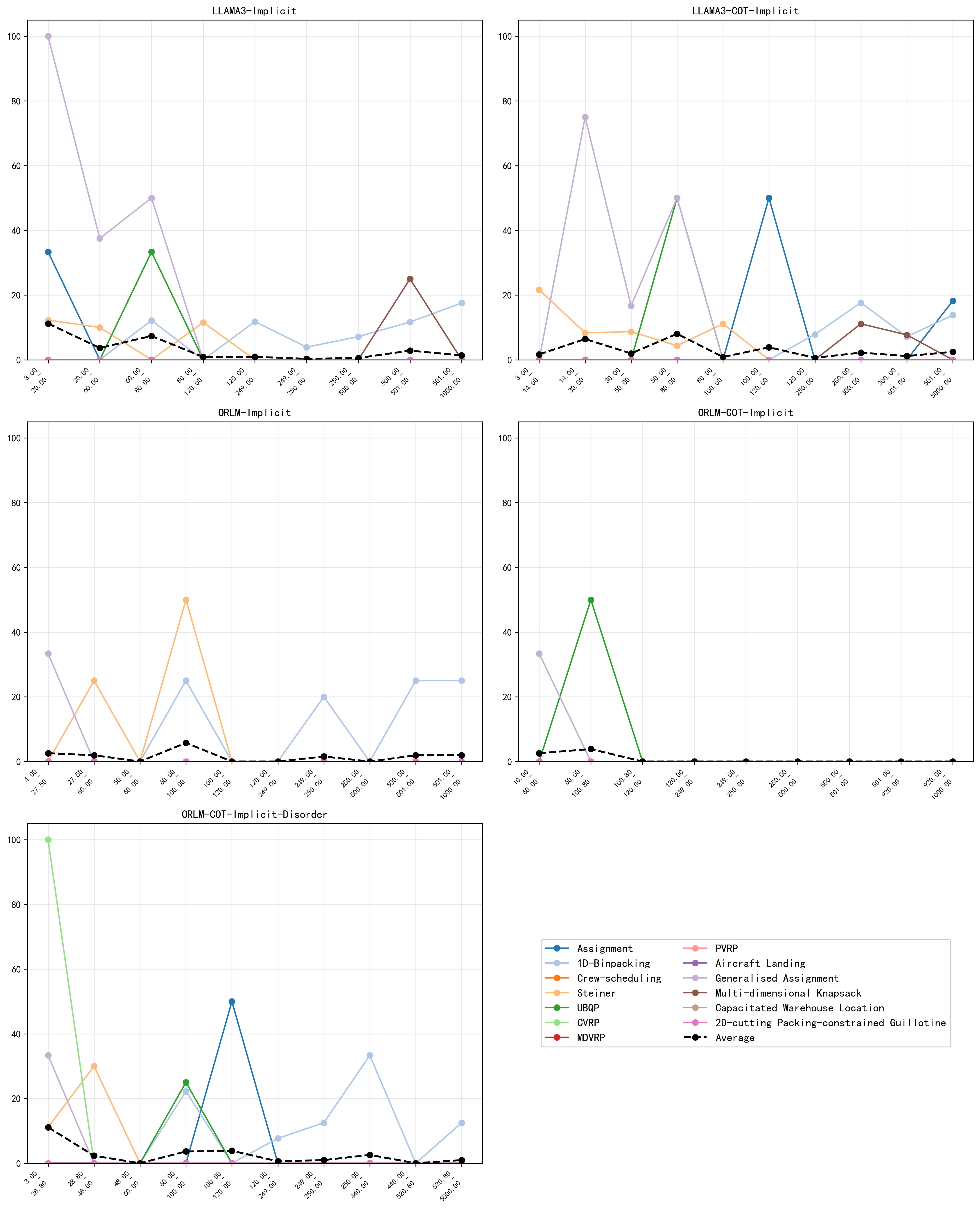} % 插入图片，文件名不需要后缀
    \caption{AR/PR Ratio Across Specified Parameter Range with LLAMA3-Implicit on Different Problems} % 图片标题
    \label{model1} % 图片标签，用于引用
\end{figure}

\clearpage

AR/PR ratio across specified parameter range with different model on different problems is shown in this section. In each figure, the values on the x-axis represent 10\% segments, indicating that for each parameter range, the proportion of problems within that range accounts for 10\% of all issues.

We can observe a common trend, for instance, in cases with fewer parameters or under certain specific conditions, such as the Multi-dimensional Knapsack problems when the number of parameters ranges from 100 to 120, where the accuracy tends to be higher. It means Multi-dimensional Knapsack problems can be more easily solved if the codes can be executed correctly.

By comparing each line chart in Fig.\ref{model1}, it significantly shows that datasets in disorder greatly improve the accuracy rate in general for weak models. By comparing each line chart in Fig.\ref{model2}, some conclusions can be made. For example, AR/PR with DeepSeek-Implicit is unique in terms of Steiner, Aircraft Landing and Crew-scheduling problems, while DeepSeek-CoT-Implicit doesn't show the same trend. In detail, the AR of Steiner problems is nearly 100\% in the case of a small number of parameters ranging from 0 to 120 while Aircraft Landing problems does when the number of parameters range from 40 to 60. Interestingly, for crew scheduling problems, higher number of parameters shows higher accuracy rate, which is contrary to the common trend. Actually, this trend is observed in many other models, which indicates that different models are good at different type of discrete optimization problems. Therefore, you can refer to Fig.\ref{model1} \ref{model2} so that every type of problem can be solved effectively. 

\clearpage

\begin{figure}[ht] % [h]表示图片尽量放在此处
    \centering % 使图片居中
    \includegraphics[width=0.69\textwidth]{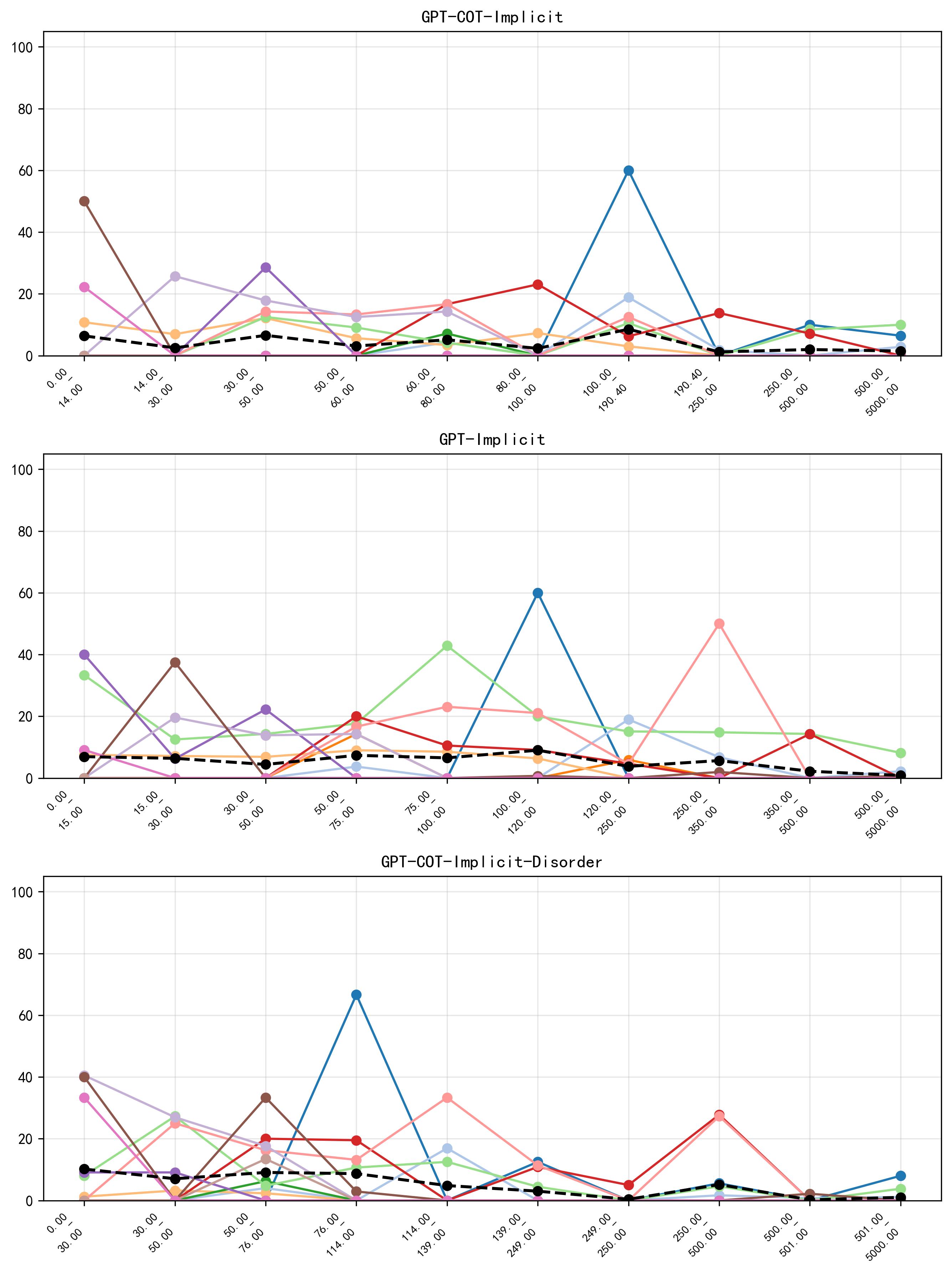} % 插入图片，文件名不需要后缀
    \caption{AR/PR Ratio Across Specified Parameter Range with GPT-4o-mini-CoT-Implicit on Different Problems} % 图片标题
    \label{model3} % 图片标签，用于引用
\end{figure}

By comparing Fig. \ref{model3}, it is observed that although GPT-4o-mini with different technique focus on different problems, the trend is similar for implicit datasets. The average line shows that datasets in disorder activate problems in the low number of parameter range while CoT does not improve the accuracy rate even worsen it in general. 

\clearpage

\begin{figure}[ht] % [h]表示图片尽量放在此处
    \centering % 使图片居中
    \includegraphics[width=0.69\textwidth]{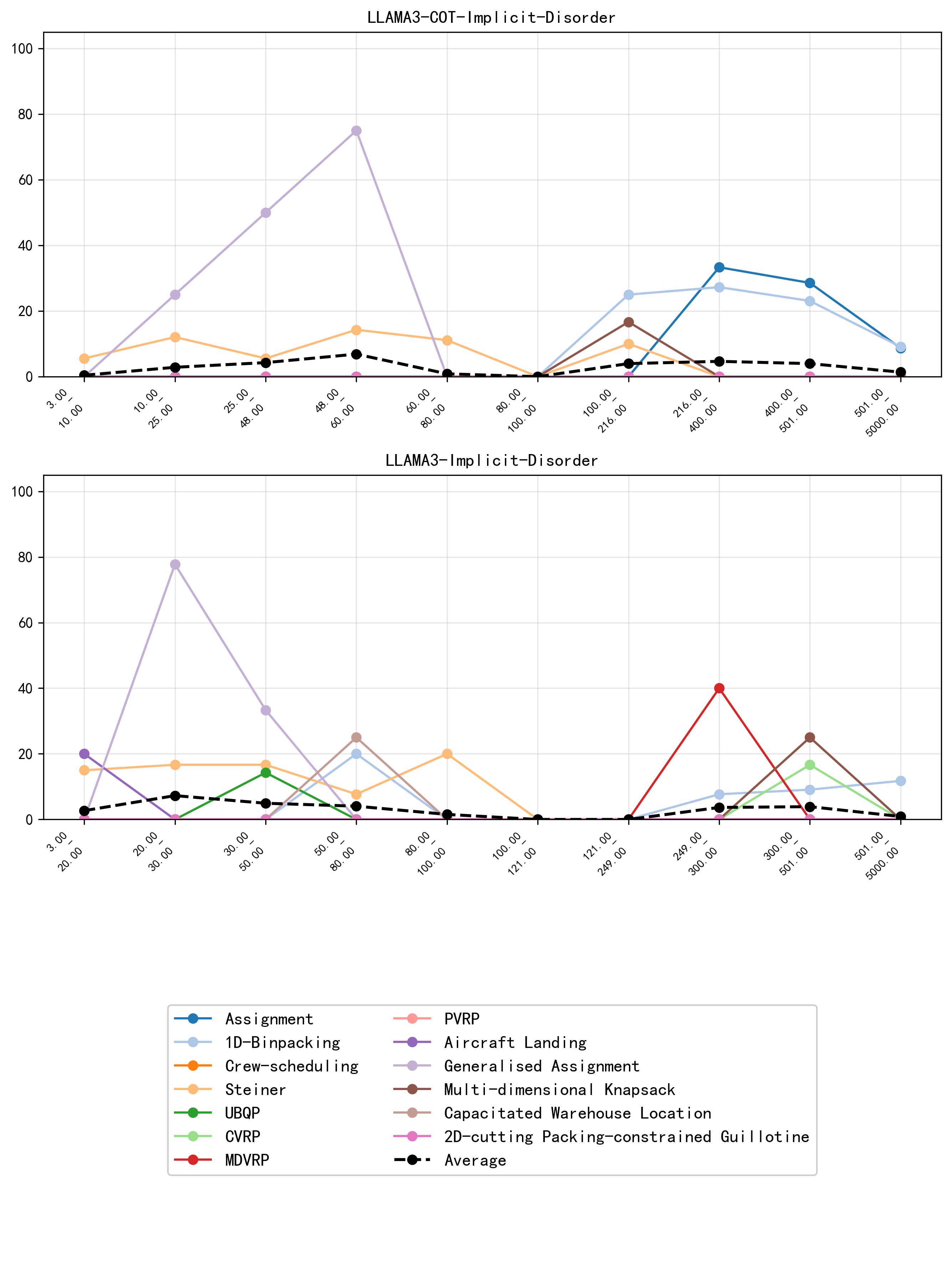} % 插入图片，文件名不需要后缀
    \caption{AR/PR Ratio Across Specified Parameter Range with LLAMA3-CoT-Implicit-Disorder on Different Problems} % 图片标题
    \label{model4} % 图片标签，用于引用
\end{figure}

Also, Fig.\ref{model4} indicates that CoT technique changes the type of problems that have accuracy rate over 0 and the distribution of accuracy rate across parameter range while some problems can not reach optimal results thoroughly. 

\clearpage

For weak models, assignment problems are suitable for specific parameter ranges (3-20, 100-120), 1D-bin packing problems are not sensitive to parameter variations, Crew scheduling problems are not recommended with this type of model, Steiner problems are appropriate for a limited number of parameters, UBQP is best suited for tasks with parameter counts between 60 and 100, CVRP is also discouraged in this context, and if used, should involve fewer parameters. Similarly, MDVRP, PVRP, Aircraft Landing, Capacitated Warehouse Location, and 2D-cutting packing problems are not advisable for this type of problem. The Generalized Assignment Problem is suitable for problems with a moderate number of parameters, while the Multi-Dimensional Knapsack problem is well-suited for LLAMA3 with high number of parameters. 

For strong models, such as DeepSeek-R1 and GPT-4o-mini, both models perform well in 1D Bin-packing problems, particularly in mid to high parameter count scenarios; DeepSeek-R1 significantly outperforms GPT-4o-mini in tackling Crew Scheduling problems; aside from DeepSeek-Implicit, performance across other models and techniques remains relatively uniform; DeepSeek-R1 is identified as more suitable for solving UBQP problems; GPT-Implicit demonstrates the best performance in CVRP; GPT-4o-mini shows a higher adaptability to MDVRP and exhibits insensitivity to parameter variations; The use of COT techniques significantly enhances performance of PVRP; Both models are suitable for low to medium parameter count scenarios for Aircraft Landing Problems; GPT-4o-mini generally outperforms in mid to low parameter ranges for Generalized Assignment Problems; No model appears to be effective in solving Multi-dimensional Knapsack Problems; Similarly, no model demonstrates suitability for Capacitated Warehouse Location problems; Solutions are feasible for low parameter counts, but mid to high parameter counts prove challenging for 2D Cutting Packing-Constrained Guillotine.

If you want to check which model with which technique can lead to higher AR result in detail, please refer to these figures.

\end{document}